\documentclass[lettersize,journal]{IEEEtran}
\usepackage{amsmath,amsfonts}
\usepackage{array}
\usepackage[caption=false,font=normalsize]{subfig}
\usepackage{textcomp}
\usepackage{caption}
\usepackage{stfloats}
\usepackage{url}
\usepackage{verbatim}
\usepackage{graphicx}
\usepackage{cite}
\usepackage{color}
\usepackage{multirow, multicol}
\usepackage{varwidth}
\DeclareCaptionFormat{myformat}{%
  \begin{varwidth}{\linewidth}%
    \centering
    #1#2#3%
  \end{varwidth}%
}
\usepackage[ruled,linesnumbered]{algorithm2e}
\begin{document}

\title{SAM4UDASS: When SAM Meets Unsupervised Domain Adaptive Semantic Segmentation in Intelligent Vehicles}

\author{Weihao Yan,
Yeqiang Qian,~\IEEEmembership{Member,~IEEE,} 
Xingyuan Chen,
Hanyang Zhuang,~\IEEEmembership{Member,~IEEE,}\\
Chunxiang Wang,~\IEEEmembership{Member,~IEEE,}
and Ming Yang,~\IEEEmembership{Member,~IEEE}
\thanks{This work is supported by the National Natural Science Foundation of China (62173228). \emph{(Corresponding author: Ming Yang; Yeqiang Qian.)}}
\thanks{Weihao Yan, Xingyuan Chen, Chunxiang Wang and Ming Yang are with the Department of Automation, Shanghai Jiao Tong University, Key Laboratory of System Control and Information Processing, Ministry of Education of China, Shanghai, 200240, China (email: mingyang@sjtu.edu.cn).}
\thanks{Yeqiang Qian is with the Global Institute of Future Technology, Shanghai Jiao Tong University, Shanghai, 200240, China (qianyeqiang@sjtu.edu.cn).}
\thanks{Hanyang Zhuang is with University of Michigan-Shanghai Jiao Tong University Joint Institute, Shanghai Jiao Tong University, Shanghai, 200240, China (zhuanghany11@sjtu.edu.cn).}}

\markboth{IEEE TRANSACTIONS ON INTELLIGENT VEHICLES, ~Vol.~14, No.~8, August~2021}%
{Yan \MakeLowercase{\textit{et al.}}: SAM for UDA of SS}


\maketitle

\begin{abstract}
 Semantic segmentation plays a critical role in enabling intelligent vehicles to comprehend their surrounding environments. However, deep learning-based methods usually perform poorly in domain shift scenarios due to the lack of labeled data for training. Unsupervised domain adaptation (UDA) techniques have emerged to bridge the gap across different driving scenes and enhance model performance on unlabeled target environments. Although self-training UDA methods have achieved state-of-the-art results, the challenge of generating precise pseudo-labels persists. These pseudo-labels tend to favor majority classes, consequently sacrificing the performance of rare classes or small objects like traffic lights and signs. 
 To address this challenge, we introduce SAM4UDASS, a novel approach that incorporates the Segment Anything Model (SAM) into self-training UDA methods for refining pseudo-labels. It involves Semantic-Guided Mask Labeling, which assigns semantic labels to unlabeled SAM masks using UDA pseudo-labels. Furthermore, we devise fusion strategies aimed at mitigating semantic granularity inconsistency between SAM masks and the target domain. SAM4UDASS innovatively integrate SAM with UDA for semantic segmentation in driving scenes and seamlessly complements existing self-training UDA methodologies. Extensive experiments on synthetic-to-real and normal-to-adverse driving datasets demonstrate its effectiveness. It brings more than 3\% mIoU gains on GTA5-to-Cityscapes, SYNTHIA-to-Cityscapes, and Cityscapes-to-ACDC when using DAFormer and achieves SOTA when using MIC. The code will be available at \url{https://github.com/ywher/SAM4UDASS}.

\end{abstract}

\begin{IEEEkeywords}
Intelligent vehicles, Semantic segmentation, Domain adaptation, Segment Anything Model.
\end{IEEEkeywords}

\section{Introduction}
\begin{figure}[tbp]
        \centering
        \captionsetup[subfloat]{font=scriptsize,labelfont=scriptsize,labelformat=empty}
        \vspace{-0.2cm}
        \subfloat{\includegraphics[width=0.248\linewidth]{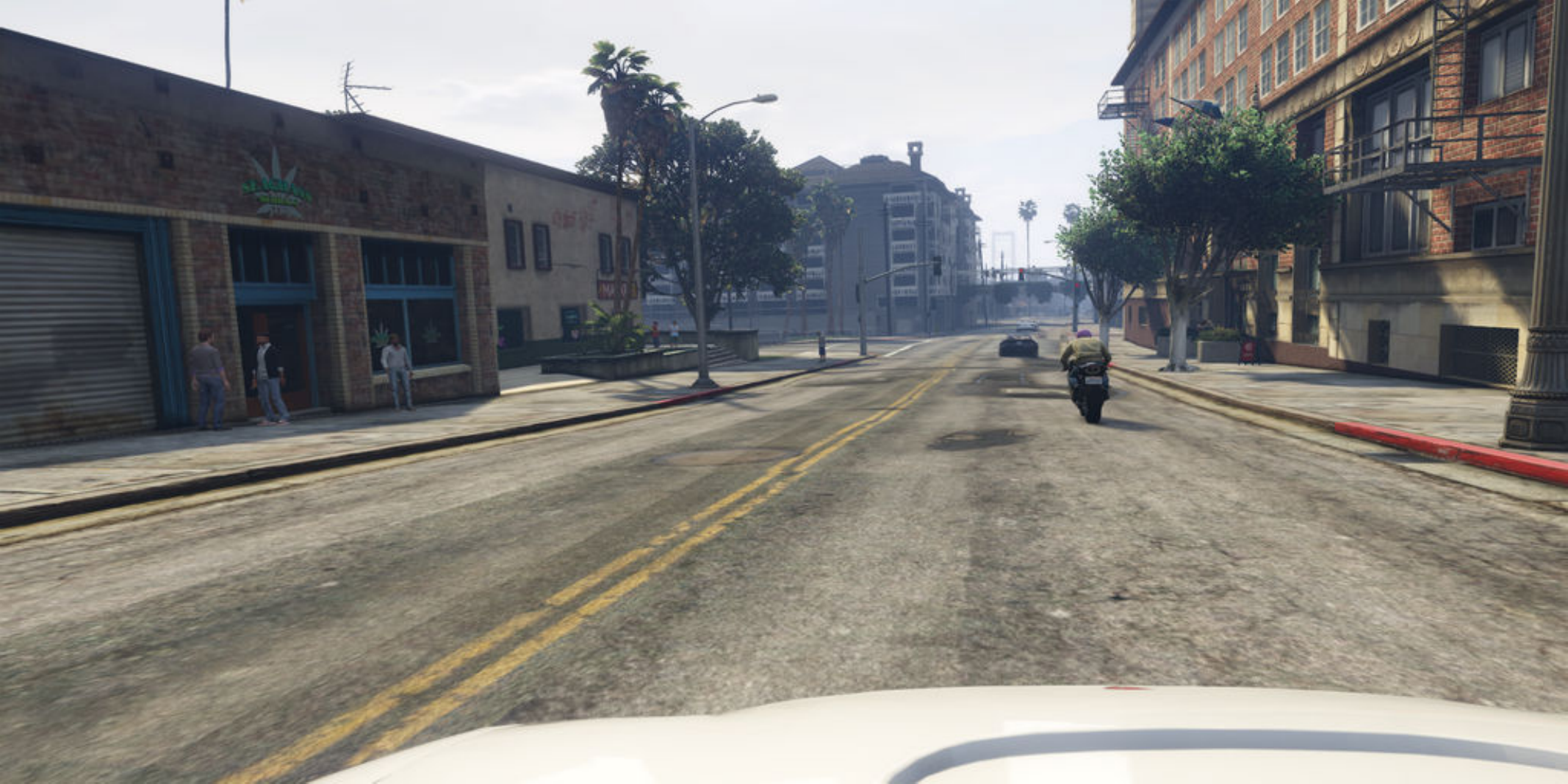}} \hfill
        \subfloat{\includegraphics[width=0.248\linewidth]{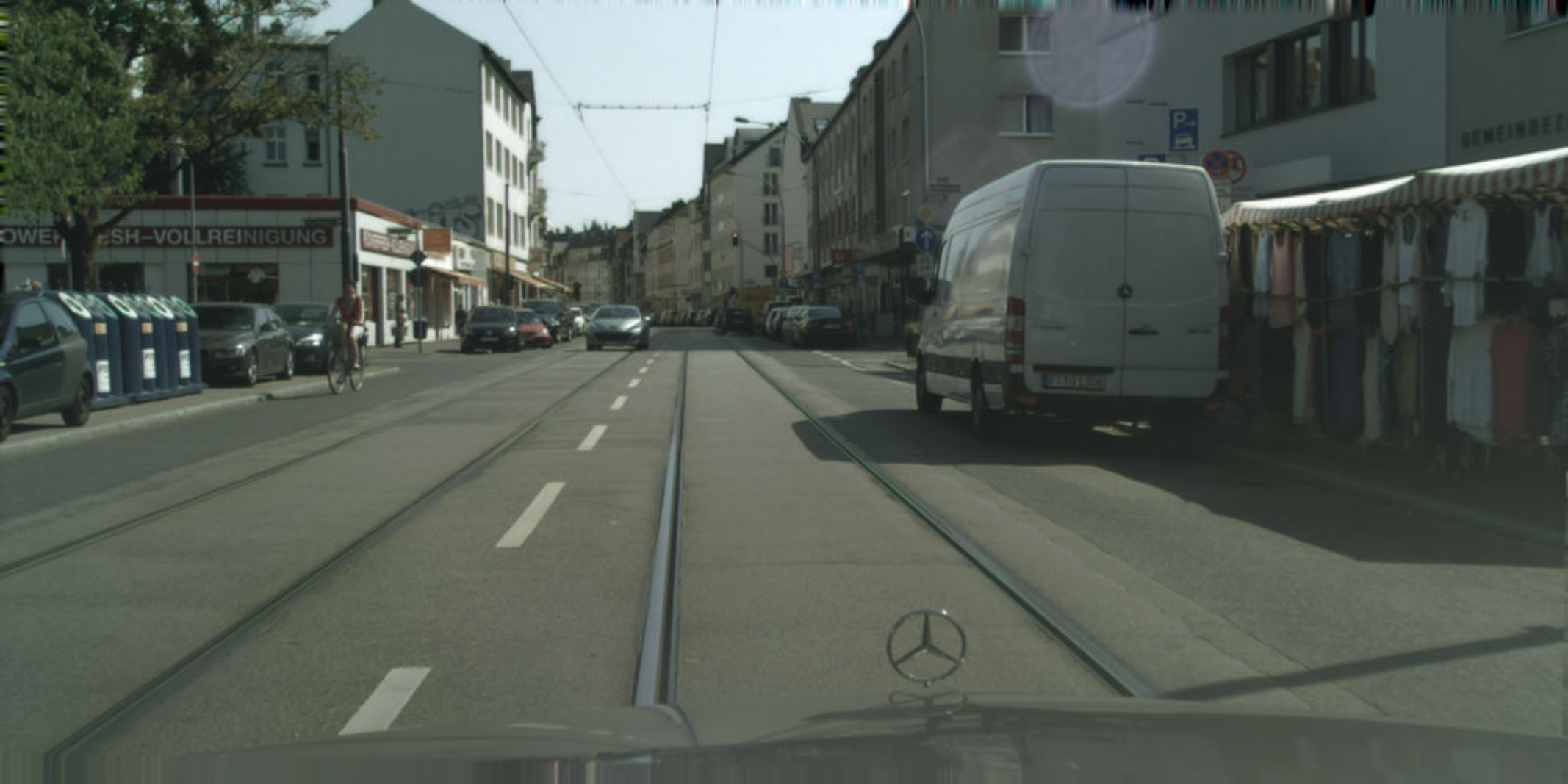}} \hfill
        \subfloat{\includegraphics[width=0.248\linewidth]{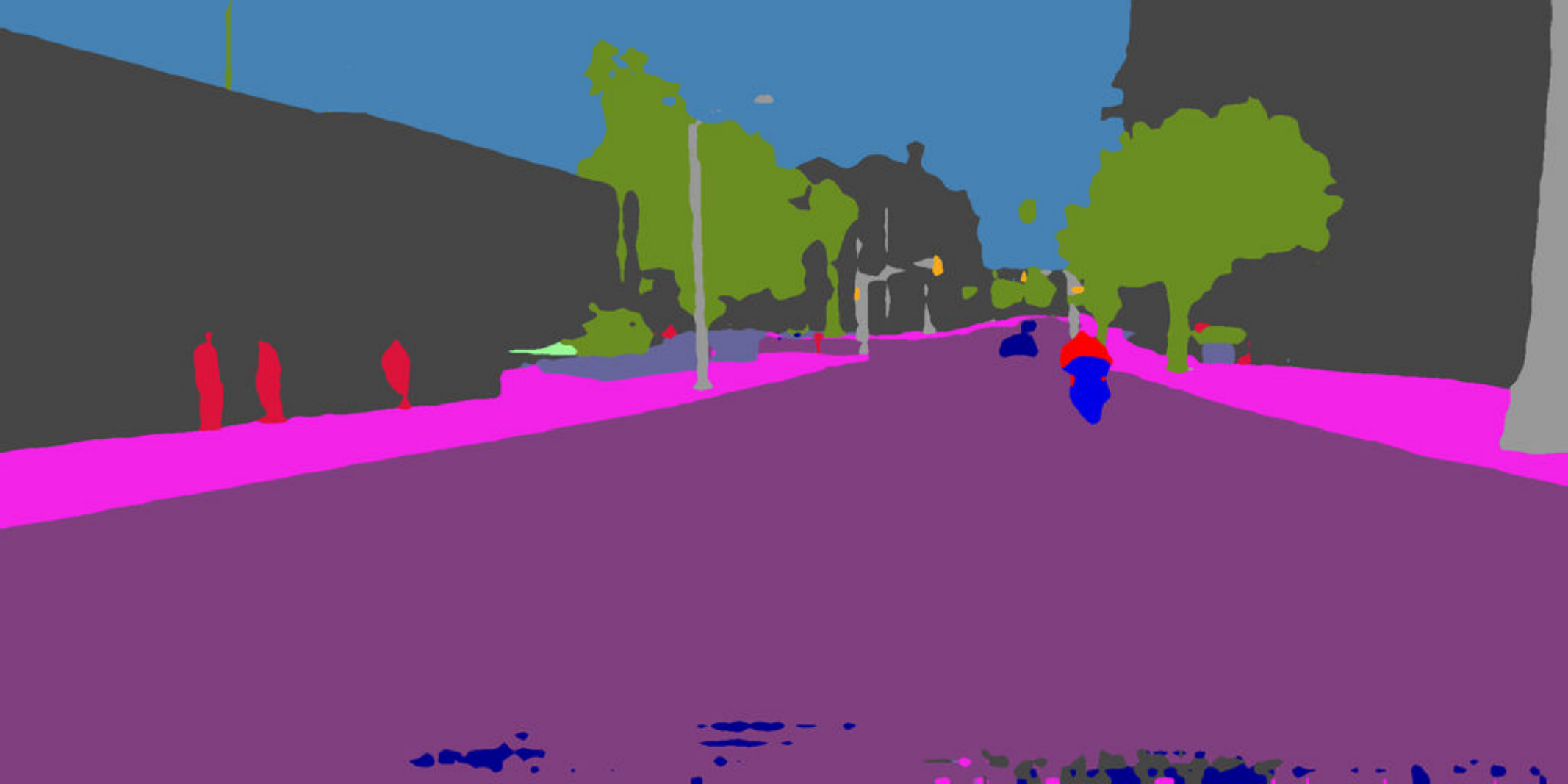}} \hfill
        \subfloat{\includegraphics[width=0.248\linewidth]{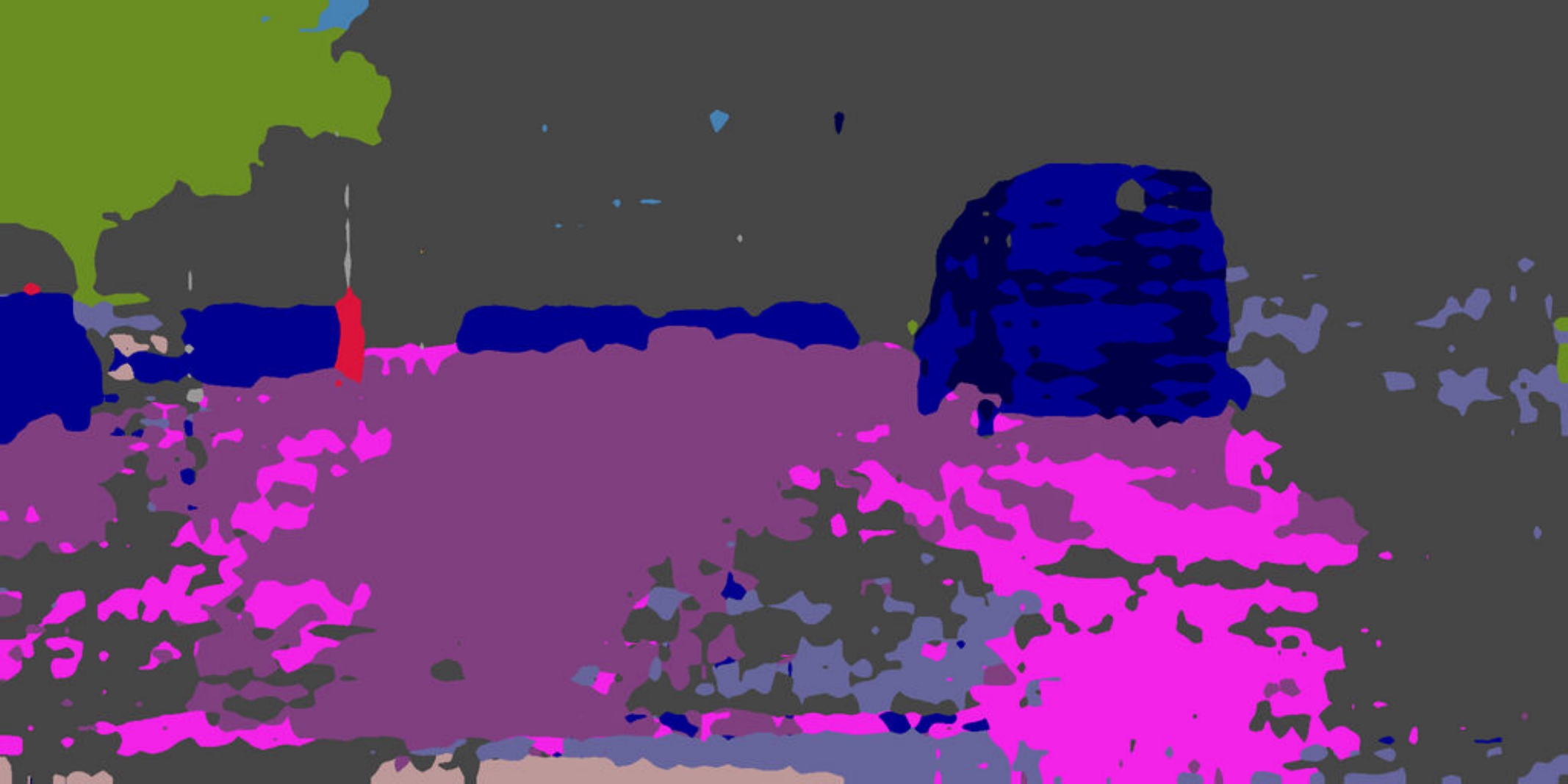}} \\ \vspace{-0.30cm}
        \subfloat[Source domain]{\includegraphics[width=0.248\linewidth]{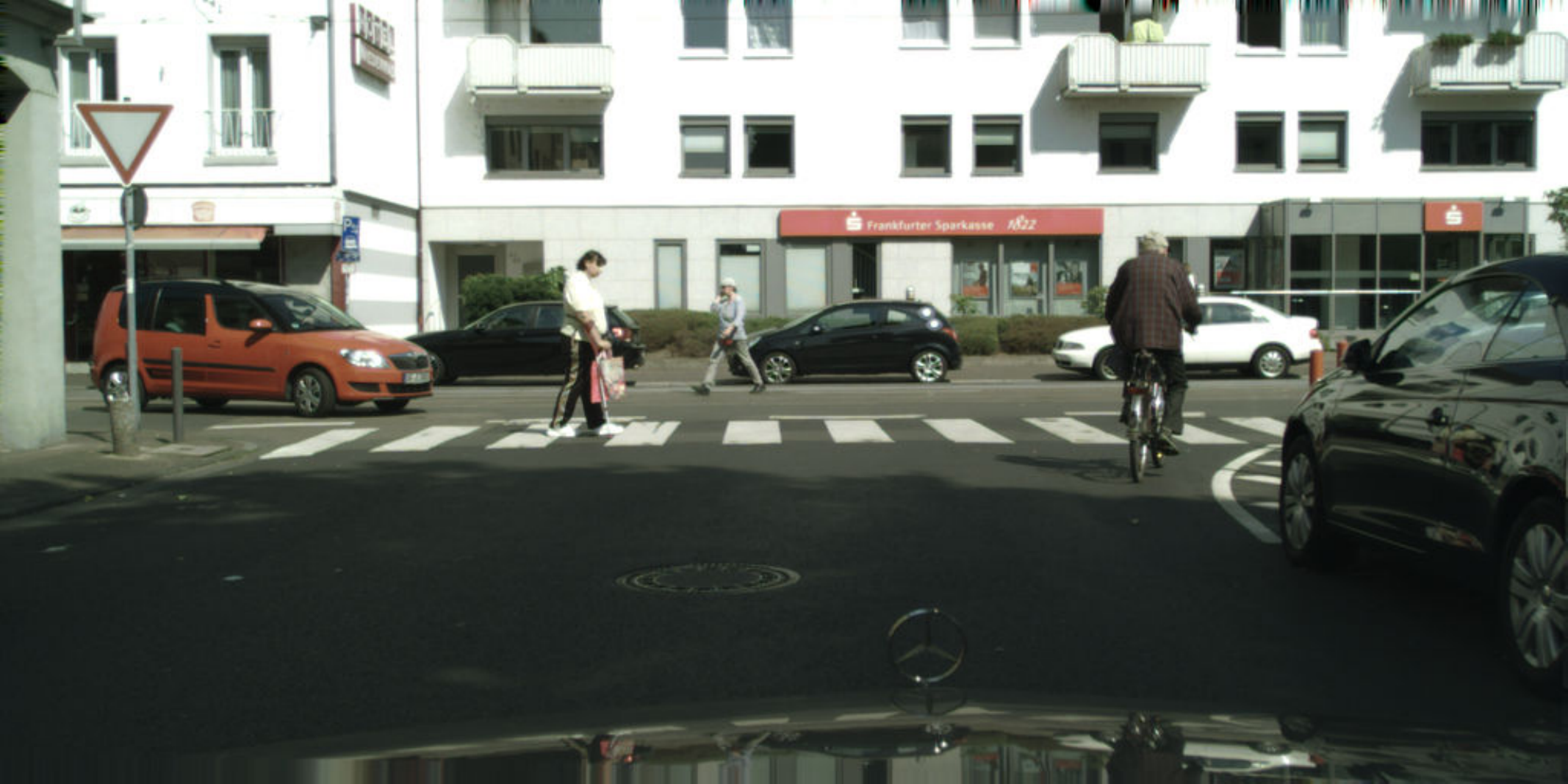}} \hfill
        \subfloat[Target domain]{\includegraphics[width=0.248\linewidth]{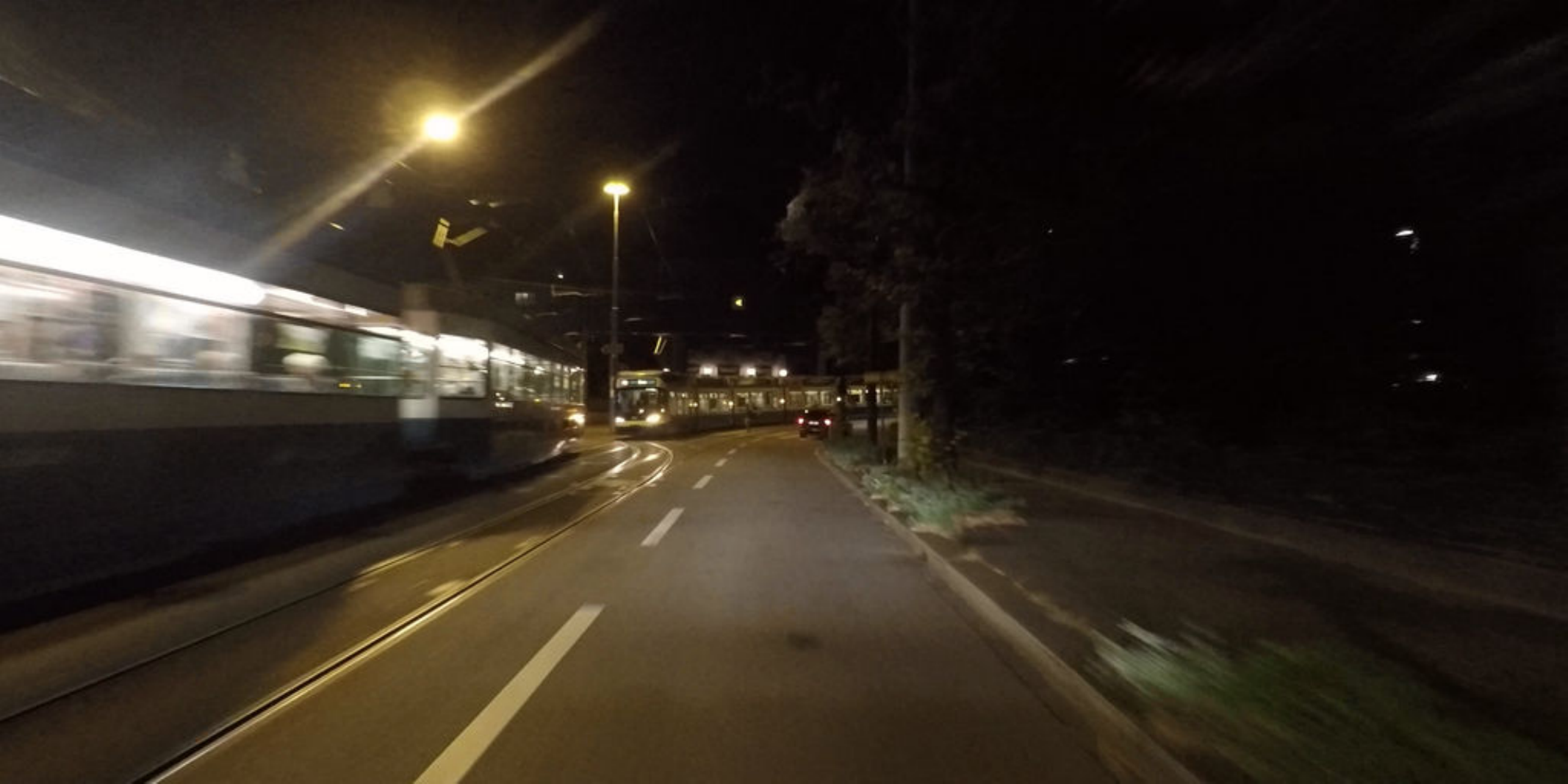}} \hfill
        \subfloat[Source prediction]{\includegraphics[width=0.248\linewidth]{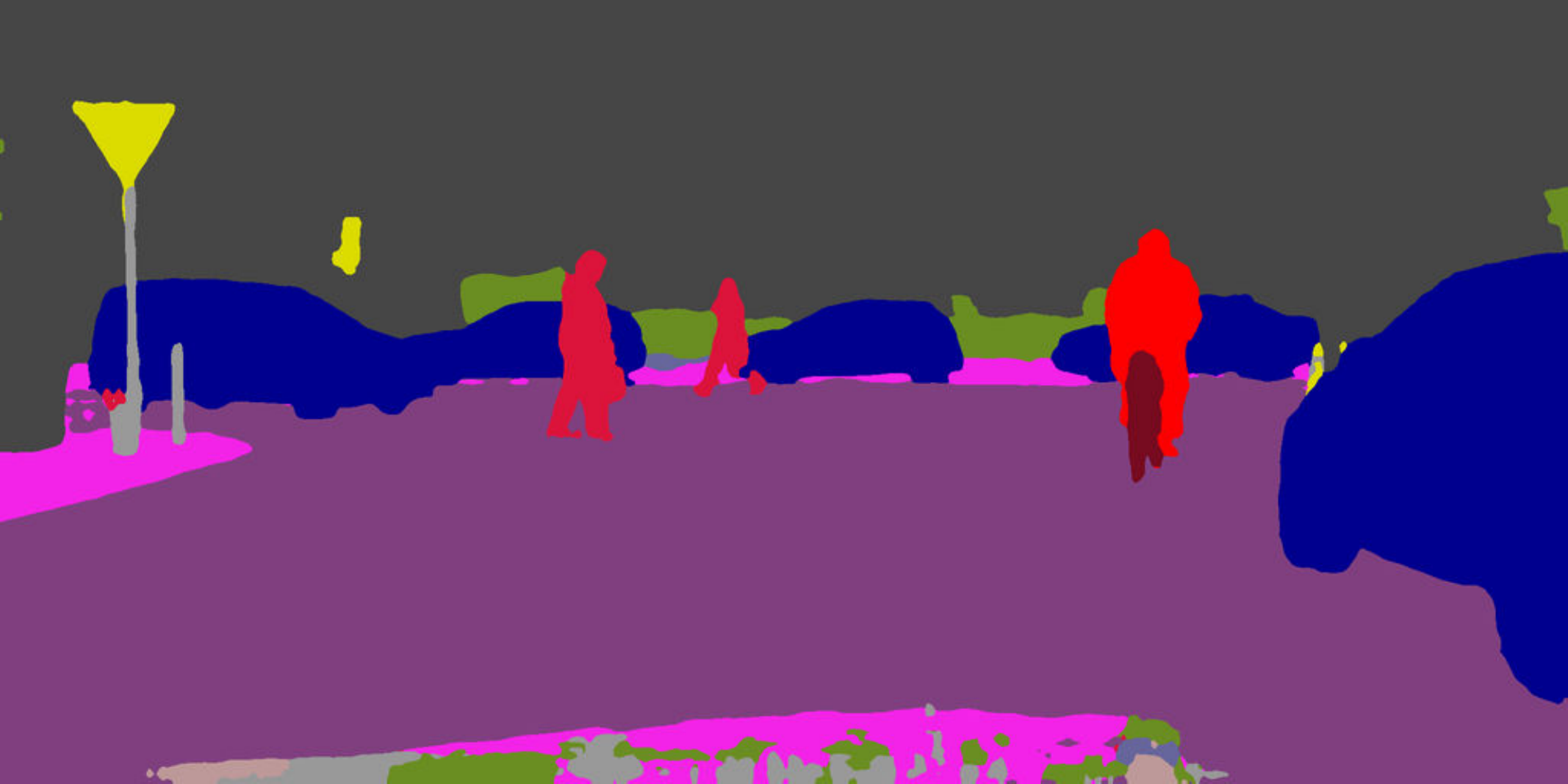}} \hfill
        \subfloat[Target prediction]{\includegraphics[width=0.248\linewidth]{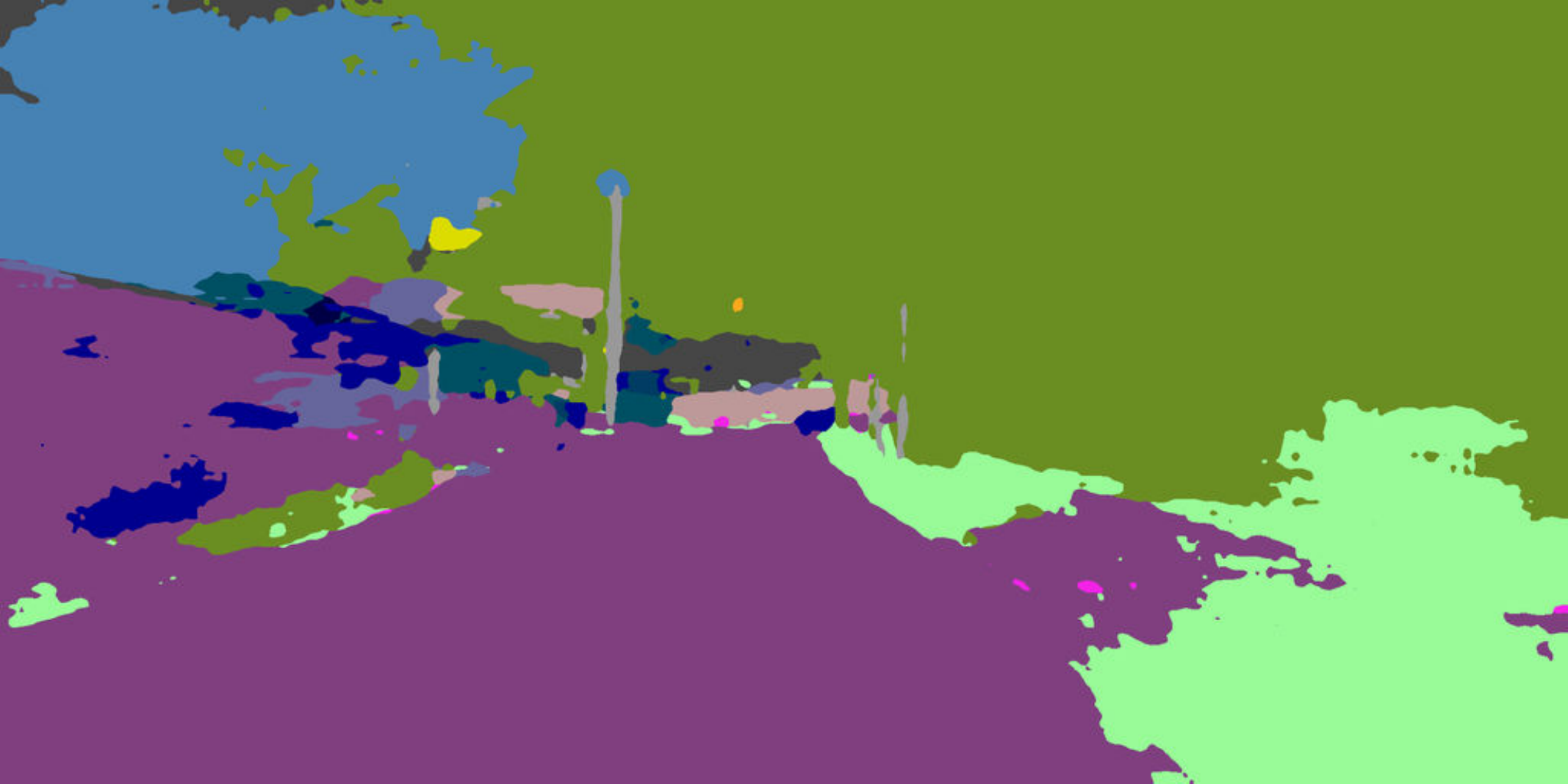}} \\ 
        \caption{Illustration of the cross-domain driving scenes faced by intelligent vehicles and the resulting semantic segmentation model performance deterioration. Left to right: source domain images, target domain images, predictions on source and target domain images using the model trained on source domain.}
        \vspace{-0.5cm}
        \label{fig:cross_domain}
\end{figure}

\IEEEPARstart{S}{emantic} segmentation is a fundamental perception task for intelligent vehicles, aiming to assign each pixel with a class label like road, pedestrian, or building. Remarkable progress has been made in semantic segmentation due to the advent of deep learning and the availability of extensively annotated datasets in recent years\cite{restricted,gated,mlfnet,hyseg}. However, deep learning-based semantic segmentation methods often falter in scenarios characterized by domain shifts, where the labeled training data is both limited and costly to obtain. The labor-intensive annotation process, taking about 1.5 hours for a single Cityscapes\cite{cityscapes} image or even 3.3 hours for a single shot under adverse weather\cite{acdc}, hinders the applicability of deep models to new driving scenes (the target domain). Such new domains may differ significantly from the source domain training dataset in terms of image style, weather conditions, or illumination. Alternatively, driving simulators can generate large-scale annotated data\cite{playingfordata,synthia}, but the disparity between simulated and real-world scenes results in significant performance drop. The cross-domain driving scenes encountered by intelligent vehicles and the resulting performance degradation of semantic segmentation models are depicted in Fig~\ref{fig:cross_domain}.

\begin{figure}[tbp]
        \centering
        \captionsetup[subfloat]{font=scriptsize,labelfont=scriptsize}
        \vspace{-0.2cm}
        \subfloat[Image with ground truth]{\includegraphics[width=0.498\linewidth]{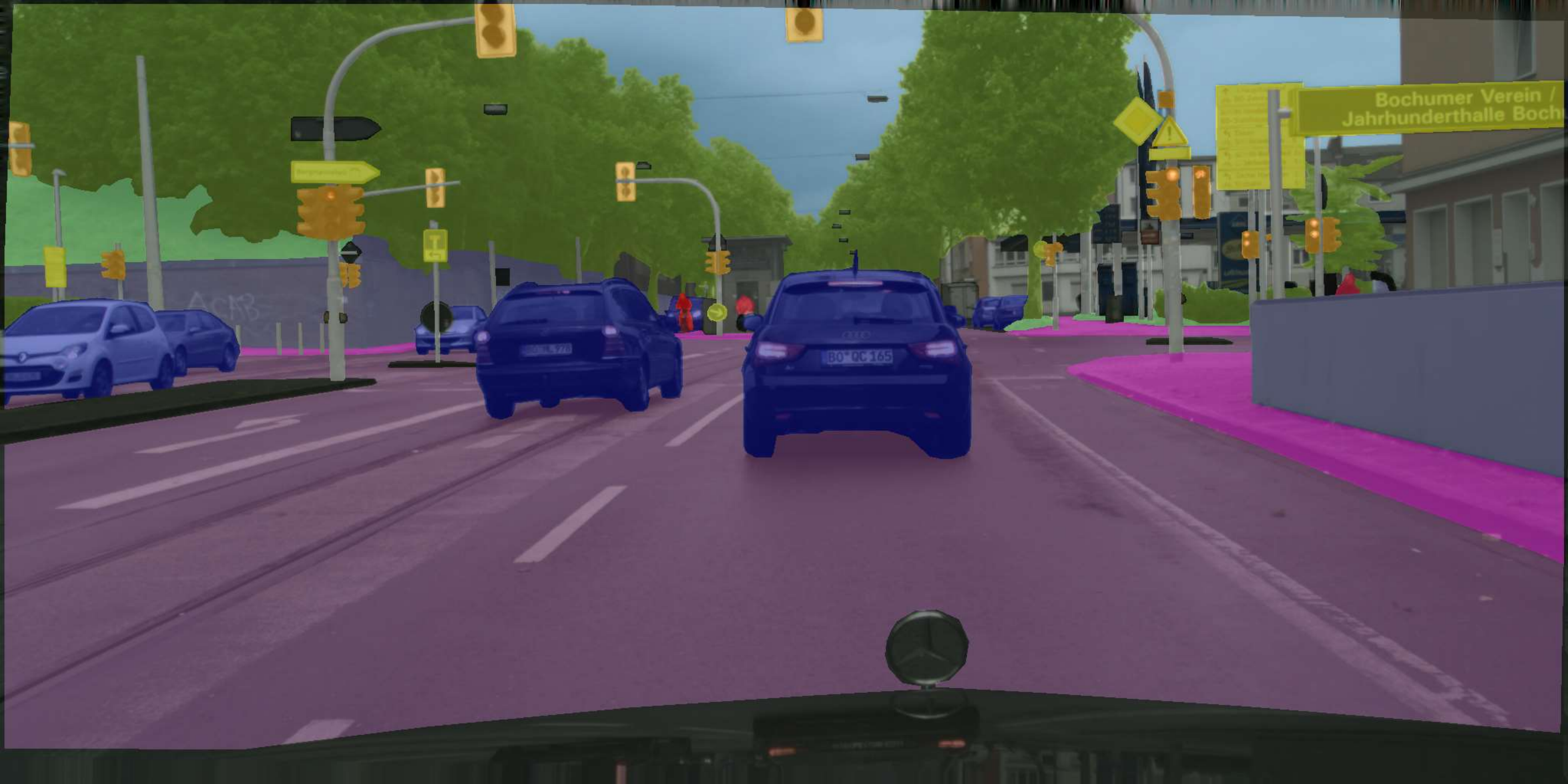}} \hfill
        \subfloat[SAM masks]{\includegraphics[width=0.498\linewidth]{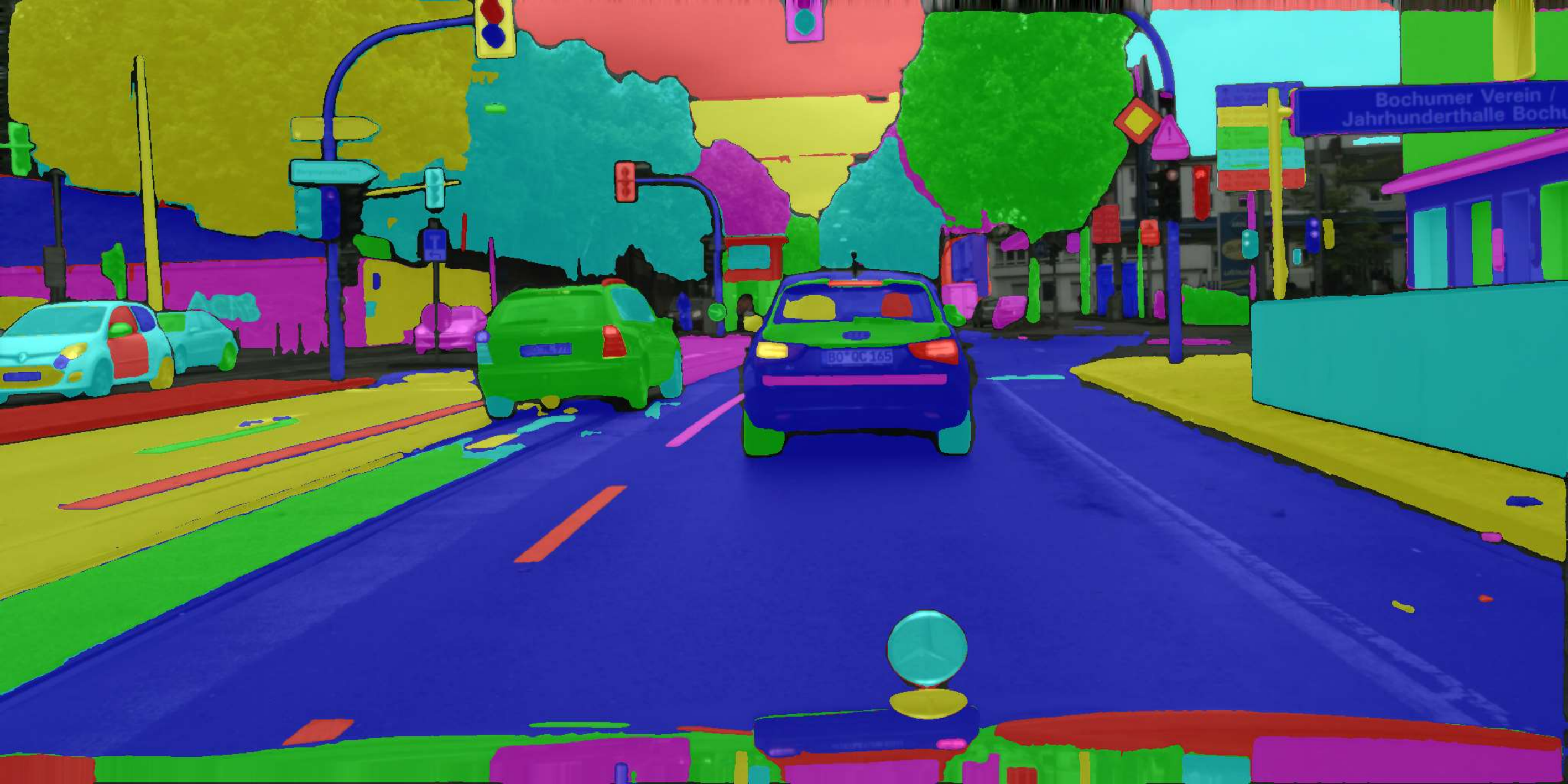}} \\ \vspace{-0.30cm}
        \subfloat[Pseudo-label from DAFormer]{\includegraphics[width=0.498\linewidth]{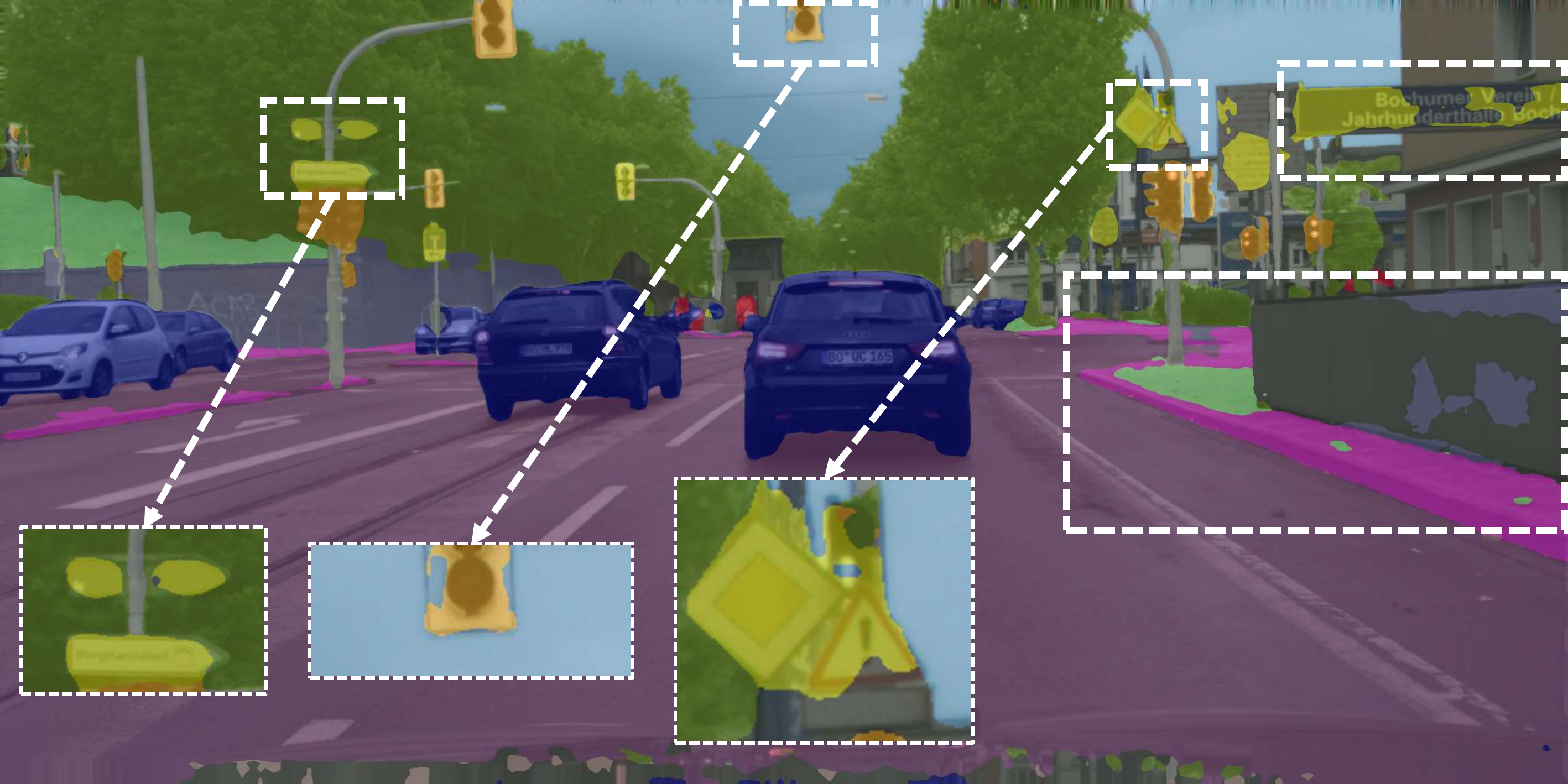}} \hfill
        \subfloat[Pseudo-label refined by SAM4UDASS]{\includegraphics[width=0.498\linewidth]{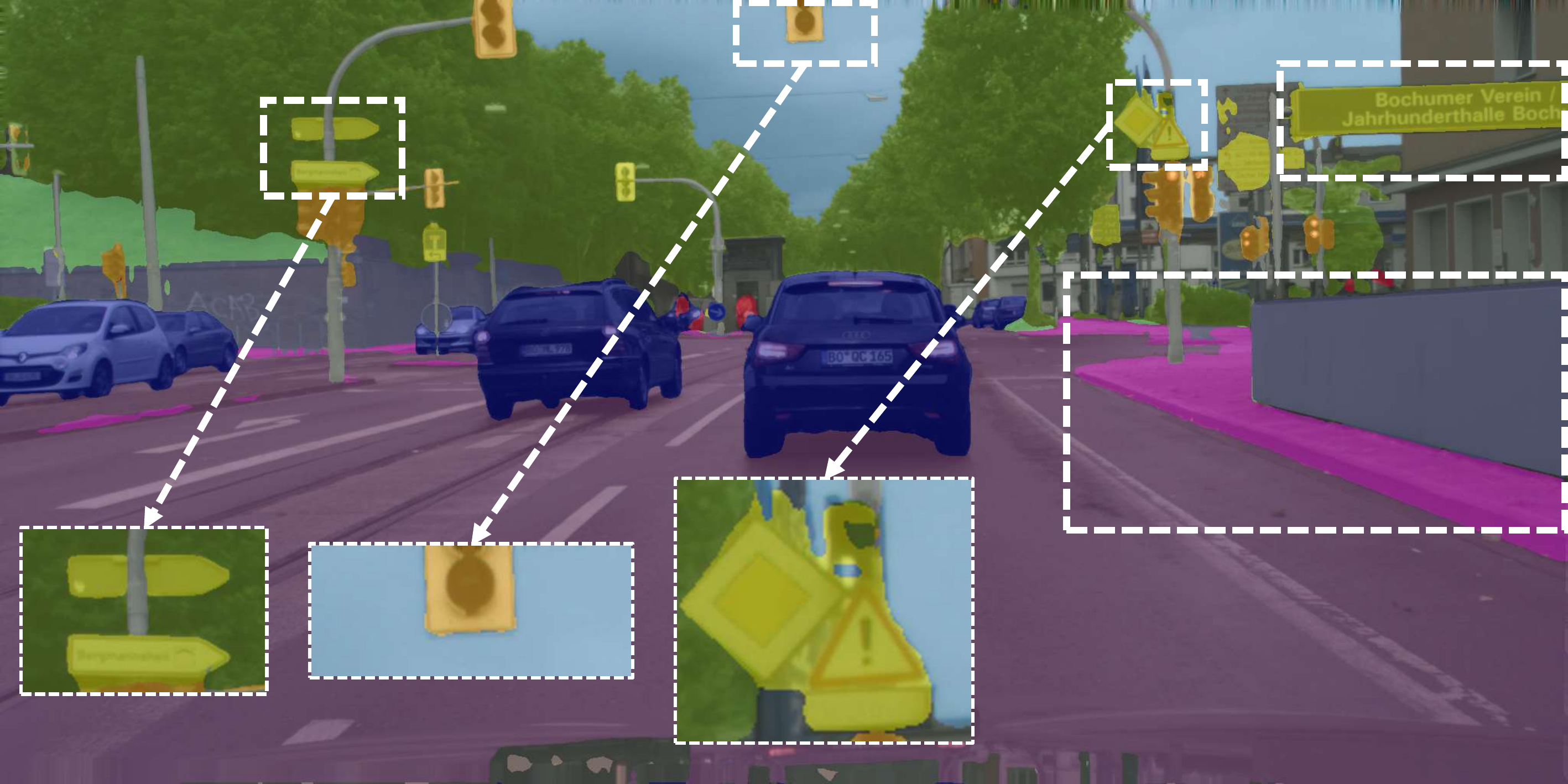}}
        \caption{The demonstration of pseudo-label refinement by SAM. 
        (a) Image with ground truth, (b) SAM masks, (c) Pseudo-label from DAFormer, 
        (d) Pseudo-label refined by SAM4UDASS.
        The improvements are marked with white boxes.}
        \label{fig:pseudo_label_refine}
        \vspace{-0.5cm}
\end{figure}

To address the domain shift challenge, researchers have turned to unsupervised domain adaptation (UDA) techniques, aiming to adapt models trained on labeled source domains to unlabeled target domains\cite{tiv_adversarial,li2022cross,TUFL,transfer,tivparallel}.
A majority of UDA methods for semantic segmentation rely on adversarial training and self-training\cite{mic}. Adversarial training methods employ Generative Adversarial Networks (GANs) \cite{gan} to facilitate domain-invariant feature learning across input, feature, and output levels \cite{advent, bidirectional, leanringoutput}. Self-training methods\cite{TUFL, daformer, mic} generate pseudo-labels for the target domain, achieving state-of-the-art adaptation performance among UDA methods. Theoretically, higher pseudo-label accuracy correlates with superior adaptation performance, as the training process mimics supervised training in the target domain. Recent efforts to enhance pseudo-label quality involve using confidence thresholds\cite{cbsl, iast} or class prototypes\cite{prototypical, cpsl}. Nevertheless, these statistics are undermined by the imbalanced data distribution in unlabeled target domains. Inaccurate pseudo-labels remain a challenge in self-training UDA methods (Fig.~\ref{fig:pseudo_label_refine} (c)), particularly at the boundaries between categories and for rare or small objects within the unbalanced target domain. The quest for reliable pseudo-label generation remains a persistent challenge in unsupervised domain adaptive semantic segmentation (UDASS), impeding the advancement of UDA methods compared with their supervised counterparts. This challenge serves as a driving force for enhancing pseudo-label quality.

The recently released foundation model, Segment Anything Model (SAM)\cite{sam} holds promise in addressing this challenge. SAM is trained on millions of images with billions of masks and demonstrates remarkable zero-shot performance in novel scenarios. While applied to domains such as medical image segmentation and video tracking \cite{sam_medical, video_track}, its potential application to UDA for semantic segmentation in driving scenes has not yet been explored. Benefiting from its extensive training dataset, SAM provides instance-level segmentation results with precise boundaries. However, these results are binary masks lacking semantic labels (Fig.~\ref{fig:pseudo_label_refine} (b)). 


In this paper, we propose SAM4UDASS, a pioneering self-training UDA framework that integrates SAM to enhance the quality of crucial pseudo-labels in cross-domain driving scenarios. We design Semantic-Guided Mask Labeling (SGML) to assign semantic labels to SAM masks based on area ratio information from semantic classes and a road assumption, thereby enhancing pseudo-labels for small or rare objects. Fusion strategies are devised to combine SAM and UDA pseudo-labels, resulting in refined labels and mitigating semantic granularity disparities between SAM masks and the target domain. We conduct extensive experiments on synthetic-to-real and normal-to-adverse driving datasets to validate SAM4UDASS's effectiveness. It achieves mIoU improvements of 3.0\%, 3.7\%, and 5.1\% respectively in GTA5-to-Cityscapes, SYNTHIA-to-Cityscapes, and Cityscapes-to-ACDC scenarios when integrated with DAFormer\cite{daformer}. Additionally, it attains state-of-the-art performance with mIoUs of 77.3\%, 69.3\%, and 70.3\% based on MIC\cite{mic}. Moreover, SAM4UDASS seamlessly integrates into existing self-training UDA methods, consistently enhancing TUFL \cite{TUFL} and SePiCo \cite{sepico}.

Our contributions can be summarized as follows:
\begin{itemize}
        \item We propose SAM4UDASS, a novel self-training UDA framework that primarily incorporates SAM into UDASS to refine the pseudo-labels in cross-domain driving scenarios faced by intelligent vehicles.
        \item We design Semantic-Guided Mask Labeling to assign semantic labels for SAM masks and devise three fusion strategies to alleviate the semantic granularity disparities between SAM masks and the target domain.
        \item Extensive experiments on synthetic-to-real and normal-to-adverse driving datasets demonstrate the effectiveness of SAM4UDASS. The code of SAM4UDASS will be accessible at \url{https://github.com/ywher/SAM4UDASS}.
\end{itemize}
\section{Related work}
\subsection{Semantic Segmentation} 
Semantic segmentation is a fundamental perception task in intelligent vehicles, providing pixel-level semantic information for various downstream tasks like drivable area detection and landmark segmentation. The surge of deep learning has catalyzed the development of advanced semantic segmentation methods. One stream emphasizes enhancing segmentation accuracy through potent model architectures. BEVSS\cite{bevss} proposes a Bird's Eye View segmentation framework integrating a two-stream compact depth transformation and feature rectification. DAFormer \cite{daformer} is tailored to achieve commendable supervised performance alongside domain-adaptation capabilities. Another stream focuses on balancing inference speed and model precision. BiSeNet\cite{bisenet} integrates a spatial path and context path, while MLFNet\cite{mlfnet} leverages a lightweight backbone with a spatial compensation branch and a multi-branch fusion extractor to facilitate real-time inference.

Deep learning models necessitate costly annotated datasets for training. Leveraging simulated data offers a cost-effective alternative. Nevertheless, the significant domain gap between simulations and real-world scenarios often leads to substantial performance degradation. This gap also persists when transitioning from normal weather\cite{cityscapes} to adverse weather\cite{acdc}.

\subsection{UDA for Semantic Segmentation} 
UDA methods for semantic segmentation generally fall into two primary categories: adversarial-based and self-training-based methods. Adversarial-based methods strive to align source and target domains by introducing adversarial learning at input, feature, or output levels. \cite{tiv_adversarial} employs a fully convolutional discriminator to align domain distributions at the output level, while \cite{bidirectional} introduces a bidirectional learning framework that aligns them at both input and output levels. However, adversarial training is susceptible to challenges such as mode collapse and unstable training dynamics.

Self-training UDA methods recently have achieved SOTA adaptation performance by generating pseudo-labels for the target domain. SePiCo\cite{sepico} designs a novel one-stage adaptation framework that highlights the semantic concepts of individual pixels to boost the performance of self-training methods. DAFormer\cite{daformer} introduces the transformer network to the UDASS and employs three training strategies to stabilize training, attaining new SOTA adaptation performance. Additionally, MIC\cite{mic} introduces a Masked Image Consistency module to enhance the spatial context relations of the target domain, further refining DAFormer's performance. Despite these advances, generating high-quality pseudo-labels remains challenging. Pseudo-label boundaries are often imprecise, and rare or small objects are overwhelmed by common objects due to the target domain's unbalanced data distribution. As a result, while self-training methods showcase SOTA adaptation performance, they still lag behind supervised learning counterparts. Improving pseudo-label quality remains a pivotal step in further enhancing their performance. 

\subsection{Segment Anything Model}
The Segment Anything Model (SAM) \cite{sam} has gained prominence for generating high-quality object masks based on prompts such as points or boxes, and can provide masks for all objects in an image. It has impressive zero-shot segmentation performance due to the large-scale SA-1B dataset, with one billion masks and 11 million images. Recent works based on SAM have extended its applications, including MedSAM\cite{sam_medical} for medical image segmentation and Track Anything Model\cite{video_track,xmem} for object tracking and segmentation. Moreover, Anything-3D\cite{anything3d} extends SAM to single-view 3D reconstruction with BLIP\cite{blip} and Stable Diffusion\cite{diffusion}.

Nonetheless, SAM-generated masks lack specific semantic labels, limiting their direct usability in semantic segmentation. Grounded SAM combines Grounding DINO\cite{dino} and SAM to detect and segment objects based on text inputs, enabling masks to carry semantic information. However, its performance relies on Grounding DINO and text input quality, which might not always be stable and accurate, especially for certain background categories. While SEEM \cite{seem} extends SAM's capabilities, its semantic labels are confined to the 80 categories in COCO\cite{coco} and may not align with those required in the target domain. Furthermore, retraining the foundation model is resource-intensive. In this study, we explore the integration of SAM and UDASS, utilizing semantic information from UDA pseudo-labels to guide unlabeled masks generated by SAM, without necessitating SAM's retraining.

\subsection{Fusion Strategy}
Prior fusion methods can be broadly categorized into four levels: signal, feature, result, and multi-level\cite{fusion}. In this context, we concentrate on the result-level for pseudo-label fusion. While\cite{modular} introduces modular fusion strategies to combine outputs from multiple modalities for semantic segmentation, these methods may not be directly applicable to our task, as SAM pseudo-labels are not dense semantic segmentation maps, often featuring unsegmented pixels with void assignments. On the other hand, image-mixing strategies like ClassMix\cite{classmix} find extensive use in UDA for semantic segmentation, serving as an intermediary domain between source and target domains\cite{TUFL,daformer,hrda,mic}. However, these strategies primarily function as data augmentation methods and do not predominantly address pseudo-label fusion. 

\section{Method}
\begin{figure*}[htbp]
        \centering
        \includegraphics[width=1.0\linewidth]{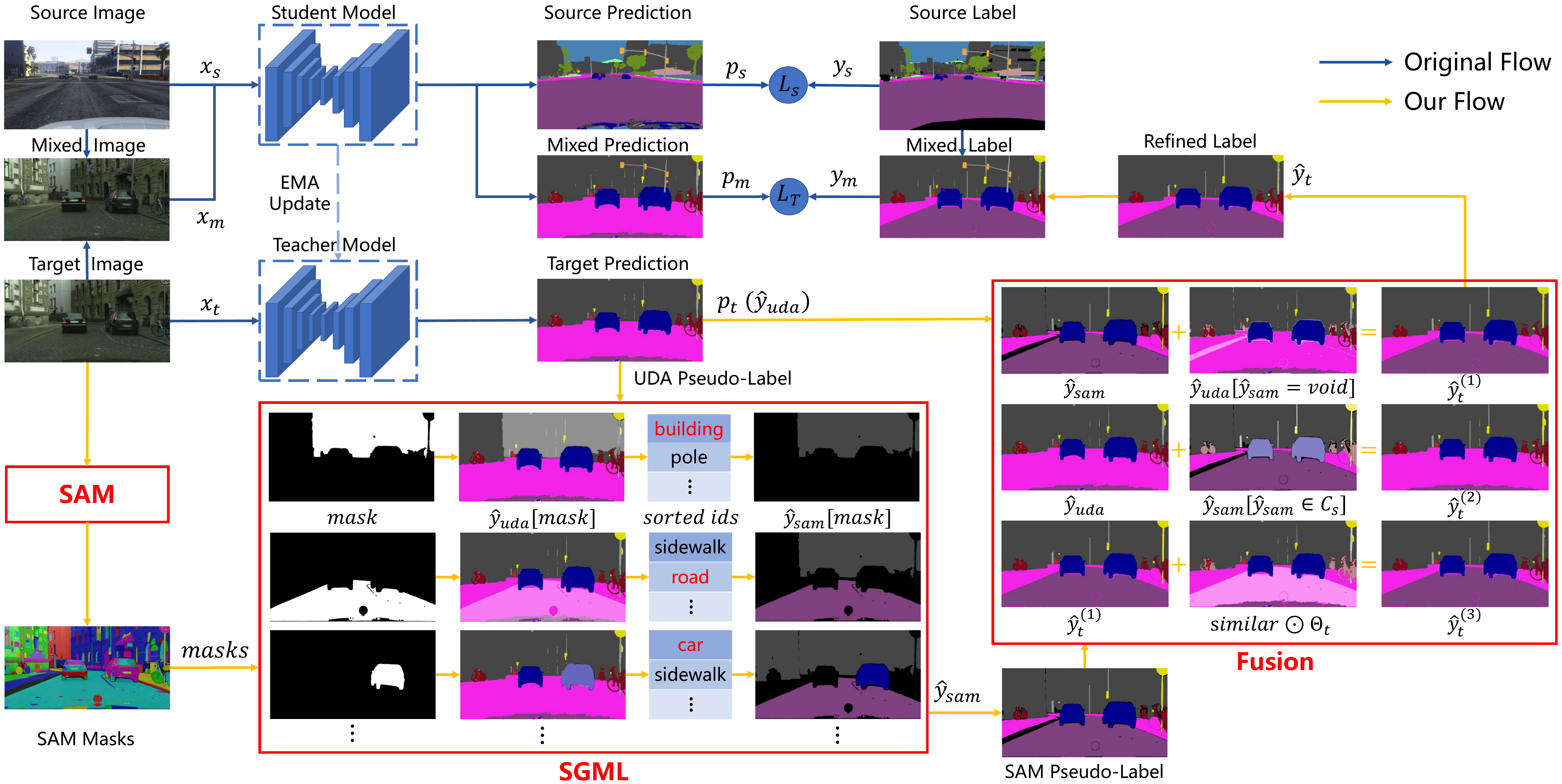}
        \caption{The overview of SAM4UDASS. The blue arrows depict the original self-training methods' flow, while the yellow ones represent our approach. The prediction $p_s$ for source image $x_s$ is supervised by the source label $y_s$. UDA pseudo-label $\hat{y}_{uda}$ and unlabeled masks $masks$ for target image $x_t$ are generated using the teacher model and SAM. Subsequently, SAM pseudo-label $\hat{y}_{sam}$ is derived through SGML. The Fusion module takes ($\hat{y}_{uda}$, $\hat{y}_{sam}$) as inputs and get the refined pseudo-label $\hat{y}_{t}$. ($x_s$, $x_t$) and ($y_s$, $\hat{y}_{t}$) are mixed using ClassMix\cite{classmix} and ($x_m$, $y_m$) are used to train the student network.
        }
        \vspace{-0.3cm}
        \label{fig:SAM4UDASS}
\end{figure*}

In this section, we firstly provide an overview of our SAM4UDASS framework, followed by detailed descriptions of its key components: SGML and three fusion strategies. Illustrated in Fig.\ref{fig:SAM4UDASS}, SAM4UDASS is built upon the self-training paradigm, adopting the teacher-student framework. Here, the teacher model's parameters are updated via Exponential Moving Average (EMA) from the student model.

\subsection{Overview of SAM4UDASS}
For source image $x_s \in R^{3\times H \times W}$, the student model produces prediction $p_s \in R^{C\times H \times W}$, where $C$ is the number of categories, $H$ and $W$ are the height and width of the image, respectively. The cross-entropy loss $L_S(p_s, y_s)$ supervises $p_s$ using ground truth $y_s \in N^{C\times H \times W}$:

\begin{equation}
        L_S(p_s, y_s) = -\frac{1}{H\cdot W} \sum_{n=1}^{H\cdot W} \sum_{c=1}^{C} y_s^{n,c} \log p_s^{n,c}
        \label{eq:source_cross_entropy}
\end{equation}
where $n$ represents the pixel location in the image and $c$ denotes the channel index.

On the other hand, the teacher network processes target image $x_t$ to generate pseudo-label $\hat{y}_{uda}$. By mixing $x_t$ with source image $x_s$ through ClassMix\cite{classmix}, a mixed image $x_m$ is produced. It undergoes student model inference, yielding mixed prediction $p_m$. In contrast to previous self-training methods that directly use pseudo-label $\hat{y}_{uda}$ for training, SAM4UDASS enhances $\hat{y}_{uda}$ by refining it with pseudo-label $\hat{y}_{sam}$ from SGML. Specifically, SAM processes $x_t$ to produce unlabeled masks $masks$, and the pseudo-label $\hat{y}_{sam}$ is obtained through the SGML module under the guidance of $\hat{y}_{uda}$. Subsequently, UDA and SAM pseudo-labels $\hat{y}_{uda}$ and $\hat{y}_{sam}$ are combined via the fusion strategies, resulting in the refined pseudo-label $\hat{y}_{t}$: 
\begin{equation}
        \begin{aligned}
                \hat{y}_{uda} &= Teacher(x_t) \\
                \hat{y}_{sam} &= SGML(SAM(x_t), \hat{y}_{uda}) \\
                \hat{y}_{t} &= Fusion(\hat{y}_{uda}, \hat{y}_{sam}) \\
        \end{aligned}
        \label{eq:y_sam}
\end{equation}

Then, the image and label pairs ($x_s$, $x_t$), ($y_s$, $\hat{y}_{t}$) from the source and target domains are mixed using the class-level mixing strategy ClassMix\cite{classmix}:
\begin{equation}
        \centering
        \begin{aligned}
                x_m &= M \cdot x_s + (1-M) \cdot x_t \\
                y_m &= M \cdot y_s + (1-M) \cdot \hat{y}_{t} \\
        \end{aligned}
        \label{eq:mixup}
\end{equation}
where M is a binary mask formed by randomly selecting half of the classes in $y_s$. The cross-entropy loss $L_T(p_m, y_m)$ is applied to supervise the student network with mixed image-label pairs ($x_m$, $y_m$):
\begin{equation}
        L_T(p_m, y_m) = -\frac{1}{H\cdot W} \sum_{n=1}^{H\cdot W} \sum_{c=1}^{C} y_m^{n,c} \log p_m^{n,c}
        \label{eq:mix_cross_entropy}
\end{equation}

The overall loss of the SAM4UDASS training process is  a combination of $L_S$ and $L_T$:
\begin{equation}
        Loss = L_S(p_s, y_s) + \lambda_t L_T(p_m, y_m)
        \label{eq:total_loss}
\end{equation}
where $\lambda_t$ is the weight of $L_T$ and set as 1.0 in the experiments following previous self-training UDA works\cite{daformer, mic, hrda}.

\subsection{Semantic-Guided Mask Labeling}

\begin{figure}[htbp]
        \centering
        \captionsetup[subfloat]{font=scriptsize,labelfont=scriptsize}
        \vspace{-0.5cm}
        \subfloat[Image]{\includegraphics[width=0.33\linewidth]{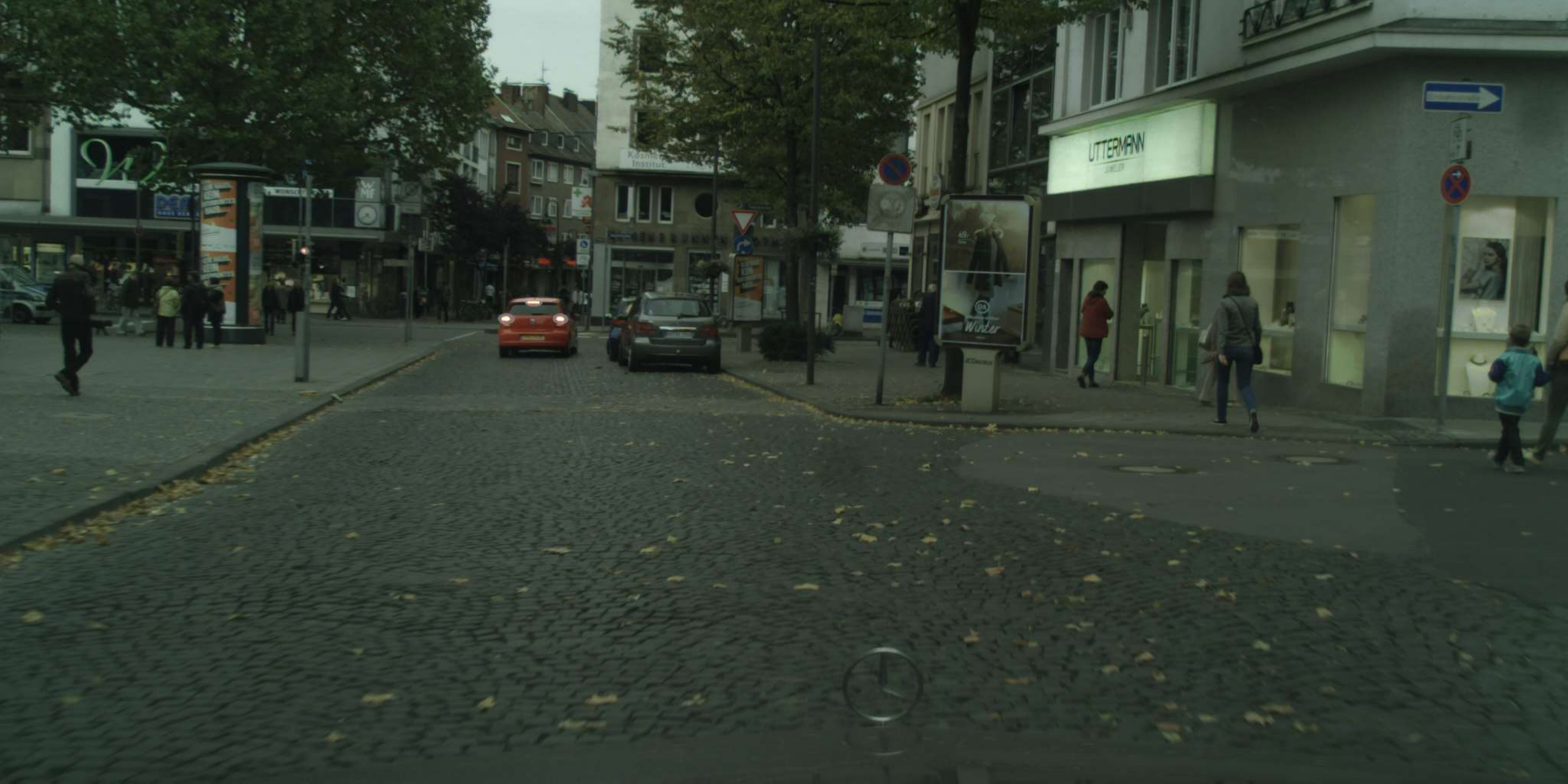}} \hfill
        \subfloat[SAM masks]{\includegraphics[width=0.33\linewidth]{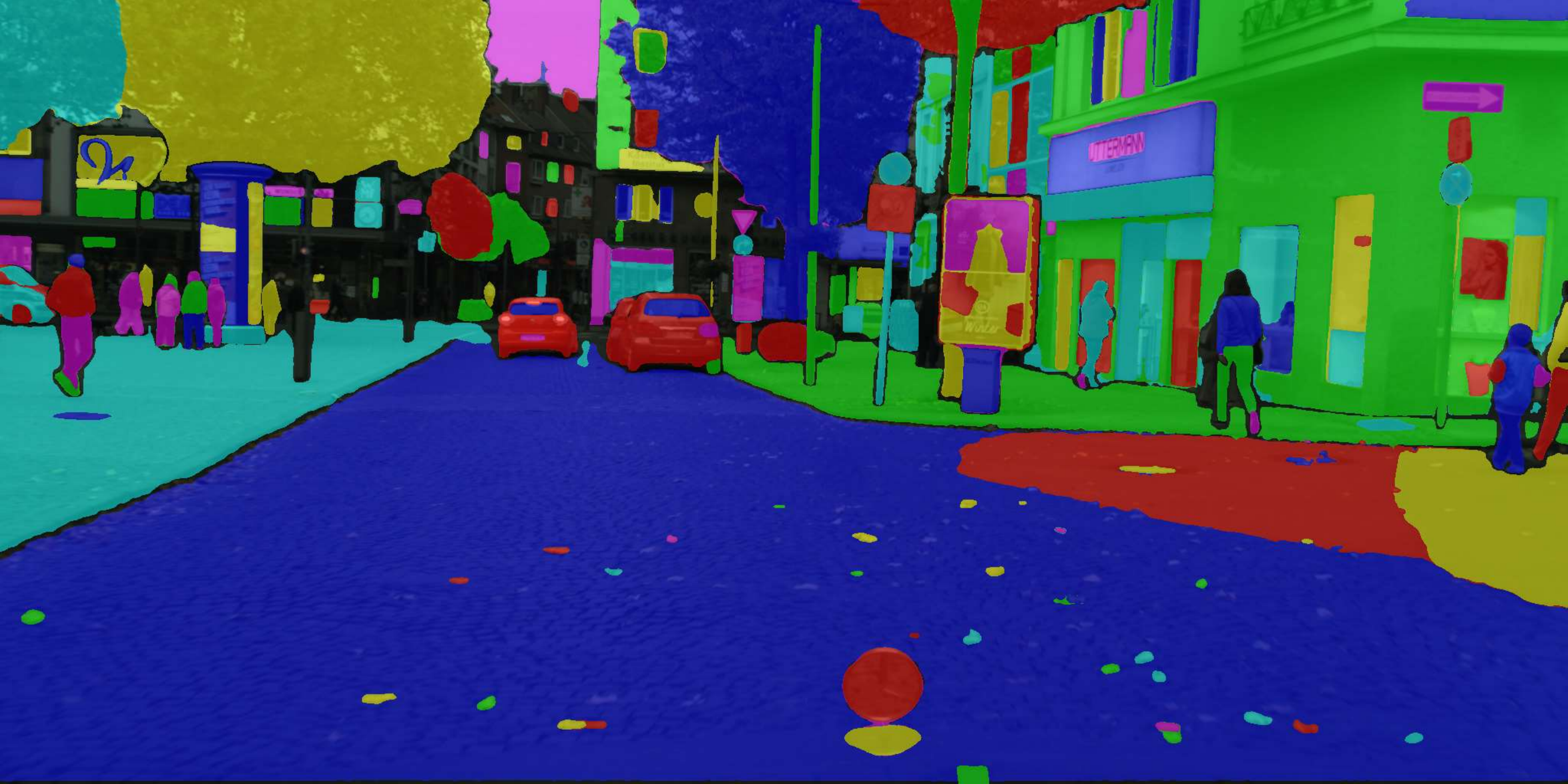}} \hfill
        \subfloat[Ground truth]{\includegraphics[width=0.33\linewidth]{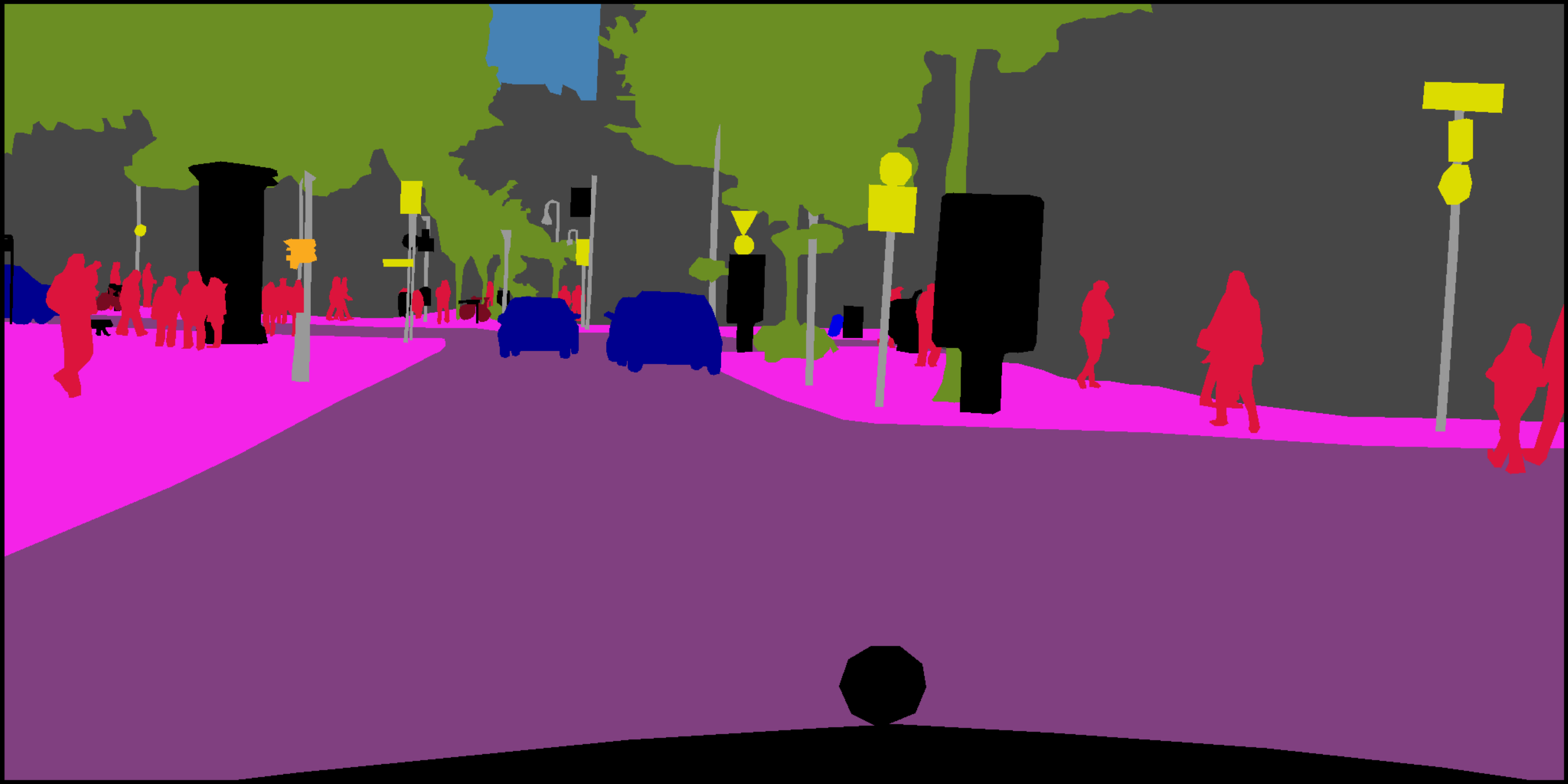}}\\ \vspace{-0.30cm}
        \subfloat[UDA pseudo-label]{\includegraphics[width=0.33\linewidth]{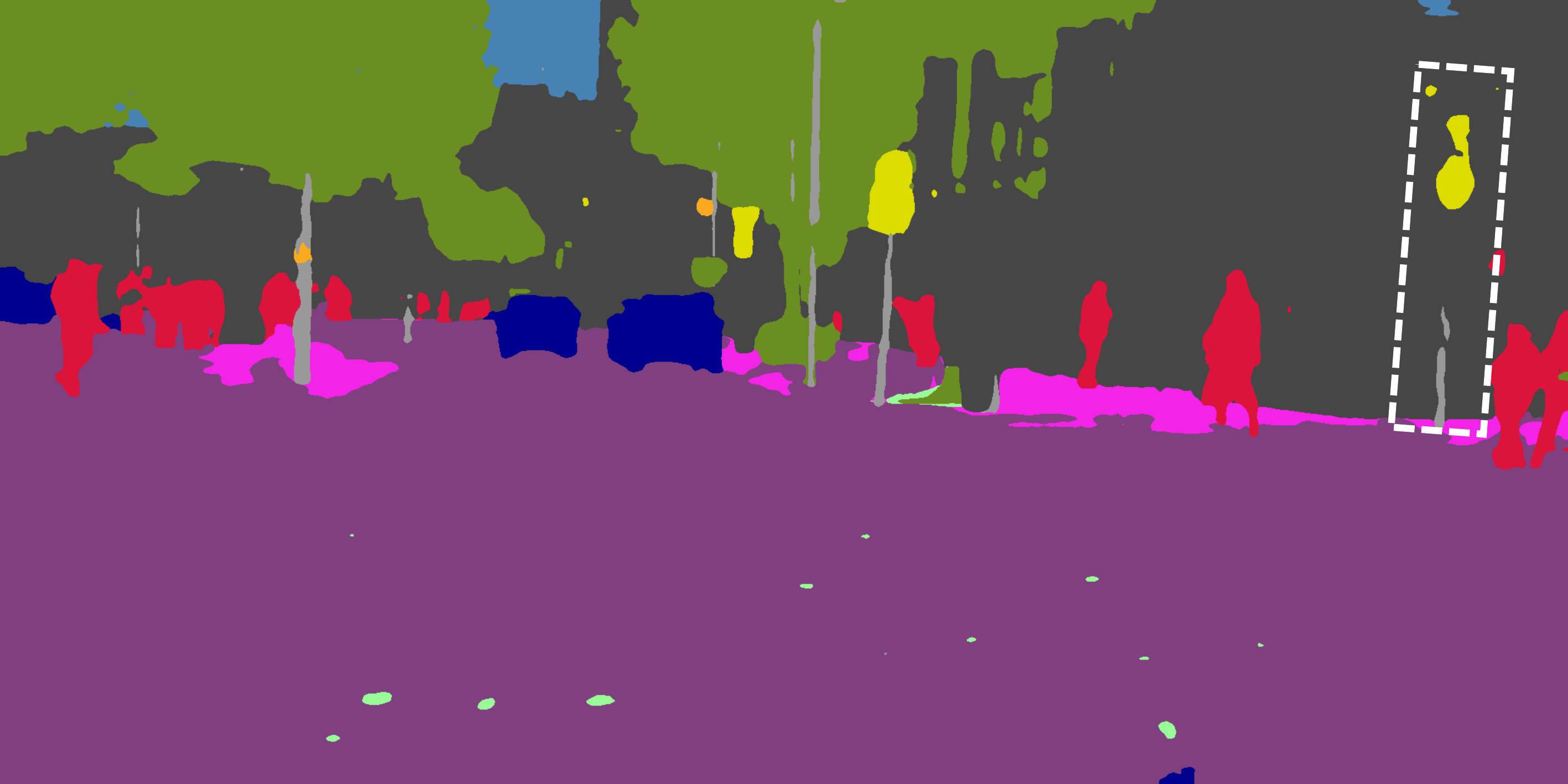}} \hfill
        \subfloat[Majority Voting]{\includegraphics[width=0.33\linewidth]{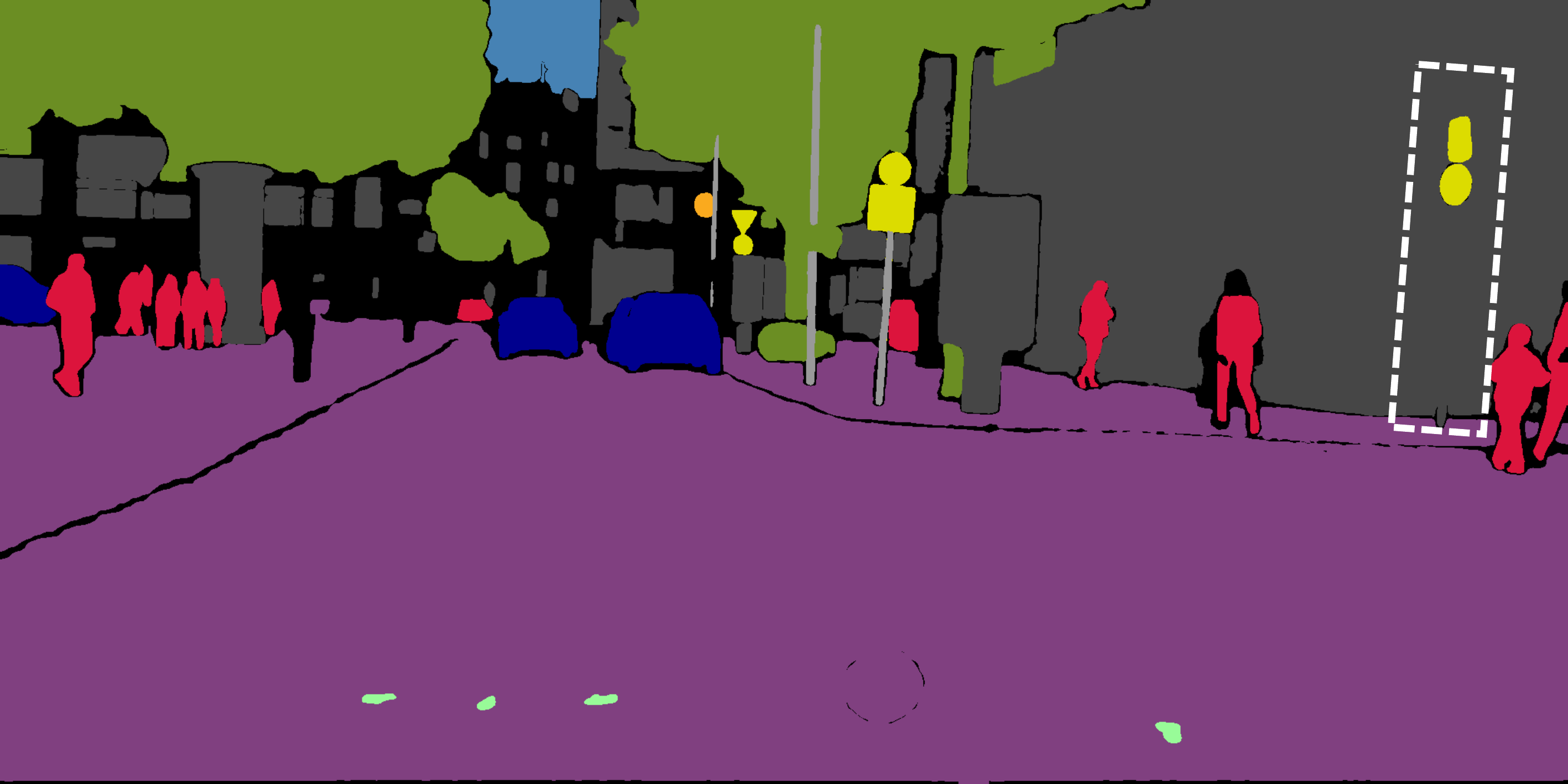}} \hfill
        \subfloat[SGML]{\includegraphics[width=0.33\linewidth]{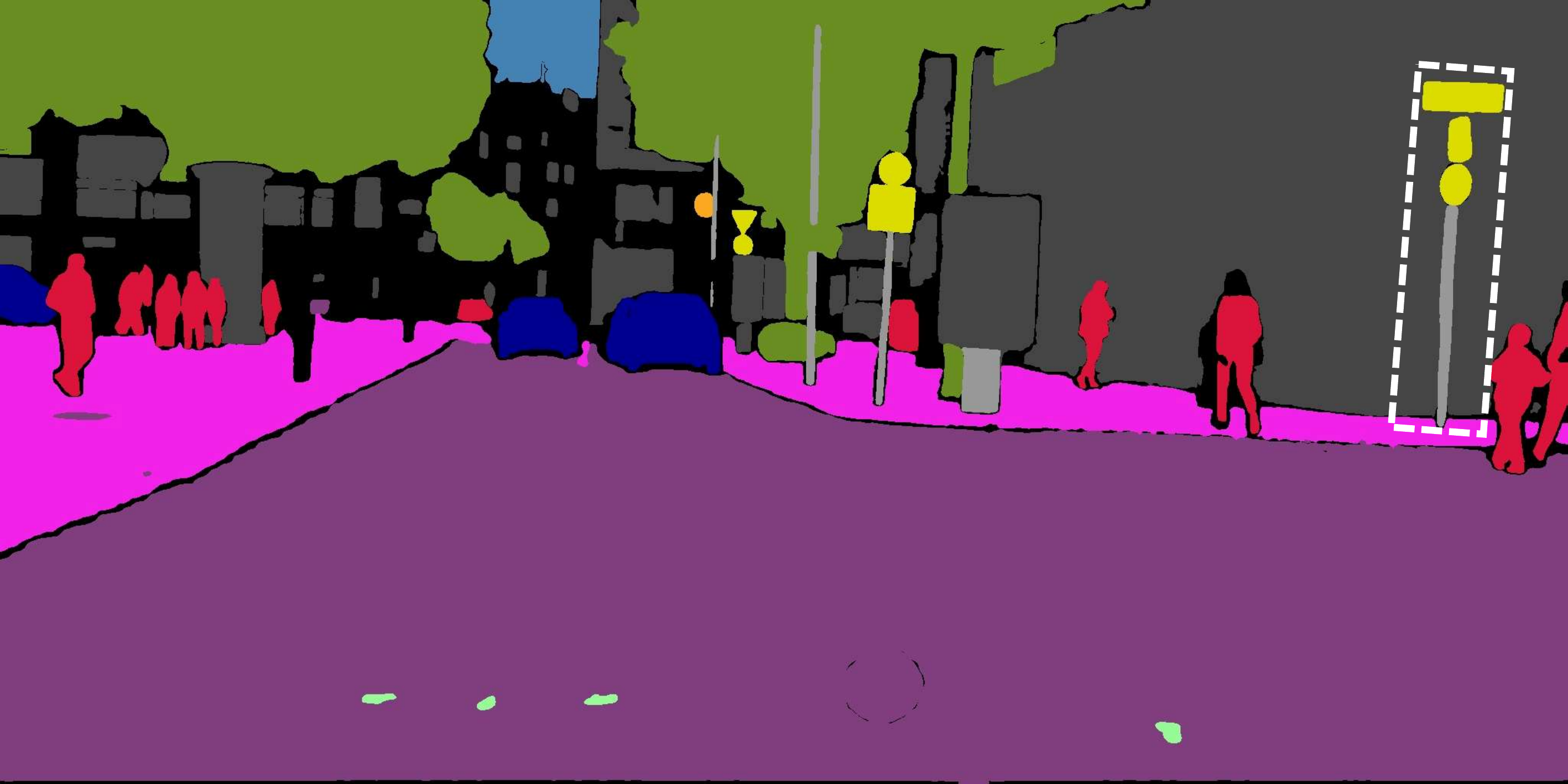}}
        \caption{The demonstration of Majority Voting and SGML.
        }
        \vspace{-0.2cm}
        \label{fig:sgml}
\end{figure}

As depicted in Fig.~\ref{fig:sgml} (b), SAM masks lack specific semantic labels, rendering them unsuitable for direct utilization in semantic segmentation. One conventional approach to obtain these labels is Majority Voting\cite{ssm}. This method involves identifying the corresponding positions of each mask in $\hat{y}_{uda}$ and then determining the most frequently occurring semantic class, thus assigning it to the mask. 

However, an inherent limitation of Majority Voting is its tendency to assign rare or small objects to prevalent or larger categories. Since the unbalanced data distribution in the target domain, $\hat{y}_{uda}$ often exhibits inaccuracies in segmenting rare or small objects like poles, traffic lights, and walls, leading to under- or mis-segmentation, as shown in Fig.~\ref{fig:sgml} (d). Despite SAM's precise instance-level segmentation of such objects, Majority Voting persists in assigning them to more frequent or larger categories, like buildings and vegetation, owing to their greater pixel count. This problem is evident in Fig.\ref{fig:sgml} (e), where the small pole is erroneously labeled as a building by Majority Voting, and the sidewalk is overwhelmed by the road.

To mitigate the above issue, we propose the SGML, introduced in Algorithm~\ref{alg:sgml} and visualized in Fig~\ref{fig:SAM4UDASS}. Specifically, we leverage area ratio information of classes to guide the labeling process. Remarkably, despite the domain discrepancy between source and target domains, size relationships between distinct categories remain consistent. To elaborate, the traffic light's area is consistently smaller than the building's in both domains. To implement this, we compute the average area for each semantic class using the ground truth $y_s$ from the source domain, subsequently identifying small-area ($C_s$) and large-area ($C_l$) classes, denoted as $[class_{s_1}, \ldots, class_{s_m}]$ and $[class_{l_1}, \ldots, class_{l_n}]$, respectively, where $m+n \le C$ (the category count). Moreover, rare classes in the target domain are appended to $C_s$, determinable from $\hat{y}_{uda}$.  Introducing an area ratio $\alpha$, we opt for the second-largest area class in $C_s$ to assign the unlabeled mask. In addition, we incorporate a road assumption into SGML to mitigate misclassification between road and sidewalk based on semantic spatial distribution. Specifically, the SAM mask, whose area ranks among the top three, with the center in the lower part of the image's central region and classified as the sidewalk, is corrected to be the road. The masks are processed in descending order of area to ensure that large objects do not overlap small ones. 

\begin{algorithm}[htbp]
        \caption{SGML}\label{alg:sgml}
        \KwIn{%
            SAM masks: $masks$, UDA pseudo-label: $\hat{y}_{uda}$,
            small and large classes: $C_s$, $C_l$, \
            area ratio: $\alpha$}
        \KwOut{SAM pseudo-label: $\hat{y}_{sam}$}
        
        Sort $masks$ in descending order based on their areas\;
        Initialize $\hat{y}_{sam}$ as an empty array of size same as $\hat{y}_{uda}$\;
        
        \For{mask in masks}{
            ids = $ \hat{y}_{uda}[mask]$\;
            ids, counts = unique\_counts(ids)\;
            ids, counts = sort\_ids\_by\_counts(ids, counts)\;
            
            \eIf{ids[0] $\in C_l$ $\land$ ids[1] $\in C_s$ $\land$ $\frac{counts[1]}{counts[0]} > \alpha$}{
                class\_id = ids[1]\;
            }{
                class\_id = ids[0]\;
            }
            \If{index(mask) $\le$ 3 $\land$ class\_id = sidewalk\_id $\land$ mask.center $\in$ lower\_central\_region}{
                class\_id = road\_id\;}
            $\hat{y}_{sam}[mask]$ = class\_id\;
        }
        \Return{$\hat{y}_{sam}$}\;
\end{algorithm}


\subsection{Fusion Strategies}

\subsubsection{Fusion Strategy 1}
This approach prioritizes $\hat{y}_{sam}$ with the intent of capitalizing on SAM's ability to deliver precise boundary segmentation. It takes $\hat{y}_{sam}$ as the baseline and populates the vacant areas with information from $\hat{y}_{uda}$.

\begin{equation}
        \begin{aligned}
                \hat{y}_{t}^{(1)} &= \hat{y}_{sam} \\
                \hat{y}_{t}^{(1)}[\hat{y}_{sam} = void] &= \hat{y}_{uda}[\hat{y}_{sam} = void]    
        \end{aligned}
        \label{eq:fusion1}
\end{equation}

\subsubsection{Fusion Strategy 2}
This strategy assumes that $\hat{y}_{uda}$ should take precedence, maximizing the use of semantic information from the target domain. Notably, SAM performs well on foreground and small objects that are typically under- or mis-segmented by $\hat{y}_{uda}$ (i.e., $C_s$). $\hat{y}_t^{(2)}$ is based on $\hat{y}_{uda}$, and the contents of $C_s$ from $\hat{y}_{sam}$ are employed for refinement: 

\begin{equation}
        \label{eq:fusion2}
        \begin{aligned}
                \hat{y}_t^{(2)} &= \hat{y}_{uda} \\
                \hat{y}_t^{(2)}[\hat{y}_{sam} \in C_s] &= \hat{y}_{sam}[\hat{y}_{sam} \in C_s]
        \end{aligned}
\end{equation}

\subsubsection{Fusion Strategy 3}
Building on Fusion Strategy 1, this approach further addresses the inconsistency in semantic granularity between SAM and the target domain. Algorithm~\ref{alg:fusion3} outlines Fusion Strategy 3, while the illustration of the three fusion strategies is available in Fig.~\ref{fig:SAM4UDASS}.

\begin{figure}[htbp]
        \captionsetup[subfloat]{font=scriptsize,labelfont=scriptsize}
        \centering
        \vspace{-0.5cm}
        \subfloat[Image with ground truth]{\includegraphics[width=0.33\linewidth]{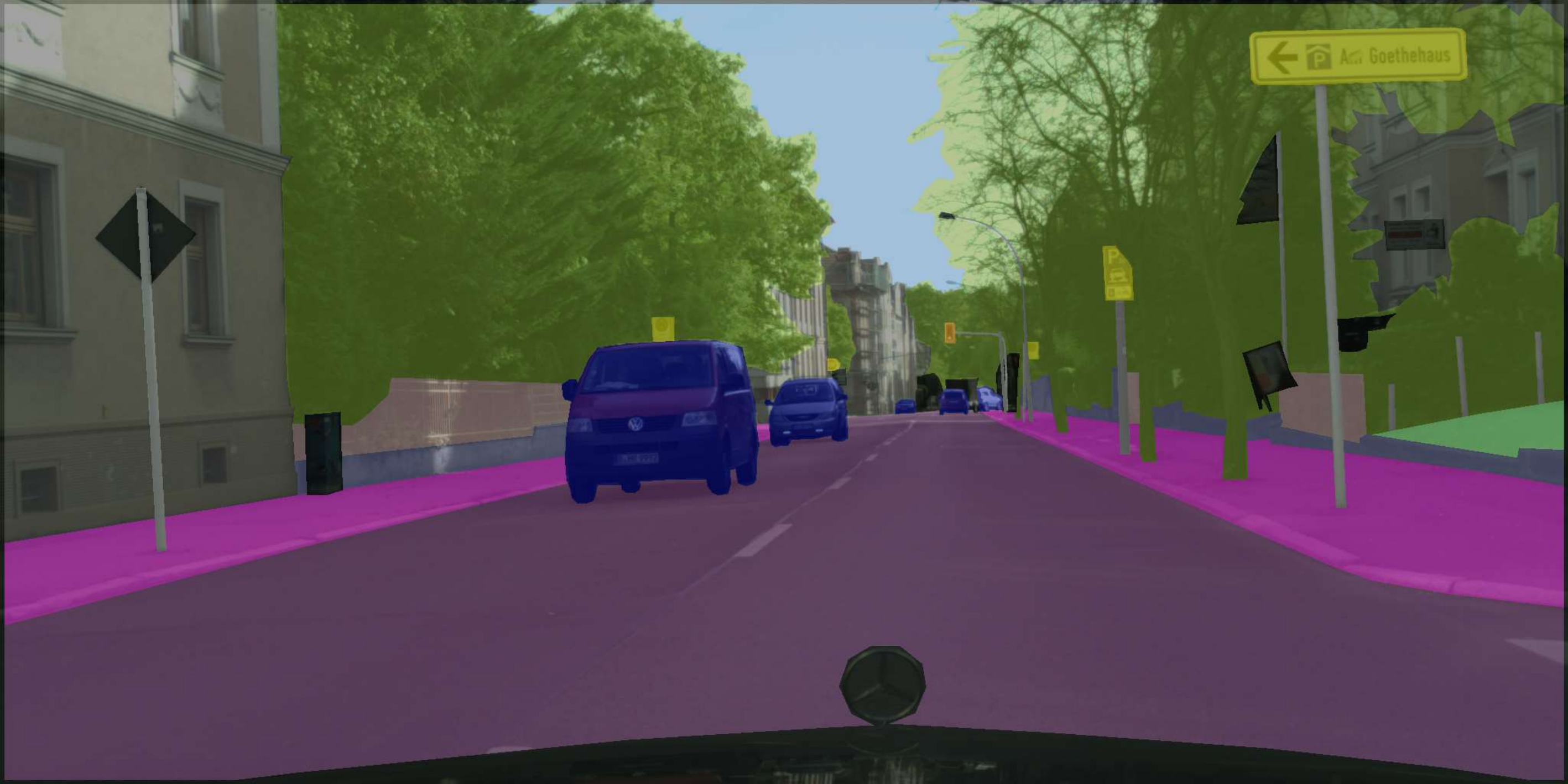}} \hfill
        \subfloat[SAM masks]{\includegraphics[width=0.33\linewidth]{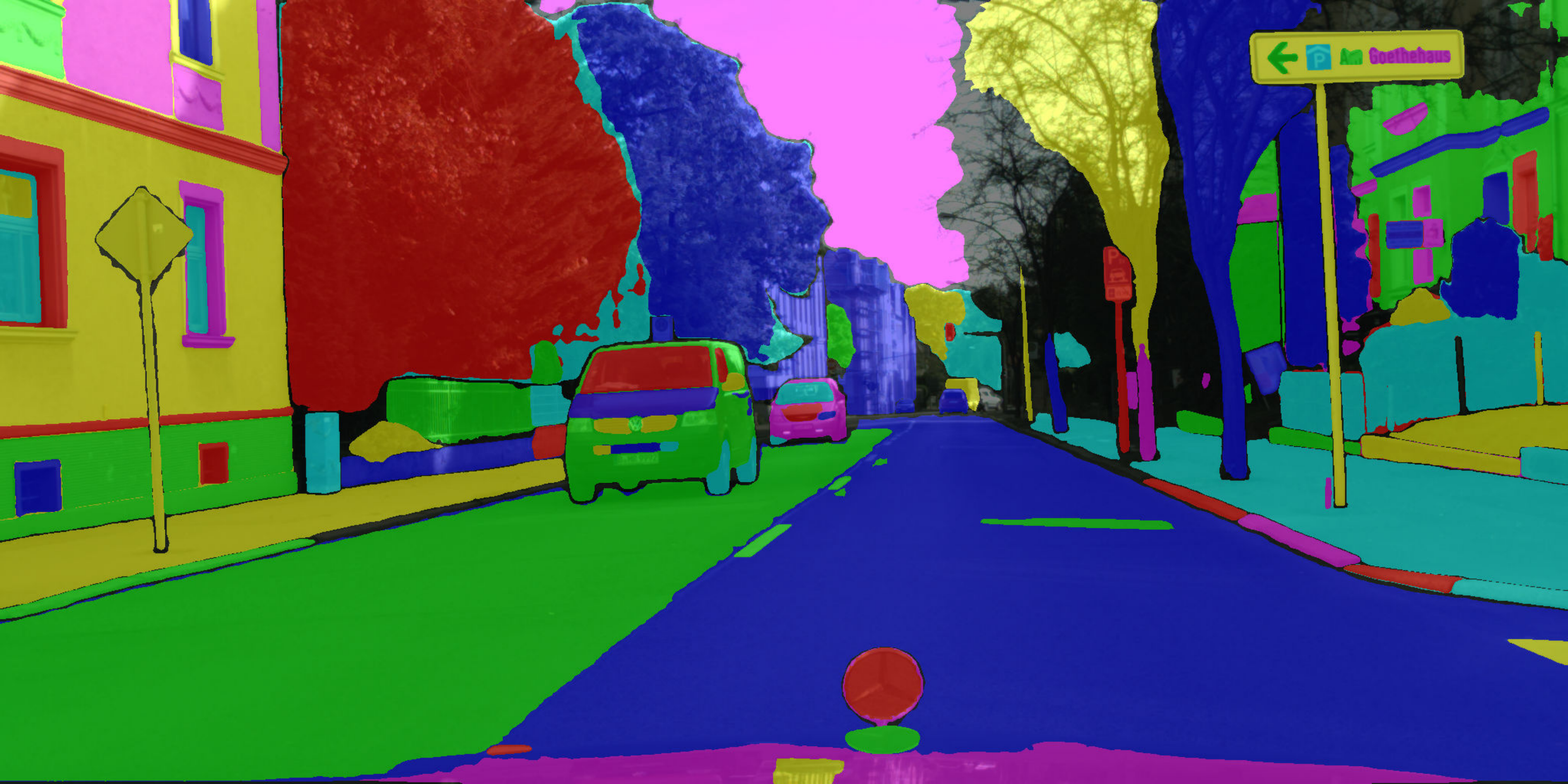}} \hfill
        \subfloat[UDA pseudo-label]{\includegraphics[width=0.33\linewidth]{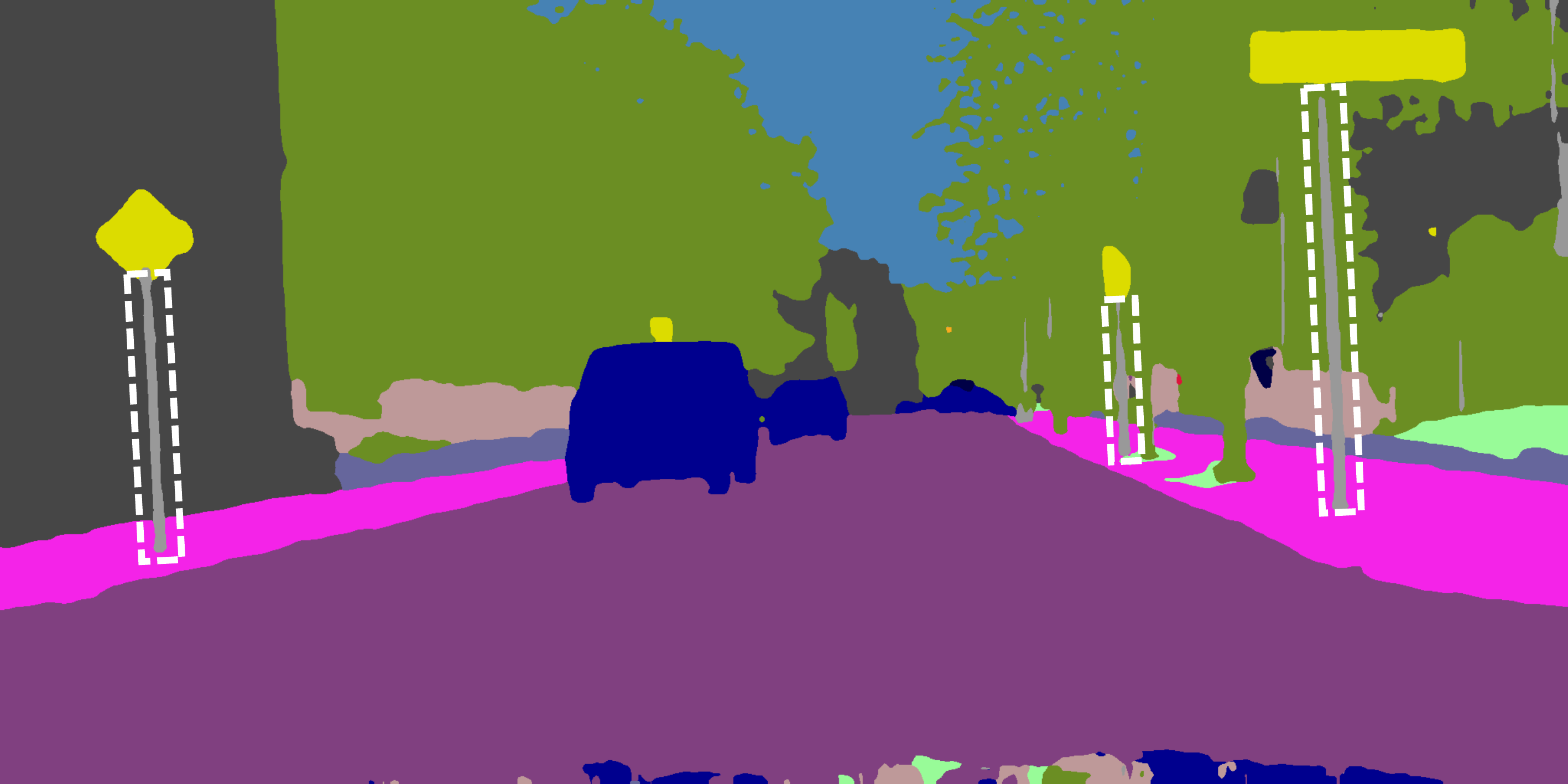}}\\ \vspace{-0.30cm}
        \subfloat[SGML]{\includegraphics[width=0.33\linewidth]{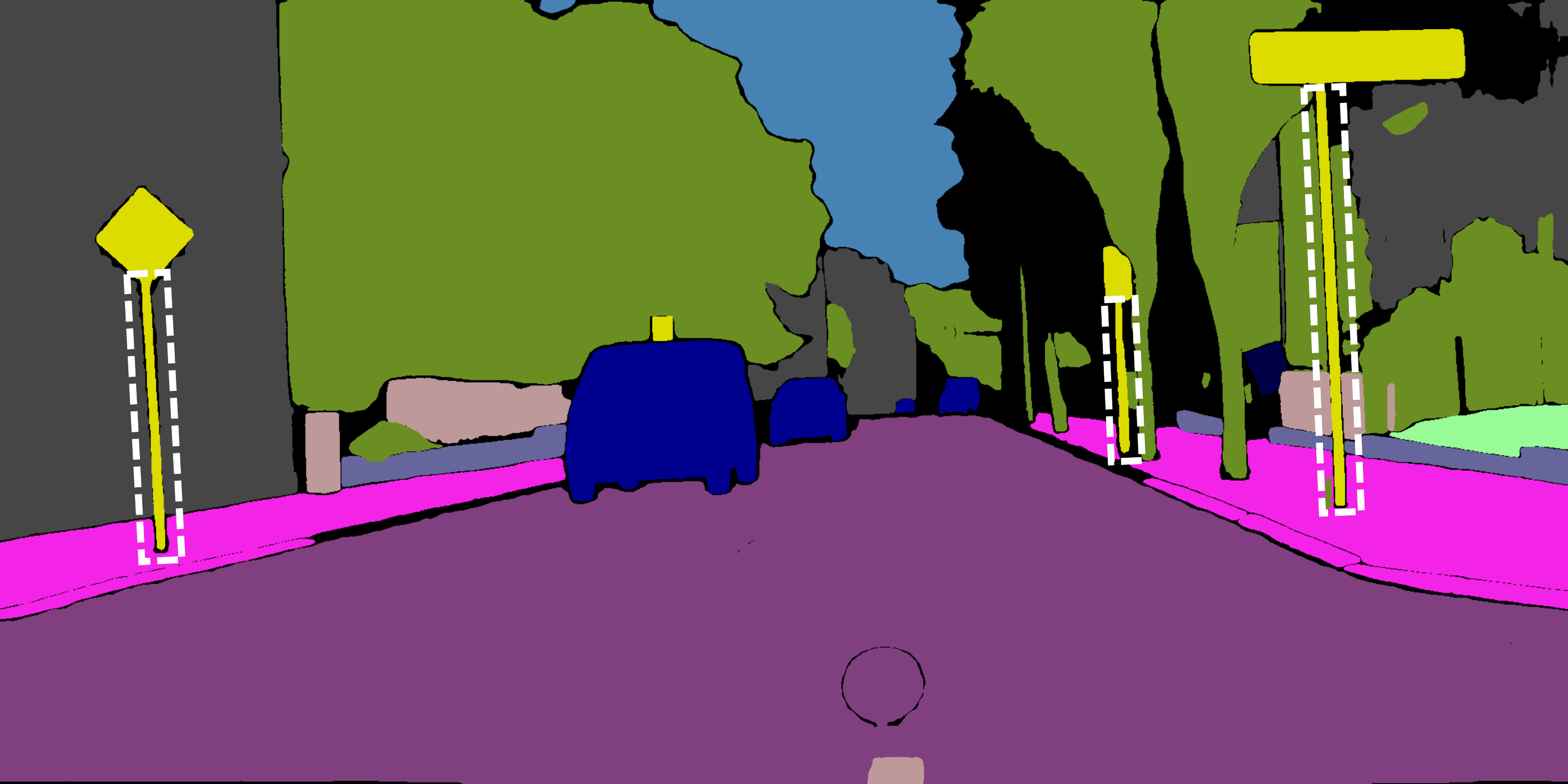}} \hfill
        \subfloat[Fusion Strategy 1]{\includegraphics[width=0.33\linewidth]{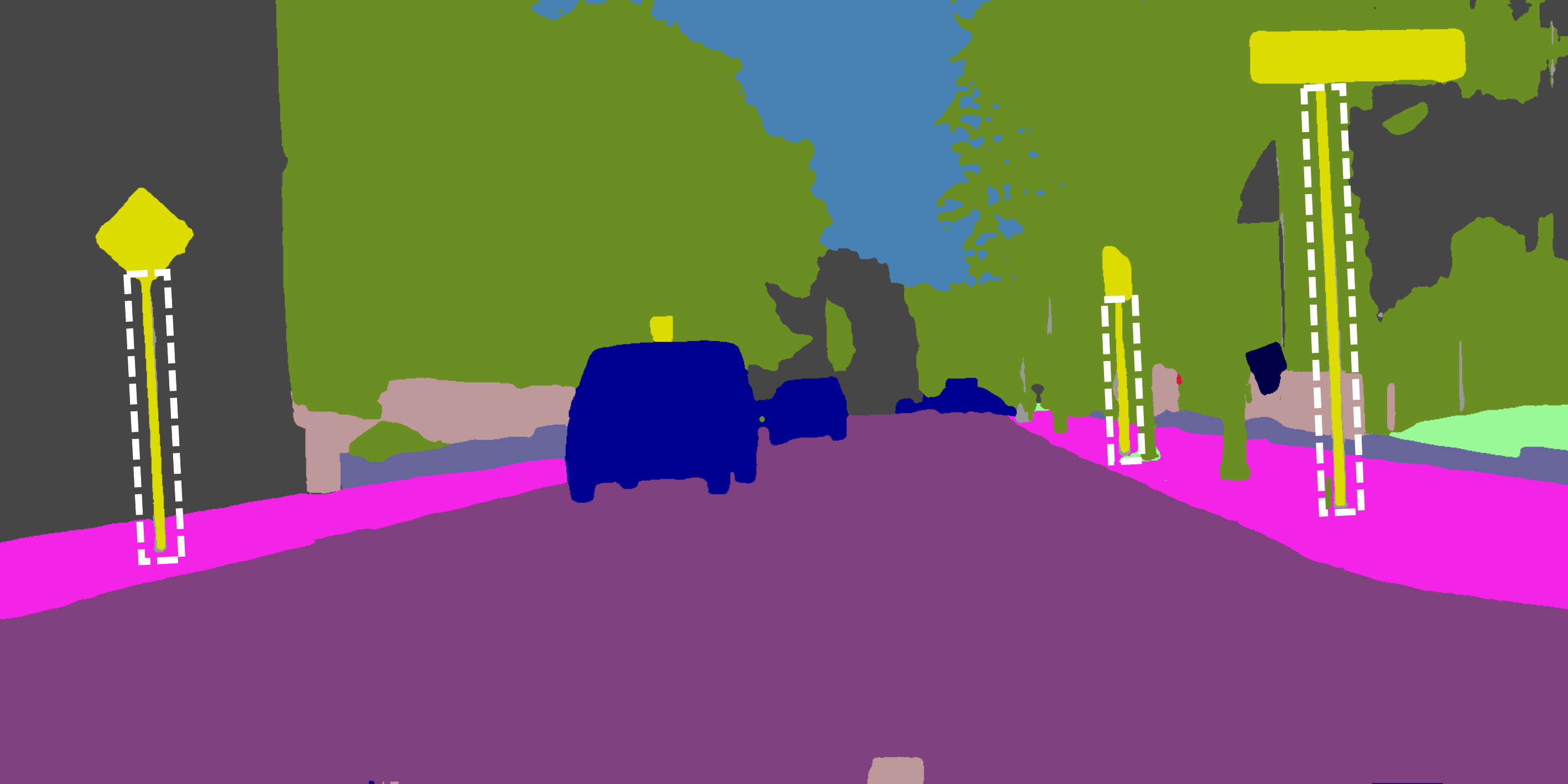}} \hfill
        \subfloat[Fusion Strategy 3]{\includegraphics[width=0.33\linewidth]{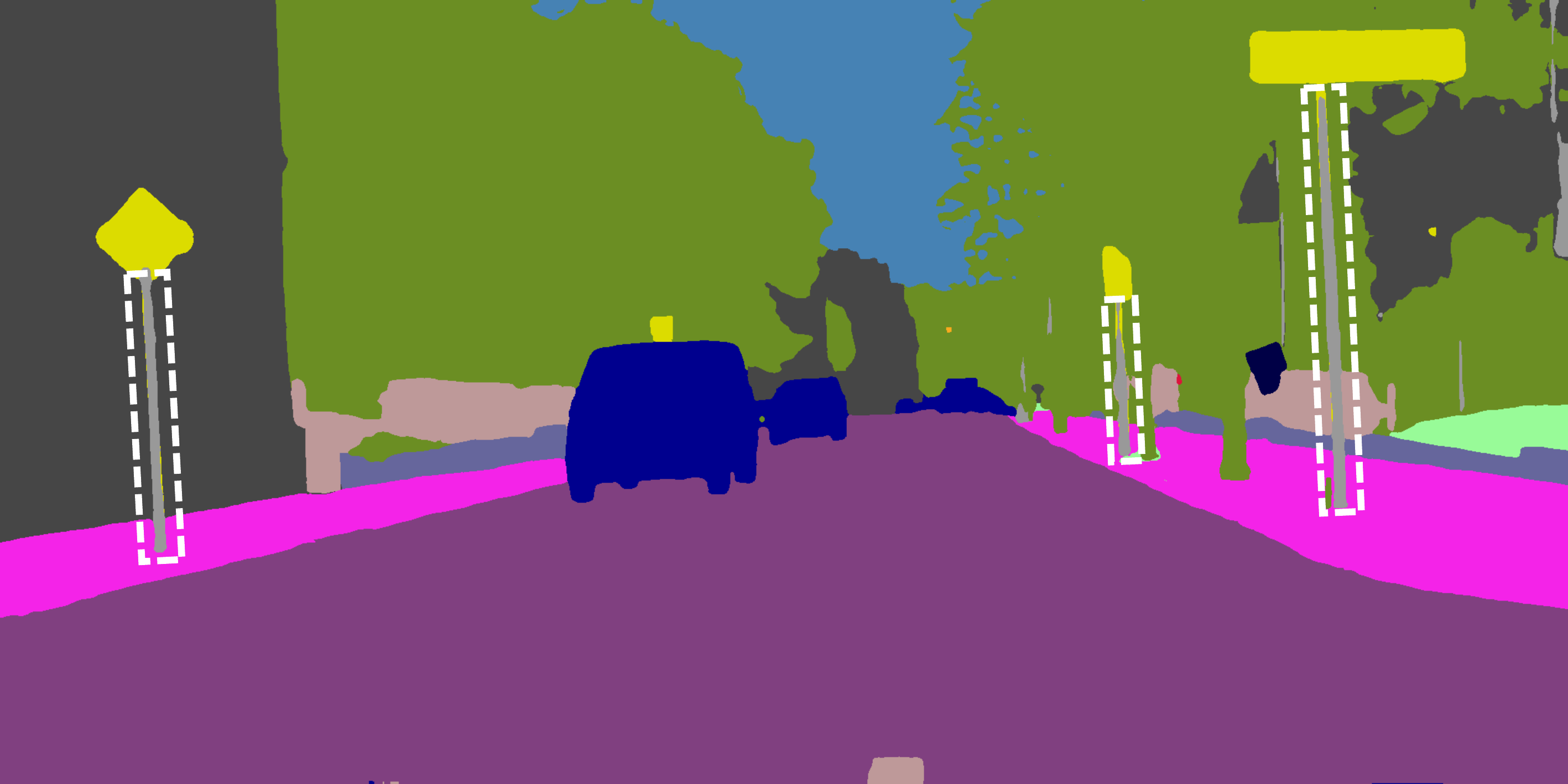}}
        \caption{The demonstration of Fusion Strategy 1 and 3.
        }
        \vspace{-0.2cm}
        \label{fig:fusion3}
\end{figure}

One significant challenge when directly using $\hat{y}_{sam}$ lies in the potential mismatch between the semantic granularity levels of SAM-generated masks for whole-image segmentation and the semantic categories in the target domain. For instance, in widely-used driving scene segmentation datasets like Cityscapes, which has 19 semantic classes, road and sidewalk represent separate classes (finer granularity), while cars are not further divided into components like wheels or windows (coarser granularity). If SAM exhibit finer granularity segmentation, Majority Voting and SGML can still correctly identify both wheels and windows as part of a car, provided the semantic information from $\hat{y}_{uda}$ is correct. However, if SAM's segmentation granularity is coarser, distinct categories could be grouped into a single category, thus leading to lower quality pseudo-labels. Fig.\ref{fig:fusion3} (b) and the first row of Fig.\ref{fig:get_sam} exemplify this, where the pole and traffic sign are segmented into a single mask by SAM, and either Majority Voting or SGML would assign the label of the traffic sign to the entire mask.

To address this concern, we introduce the confidence score of the target image $\Theta_t=\max(p_t)$ into the fusion process. We define a category similarity matrix $S$ as follows:

\begin{equation}
        \begin{aligned}
                S &= \begin{bmatrix}
                        s_{11} & s_{12} & \ldots & s_{1C} \\
                        s_{21} & s_{22} & \ldots & s_{2C} \\
                        \vdots & \vdots & \ddots & \vdots \\
                        s_{C1} & s_{C2} & \ldots & s_{CC} \\
                    \end{bmatrix} \\
        \end{aligned}\\
        \label{eq:similarity_matrix}
\end{equation}
where $s_{ij}$ represents the ``similarity" between class $i$ and $j$. It is 1 if SAM may segment them as one mask and 0 otherwise.

Denote the fusion result of Fusion Strategy 3 as $\hat{y}_t^{(3)}$, and initialize it with $\hat{y}_t^{(1)}$. For each $class_i$ in $\hat{y}_{uda}$, we identify its similar classes as $\{class_j, s_{ij} = 1\}$. Considering the potential inconsistency in semantic granularity between SAM masks and the target domain, we define pixels that may be misclassified into similar classes as $mask_i$: 
\begin{equation}
        mask_i = [\hat{y}_{uda}=class_i] \odot  [\hat{y}_t^{(1)}=class_j, s_{ij}=1]
        \label{eq:class_i}
\end{equation}
where $\odot$ represents the logical AND at pixel-level. If the corresponding area in $\Theta_t$ is higher than the confidence threshold $\beta$, the $class_i$ is assigned to those pixels in $\hat{y}_t^{(3)}$.
\begin{equation}
        \hat{y}_t^{(3)}[\Theta_t[mask_i]>\beta] = class_i
        \label{eq:yt3}
\end{equation}

\begin{algorithm}[h]
        \caption{Fusion Strategy 3}\label{alg:fusion3}
        \KwIn{Fusion Strategy 1 result: $\hat{y}_{t}^{(1)}$,
                category similarity matrix: $S$,
                UDA pseudo-label: $\hat{y}_{uda}$,
                confidence score and threshold: $\Theta_t$, $\beta$,
                }
        \KwOut{Fusion Strategy 3 result: $\hat{y}_{t}^{(3)}$}
        
        Initialize $\hat{y}_{t}^{(3)}$ as $\hat{y}_{t}^{(1)}$\;
        
        Get unique class IDs from $\hat{y}_{uda}$: $ids = \text{unique}(\hat{y}_{uda})$\;
        
        \For{i in ids}{
            $mask_i = [\hat{y}_{uda}=class_i] \odot [\hat{y}_t^{(1)}=class_j, s_{ij}=1]$\;
            $\hat{y}_t^{(3)}[\Theta_t[mask_i]>\beta] = class_i$;
            
        }
        \Return{$\hat{y}_{t}^{(3)}$}\;
\end{algorithm}


\section{Experiments}
\subsection{Datasets}
We conduct comprehensive experiments across two distinct scenarios: synthetic-to-real and normal-to-adverse domain adaptation, employing the GTA5-to-Cityscapes, SYNTHIA-to-Cityscapes, and Cityscapes-to-ACDC datasets to evaluate the effectiveness of SAM4UDASS. The synthetic-to-real case involves GTA5 and SYNTHIA as source domains, while Cityscapes serves as the target domain. Conversely, in the normal-to-adverse scenario, Cityscapes is the source domain, and ACDC is the target domain.

GTA5 encompasses 24,966 annotated images, sourced from the video game Grand Theft Auto V, with a resolution of 1914 $\times$ 1052. SYNTHIA offers 9400 photo-realistic frames of size 1280 $\times$ 760, generated from a virtual city environment. Notably, both GTA5 and SYNTHIA's label classes match Cityscapes, comprising 19 and 16 categories, respectively. Cityscapes comprises 2975, 500, and 1525 precisely annotated images for training, validation, and testing, respectively, taken from diverse urban street scenes across 50 different cities. ACDC, designed to foster semantic understanding in challenging visual conditions, comprises 1000 foggy, 1006 nighttime, 1000 rainy, and 1000 snowy images, each meticulously annotated at a resolution of 1920 $\times$ 1080.

\subsection{Self-training UDA Methods}
We establish baselines using advanced self-training UDA methods DAFormer\cite{daformer} and MIC\cite{mic}. DAFormer is adopted in most of our experiments for its faster training speed, while MIC is employed to achieve higher metrics. Additionally, to demonstrate the plug-and-paly capability of SAM4UDASS, we also conduct experiments using the SePiCo\cite{sepico} based on DeeplabV2\cite{deeplabv2} and TUFL\cite{TUFL} based on BiSeNet\cite{bisenet}.

\subsection{Evaluation Methods and Implementation Details}
The widely used Interaction-over-Union (IoU) and mean IoU (mIoU) are used as the evaluation metrics:
\begin{equation}
        \begin{aligned}
                \textrm{IoU} &= \frac{\textrm{TP}}{\textrm{TP}+\textrm{FP}+\textrm{FN}} \\
                \textrm{mIoU} &= \frac{1}{C} \sum_{c=1}^{C} \textrm{IoU}_c
        \end{aligned}
        \label{eq:iou}
\end{equation}
where $\textrm{TP}$, $\textrm{FP}$, and $\textrm{FN}$ represent the number of true positive, false positive, and false negative pixels, respectively.

For SAM, sample points per side, prediction iou threshold, and stability score threshold are 32, 0.86, and 0.92, respectively. For SGML, the small and large classes ($C_s$ and $C_l$) are \{wall, fence, traffic sign, traffic light, pole, bike\} and \{building, vegetation, sidewalk, road\}. The area ratio $\alpha$ is configured to 0.2. The confidence threshold $\beta$ is 0.9 for MIC and 0.99 for DAFormer in Fusion Strategy 3. The ``similar'' classes with $s_{ij}=1$ are shown in Table~\ref{tab:sim}.
\begin{table}
        \centering
        \caption{Similar classes for Fusion Strategy 3.}
        \label{tab:sim}
        \begin{tabular}{c|c}
                \hline
                $\textrm{class}_i$ & $\textrm{class}_j$ \\ \hline
                road & sidewalk \\
                fence, pole, traffic light, traffic sign & building \\
                fence, traffic light, traffic sign, terrain & vegetation \\
                traffic light, traffic sign & pole \\
                \hline
        \end{tabular}
        \vspace{-0.2cm}
\end{table}
These settings are based on the target domain and should be adjusted according to the specific driving scenes.

Most of training settings are following the self-training methods\cite{daformer,mic,sepico,TUFL}. For DAFormer and MIC, the AdamW optimizer is utilized, with learning rates of $6\times 10^{-5}$ for the encoder and $6\times10^{-4}$ for the decoder. The crop size is set at 512 $\times$ 512 for DAFormer and 1024 $\times$ 1024 for MIC, with a batch size of 2 and 40,000 training iterations. For SePiCo, the images are cropped to 640 $\times$ 640 for training, and the iterations are also 40,000. AdamW with betas (0.9, 0.999) is used as the optimizer, with a weight decay of 0.01. For TUFL, the SGD optimizer is used with a learning rate of 0.5$\times 10^{-5}$, training for 20,000 iterations. Two 3090 GPUs are used for the experiments, and the CPU is AMD EPYC 7542. For more specific experimental settings, our code will be available at \url{https://github.com/ywher/SAM4UDASS}.

\subsection{Experimental Results}
\subsubsection{GTA5-to-Cityscapes}

\begin{table*}[!ht]
        \centering
        \setlength{\tabcolsep}{3pt}
        \renewcommand{\arraystretch}{1.1}
        \captionsetup{format=myformat}
	\caption{The adaptation performance and comparison on GTA5-to-Cityscapes} 
	\label{tab:gta5_comparison}
	\resizebox{\linewidth}{!}{
	\begin{tabular}{c|c|ccccccccccccccccccc|c}
		\hline
		Methods & Network & road & sw & build & wall & fence & pole & light & sign & vege & terrain & sky & person & rider & car & truck & bus & train & motor & bike & mIoU \\ \hline
		TUFL\cite{TUFL} & BiSeNet & 92.0  & 58.7  & 87.1  & 40.0  & 25.3  & 44.3  & 51.1  & 52.9  & 87.5  & 42.5  & 86.9  & 62.8  & 23.2  & 88.7  & 43.7  & 50.0  & 2.1  & 17.1  & 39.9 & 52.4 \\ 
		SePiCo\cite{sepico} & DeeplabV2 & 95.4  & 69.3  & 88.6  & 41.7  & 39.6  & 45.8  & 56.1  & 64.8  & 88.4  & 44.4  & 87.7  & 73.2  & 48.4  & 90.9  & 56.9  & 58.6  & 1.0  & 42.4  & 60.6  & 60.7 \\ \hline 
                TUFL+\textbf{SAM4UDASS} & BiSeNet & 92.5  & 60.3  & 87.5  & 41.6  & 29.3  & 46.8  & 52.5  & 58.0  & 87.8  & 41.7  & 88.2  & 63.9  & 21.7  & 89.4  & 45.7  & 49.0  & 2.3  & 22.8  & 50.6 & 54.3 \\ 
                SePiCo+\textbf{SAM4UDASS} & DeeplabV2 & 96.8  & 76.3  & 89.7  & 46.0  & 48.3  & 46.8  & 58.5  & 68.5  & 89.0  & 44.1  & 91.3  & 74.5  & 51.4  & 92.2  & 62.6  & 66.4  & 2.0  & 44.3  & 65.4 & 63.9 \\ \hline
                DAFormer\cite{daformer} & \multirow{5}{*}{\begin{tabular}{c} DAFormer \end{tabular}} & 94.6  & 64.5  & 89.3  & 52.0  & 44.8  & 48.1  & 56.1  & 59.2  & 89.9  & 50.4  & 91.5  & 71.9  & 45.0  & 92.5  & 75.5  & 80.6  & 74.3  & 55.1  & 62.0 & 68.3 \\
                HRDA\cite{hrda} & & 96.3  & 74.2  & 91.2  & \textbf{61.4} & 54.1  & 57.8  & 64.7  & 69.8  & 91.5  & \textbf{51.8} & 94.2  & 79.4  & 53.9  & 94.2  & 80.5  & 83.6  & 73.6  & 64.1  & 67.9 & 73.9 \\
                MIC\cite{mic} & & 97.0  & 78.2  & 91.6  & 59.9  & 57.3  & 60.5  & 64.5  & 71.3  & 91.3  & 49.2  & 93.9  & 79.9  & 56.0  & 94.5  & 82.7  & 91.0  & 81.1  & 64.2  & 69.0 & 75.4 \\ \cline{1-1} \cline{3-22}
                DAFormer+\textbf{SAM4UDASS} & & 95.2  & 68.4  & 90.5  & 57.2  & 50.7  & 51.2  & 59.5  & 62.2  & 90.8  & 51.3  & 92.6  & 73.5  & 47.8  & 93.5  & 80.3  & 86.1  & 77.6  & 58.7  & 67.9 & 71.3  \\
                MIC+\textbf{SAM4UDASS} & & \textbf{97.3} & \textbf{79.2} & \textbf{92.3} & 59.2  & \textbf{60.2} & \textbf{63.4} & \textbf{66.8} & \textbf{74.1} & \textbf{91.9} & 49.9  & \textbf{94.7} & \textbf{82.1} & \textbf{60.6} & \textbf{95.3} & \textbf{87.7} & \textbf{91.4} & \textbf{81.4} & \textbf{65.1} & \textbf{75.2} & \textbf{77.3} \\ \hline
	\end{tabular}}
\end{table*}

\begin{figure*}[!htbp]
	\centering
	\captionsetup[subfloat]{font=scriptsize,labelfont=scriptsize,labelformat=empty}
        \vspace{-0.5cm}
	\subfloat{\includegraphics[width=0.198\linewidth]{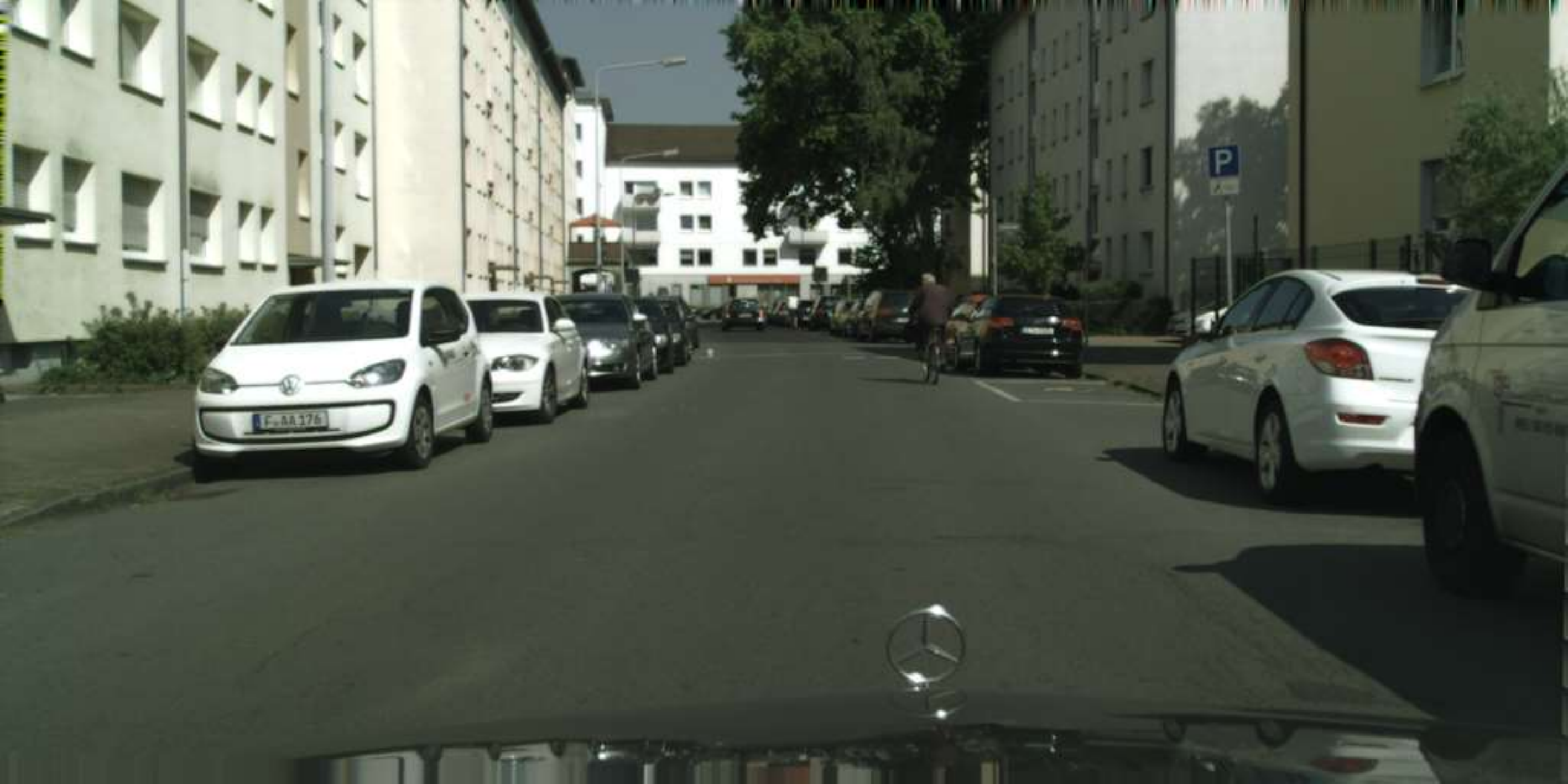}} \hfill
	\subfloat{\includegraphics[width=0.198\linewidth]{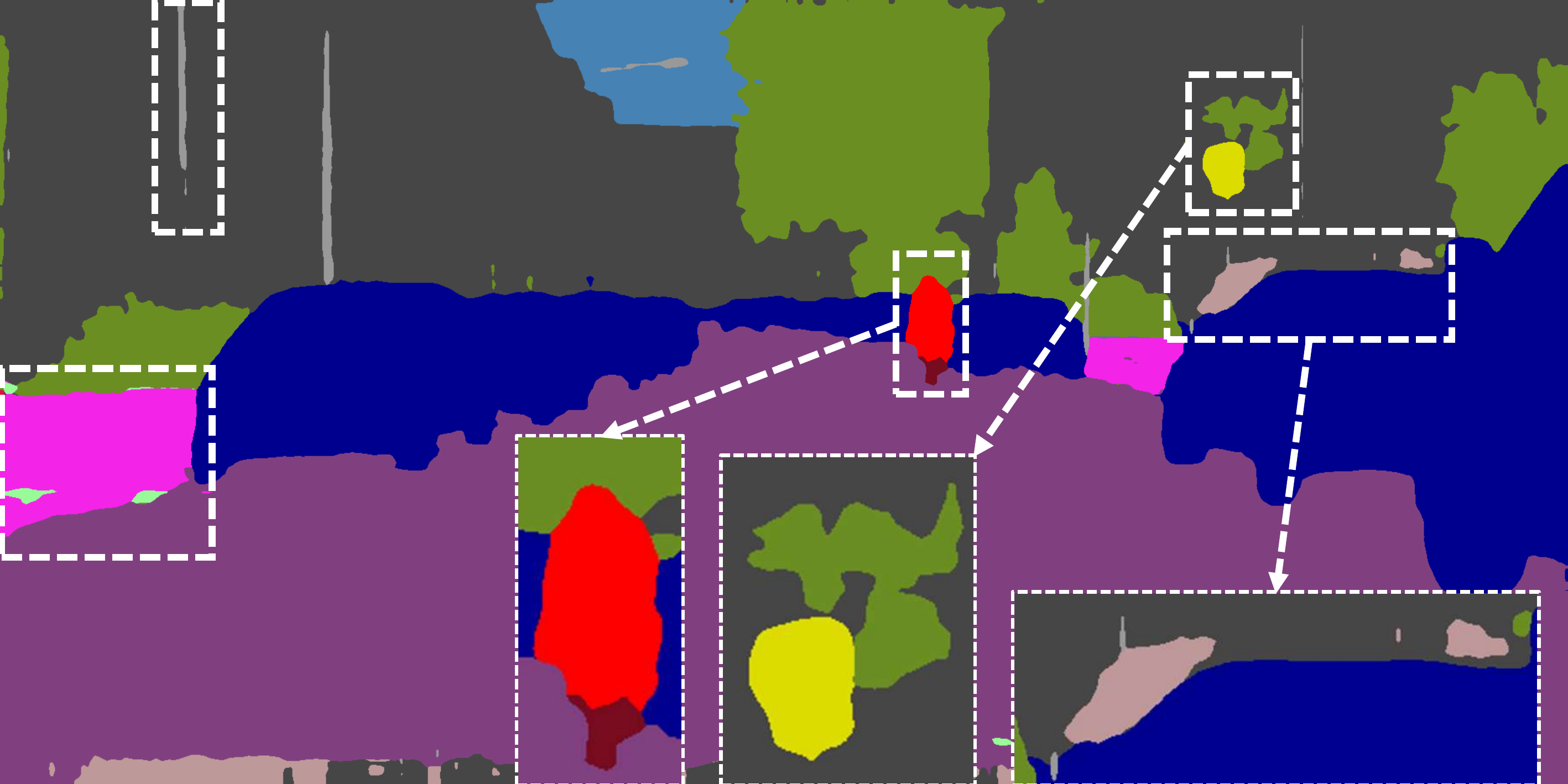}} \hfill
	\subfloat{\includegraphics[width=0.198\linewidth]{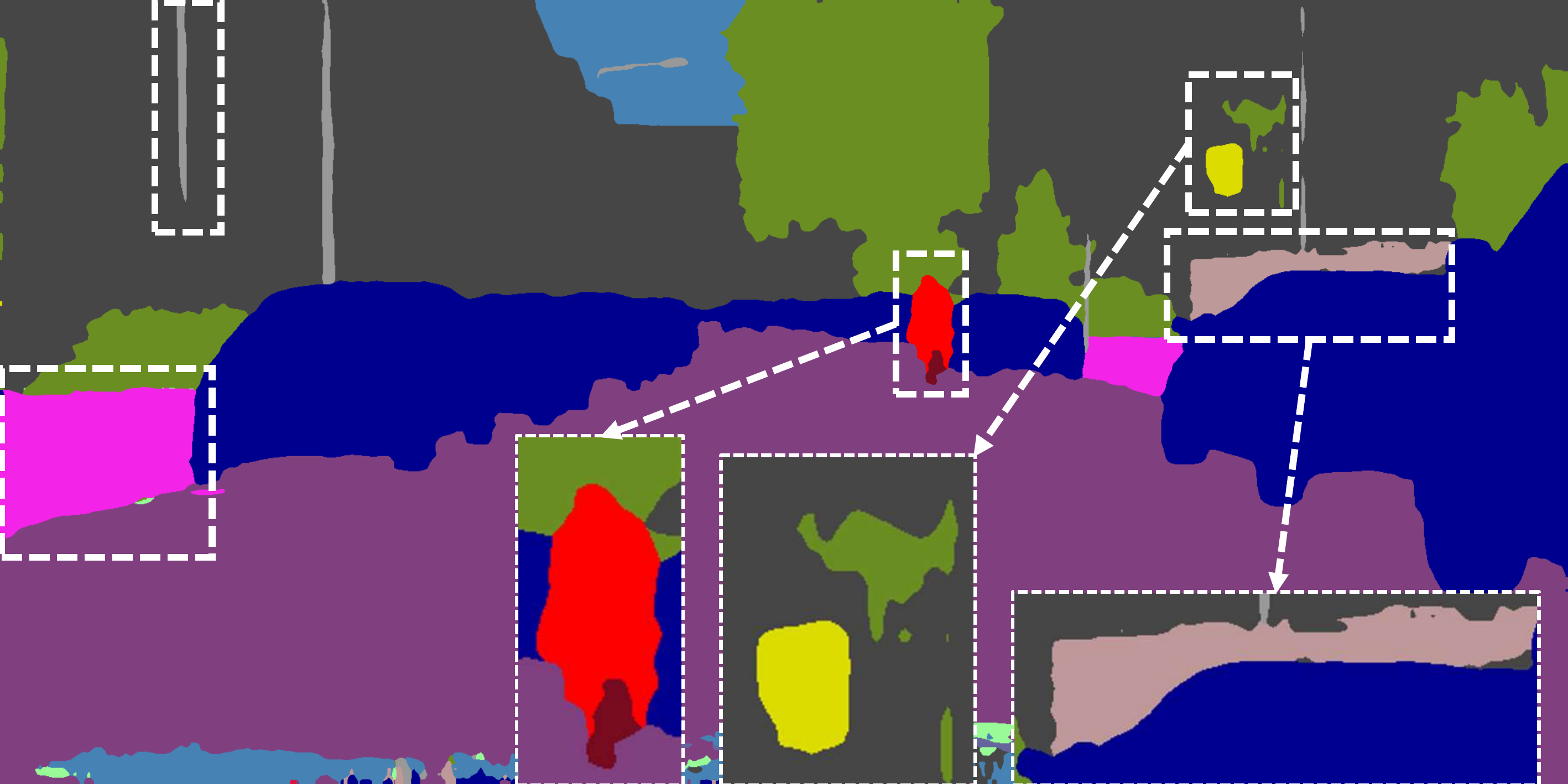}} \hfill
	\subfloat{\includegraphics[width=0.198\linewidth]{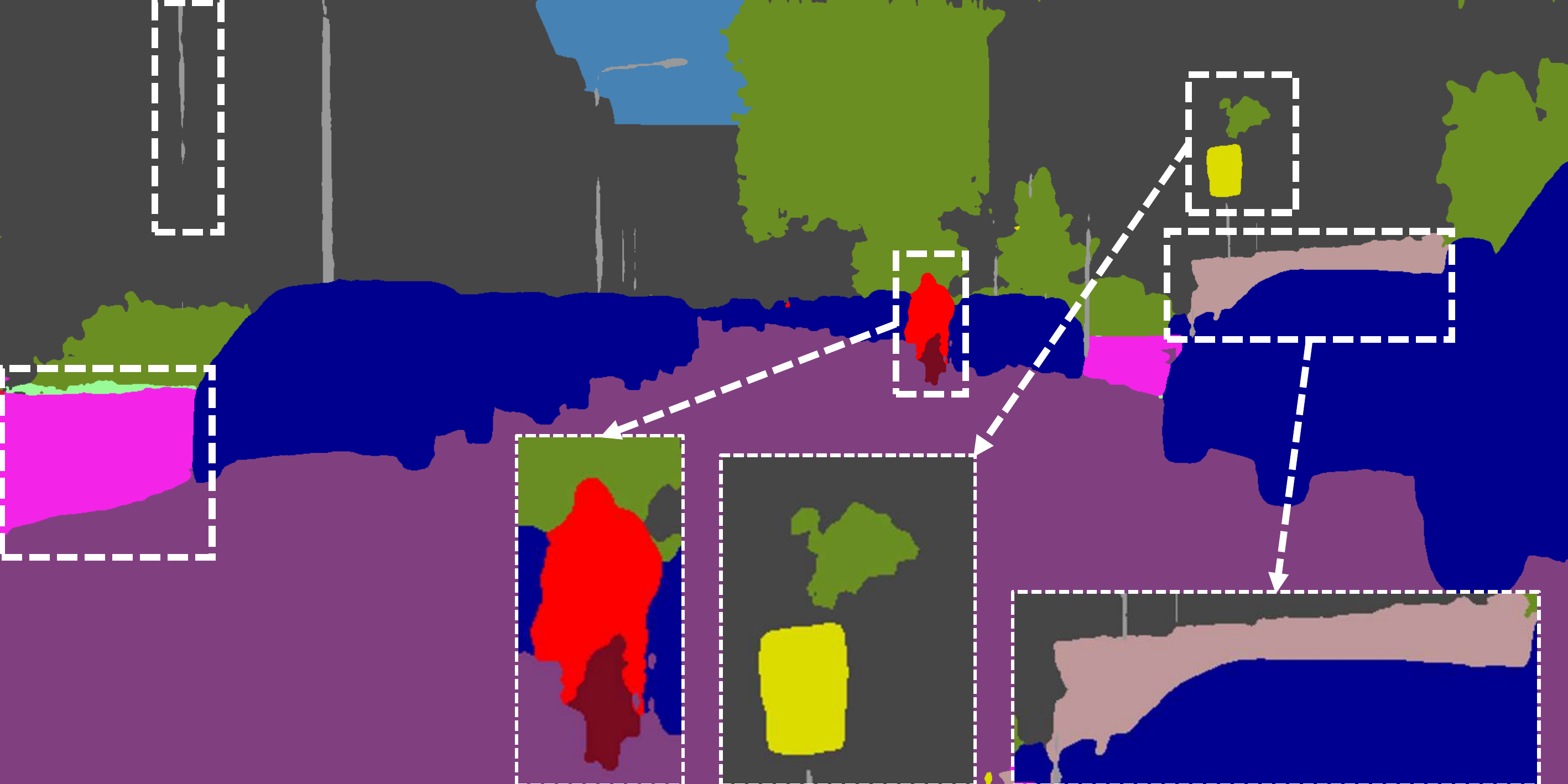}} \hfill
	\subfloat{\includegraphics[width=0.198\linewidth]{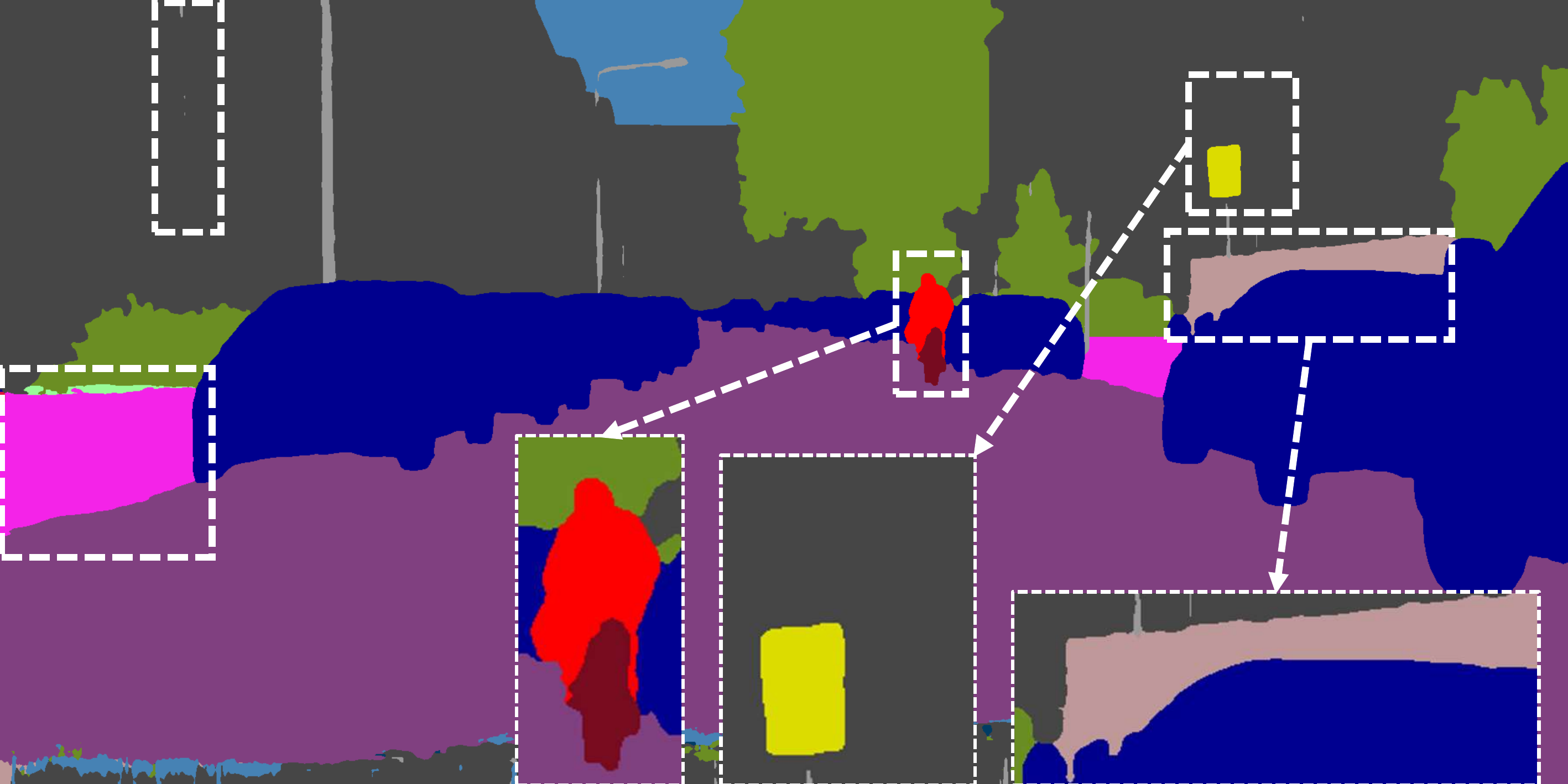}} \\ \vspace{-0.30cm}
	\subfloat{\includegraphics[width=0.198\linewidth]{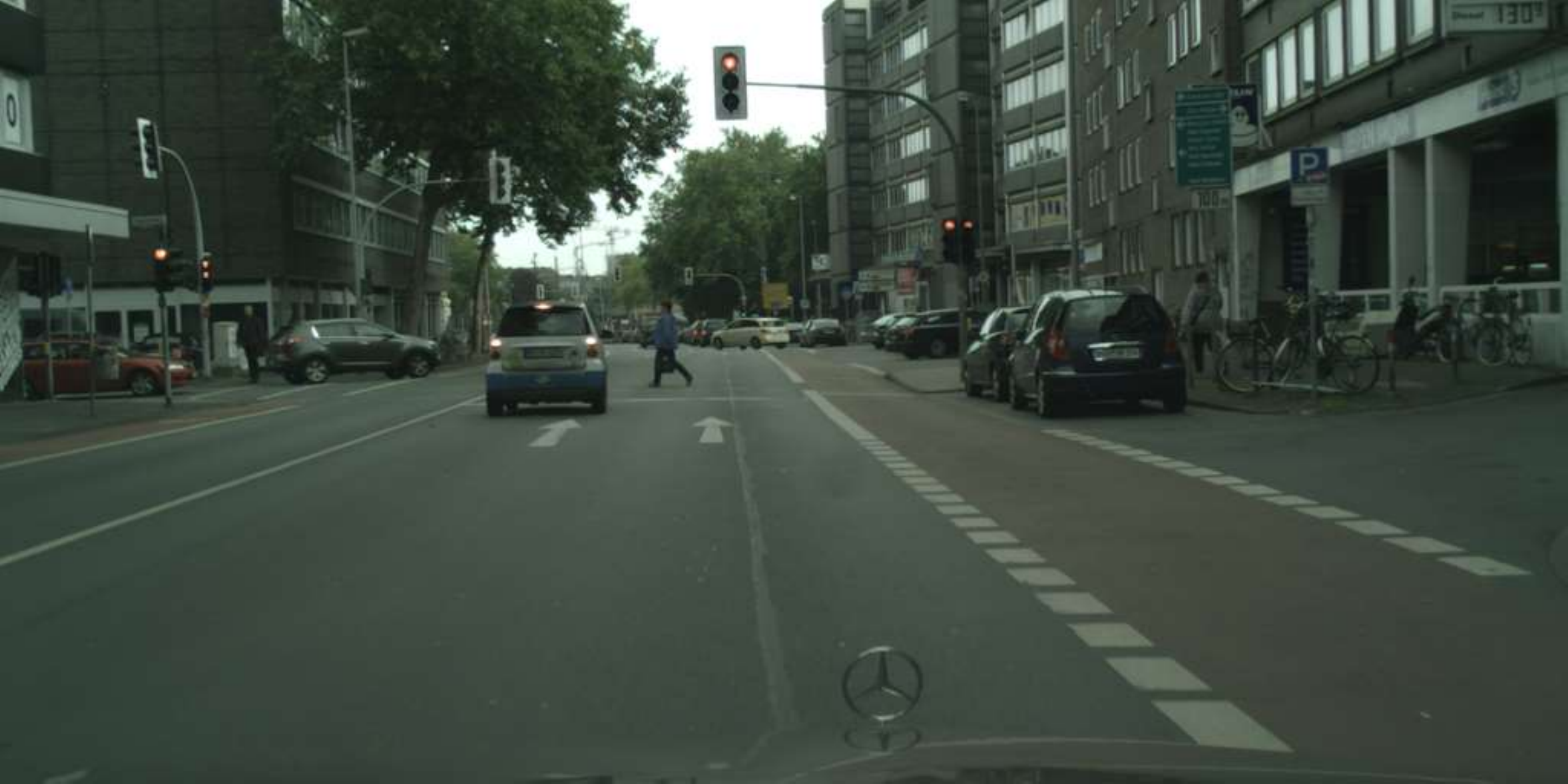}} \hfill
	\subfloat{\includegraphics[width=0.198\linewidth]{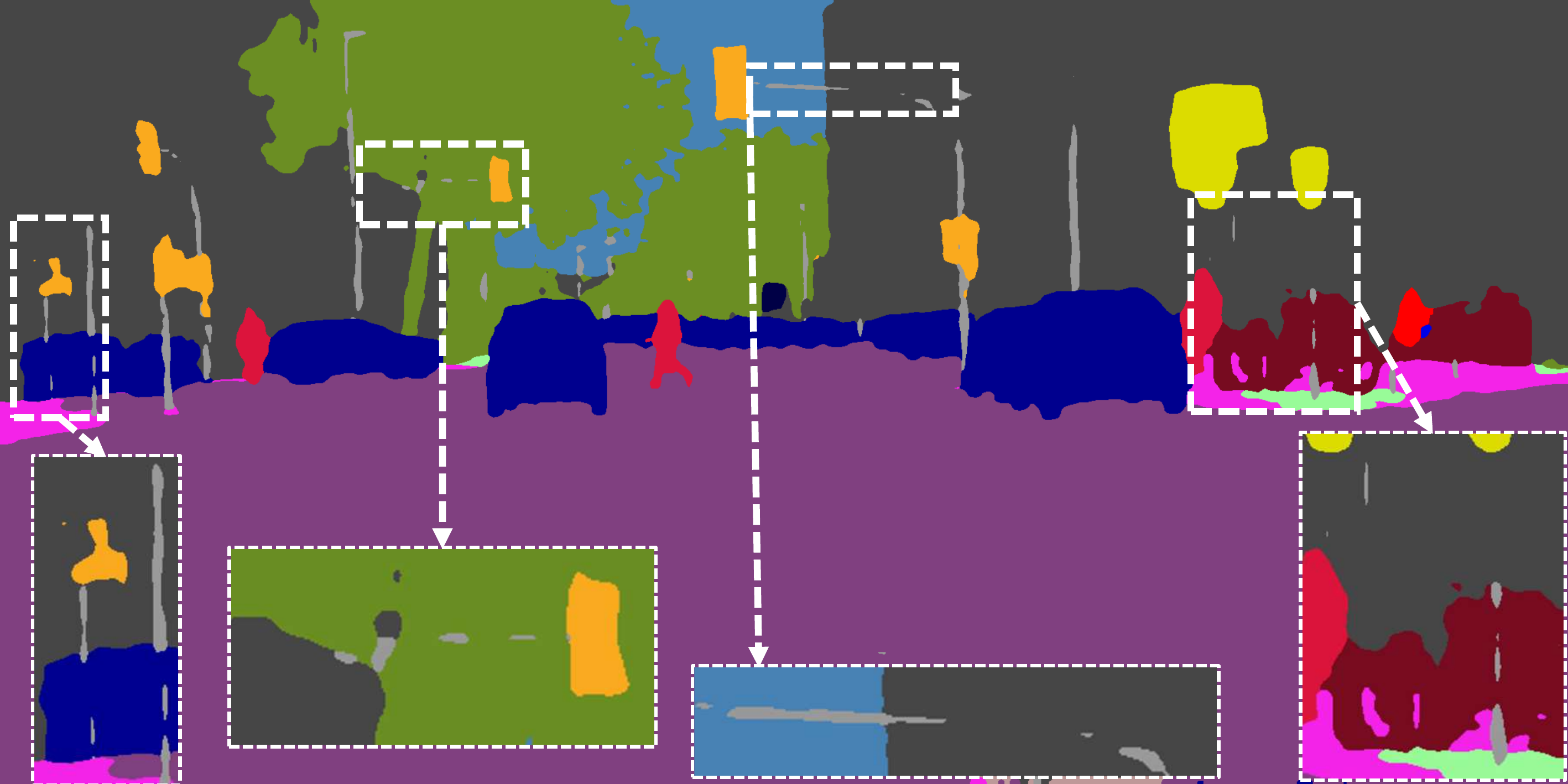}} \hfill 
	\subfloat{\includegraphics[width=0.198\linewidth]{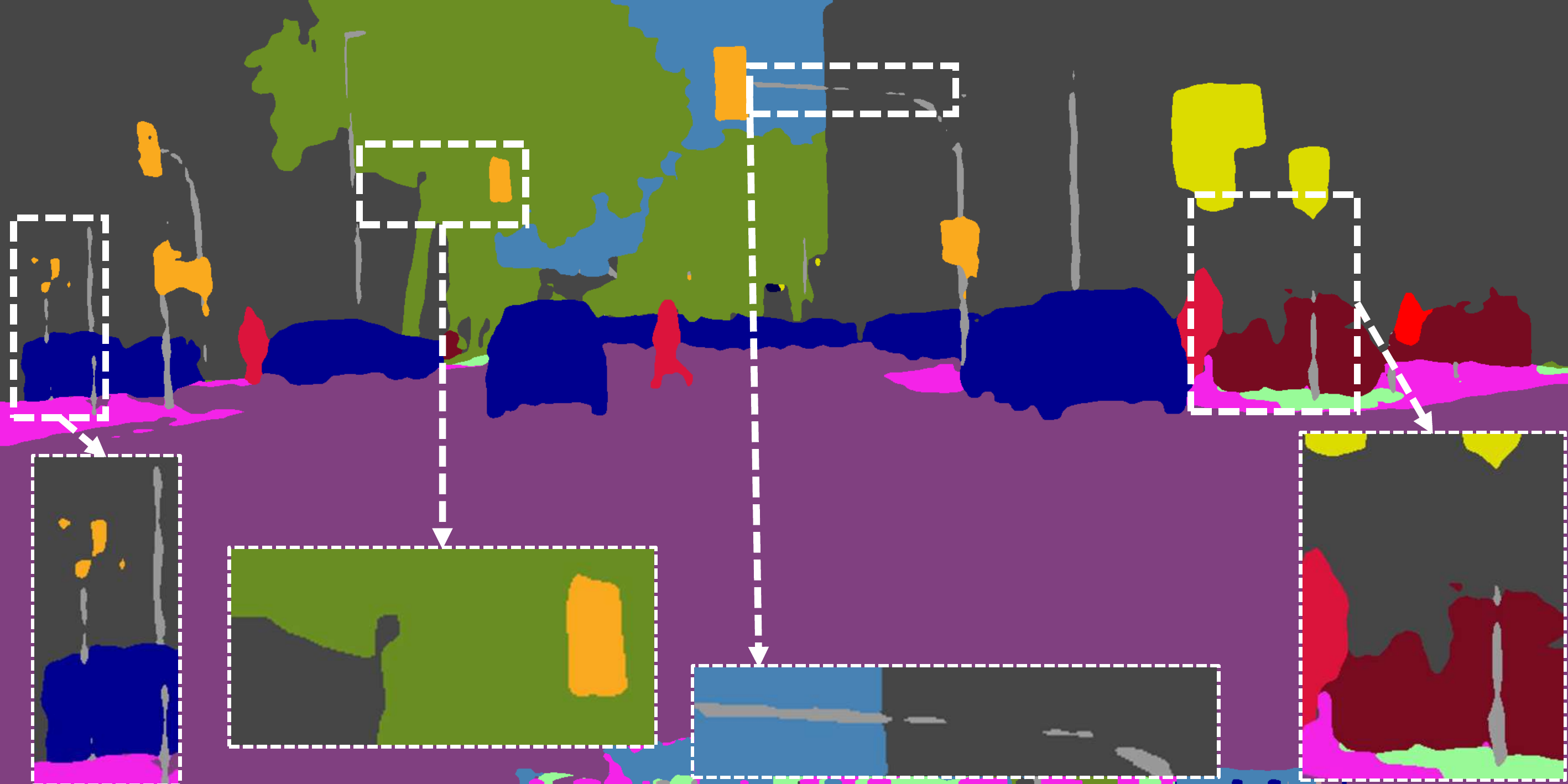}} \hfill 
	\subfloat{\includegraphics[width=0.198\linewidth]{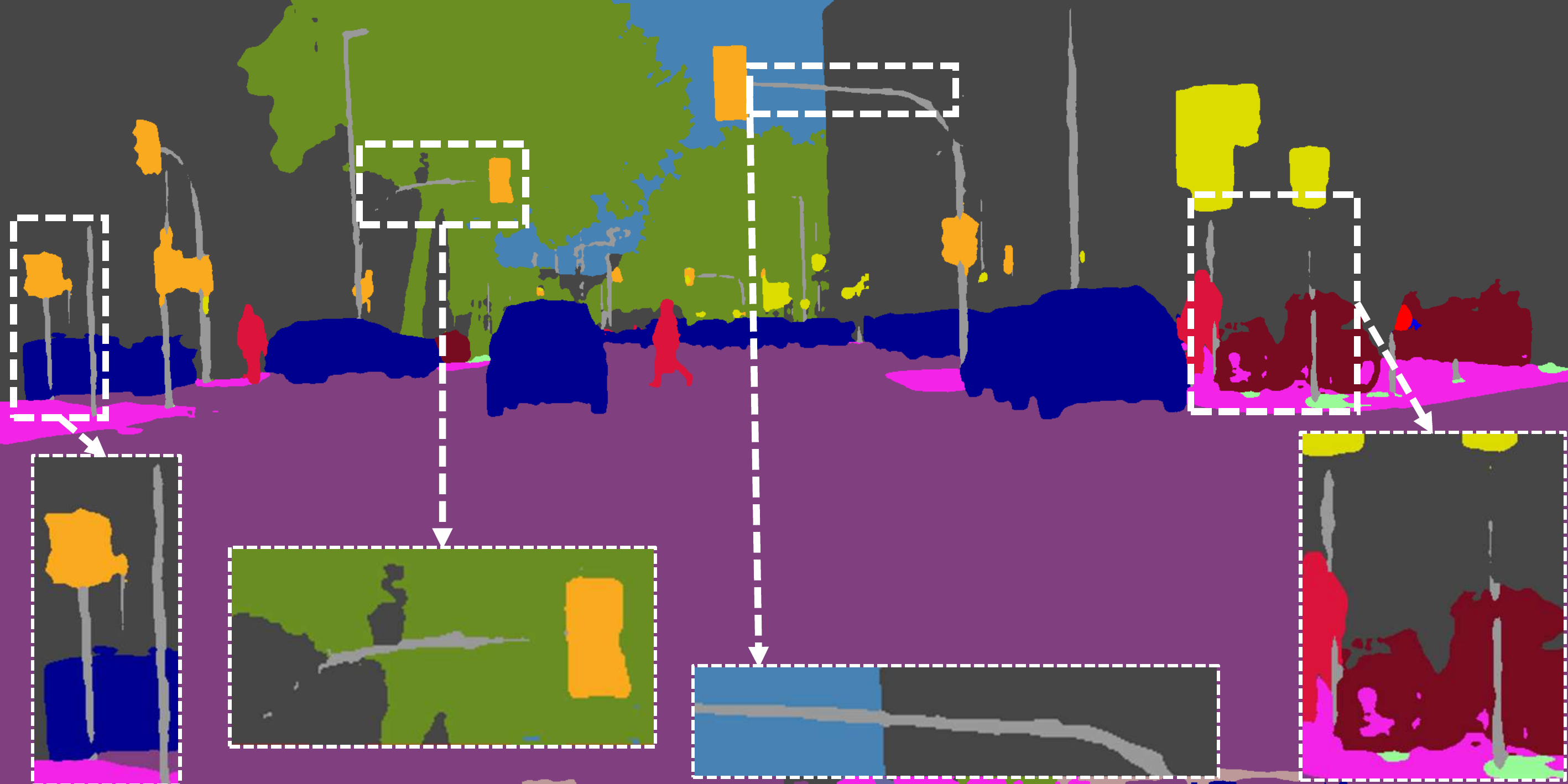}} \hfill 
	\subfloat{\includegraphics[width=0.198\linewidth]{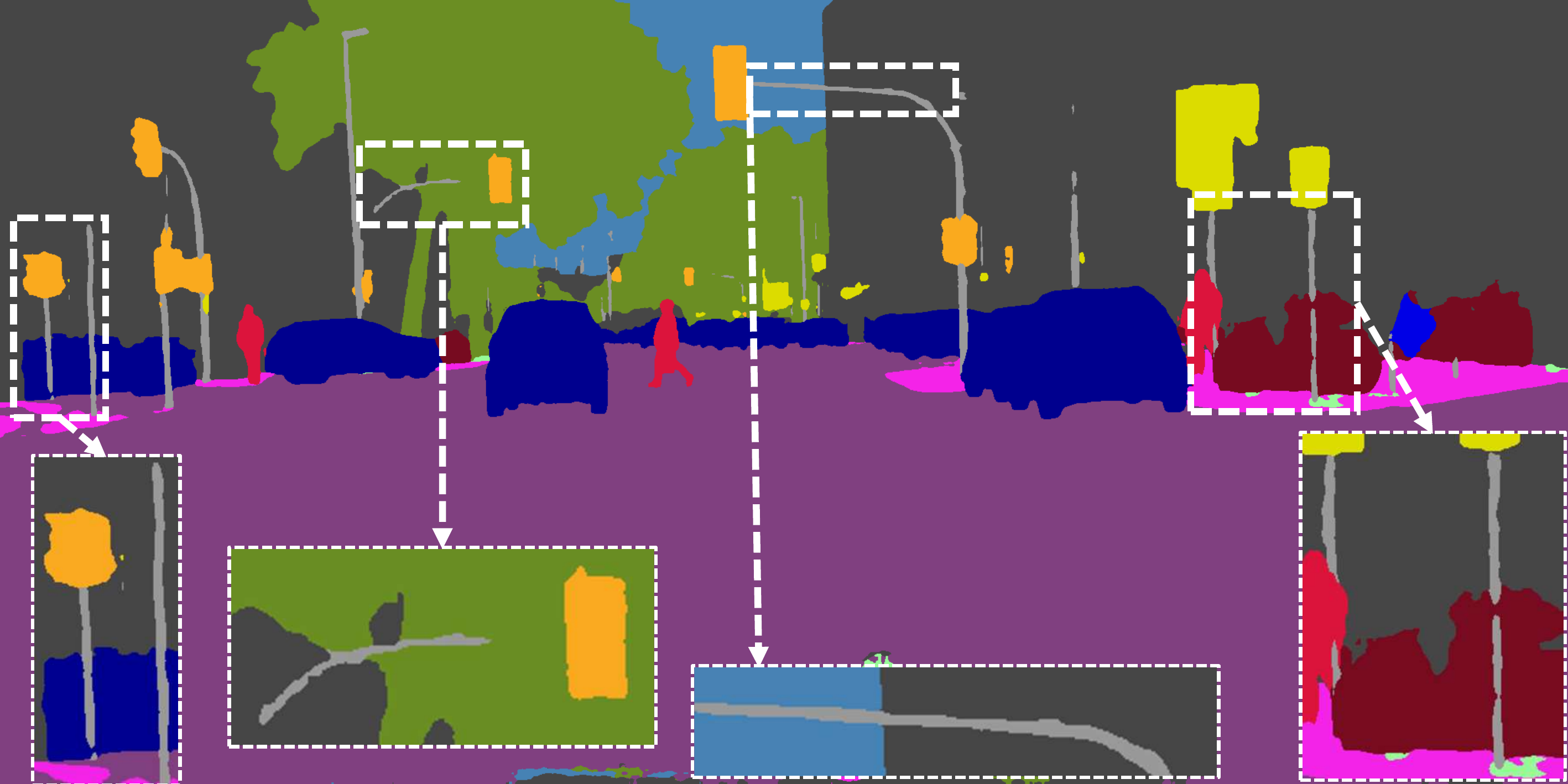}} \\ \vspace{-0.30cm} 
        \subfloat{\includegraphics[width=0.198\linewidth]{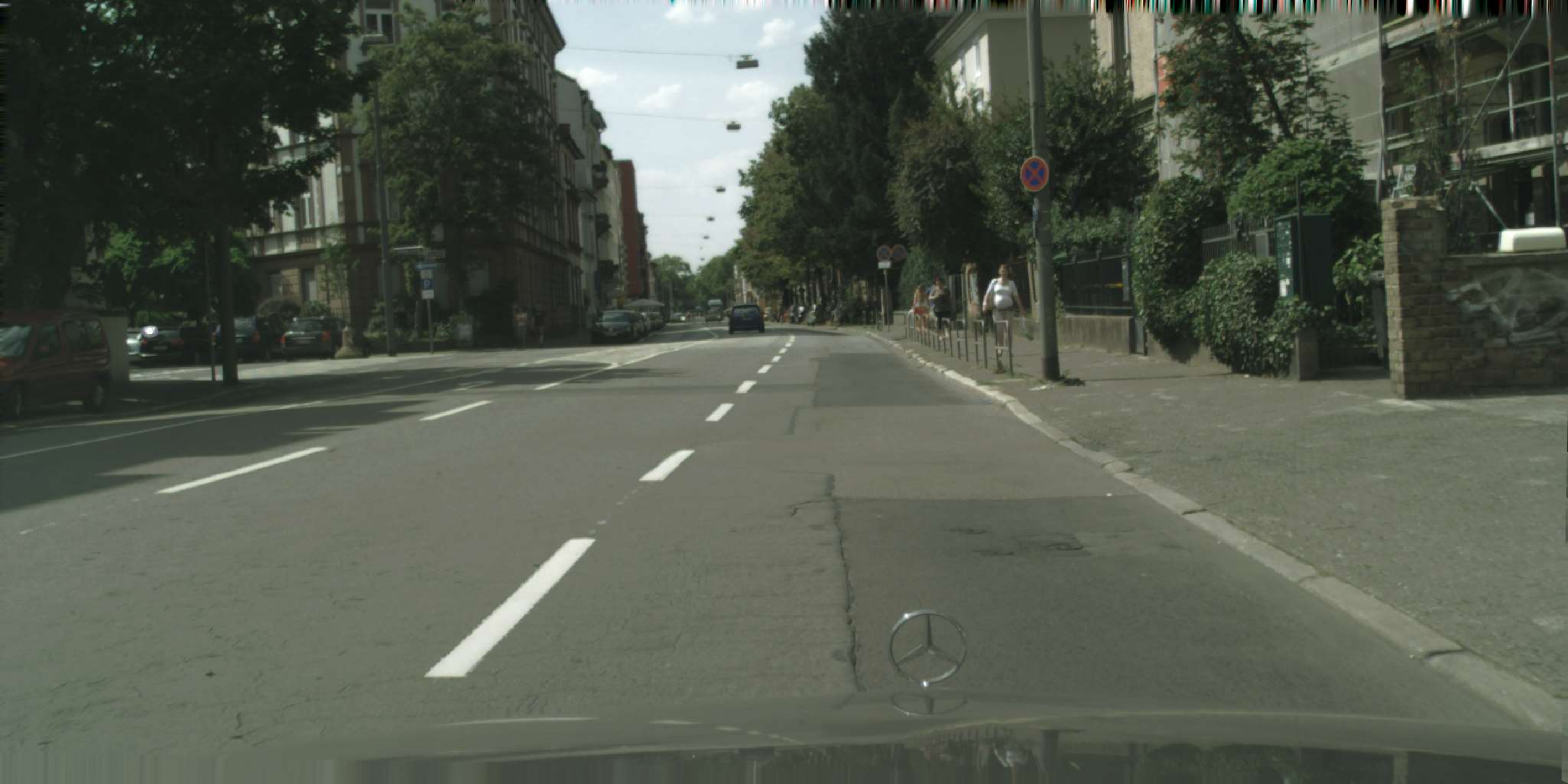}} \hfill
	\subfloat{\includegraphics[width=0.198\linewidth]{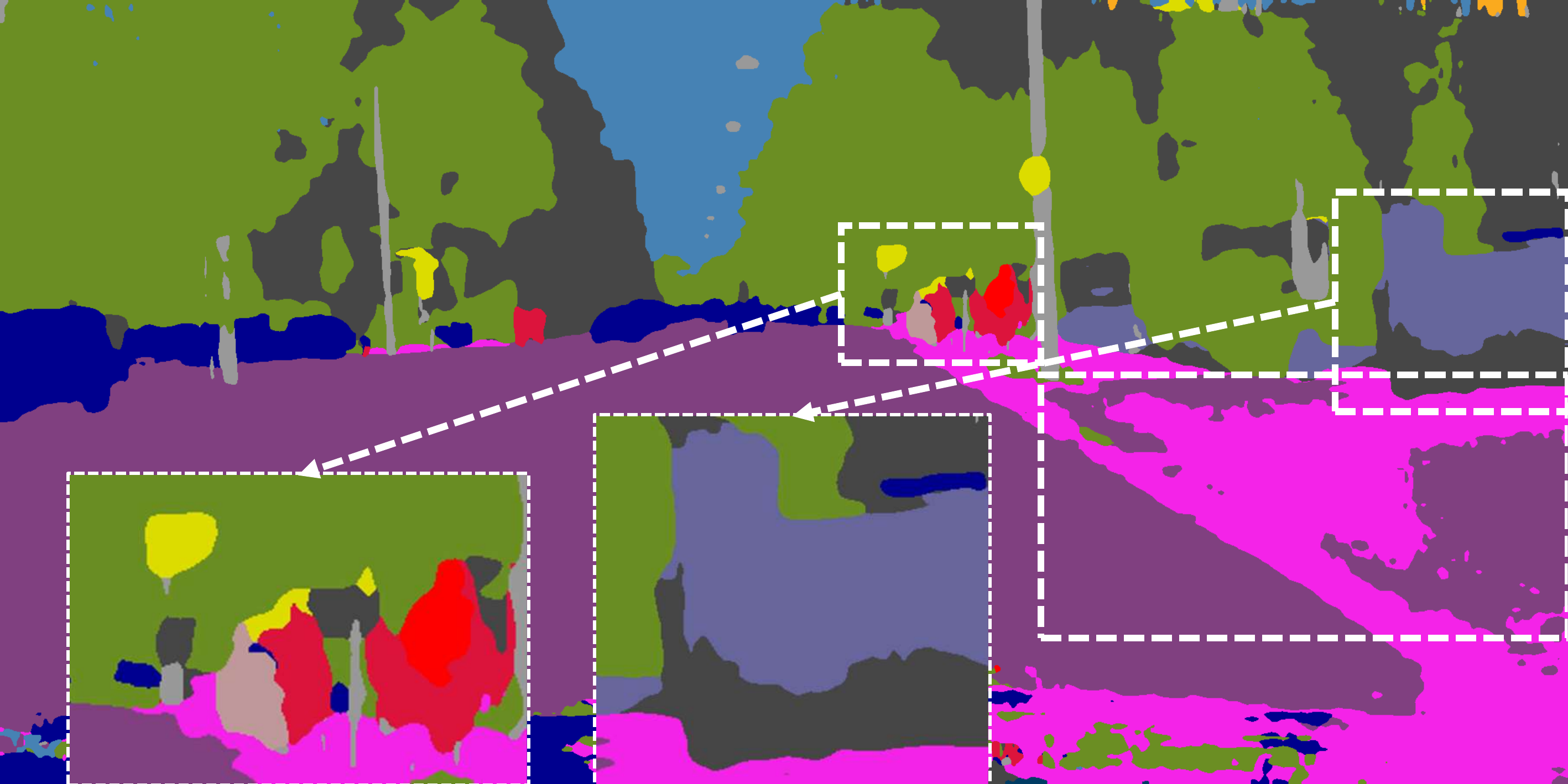}} \hfill
	\subfloat{\includegraphics[width=0.198\linewidth]{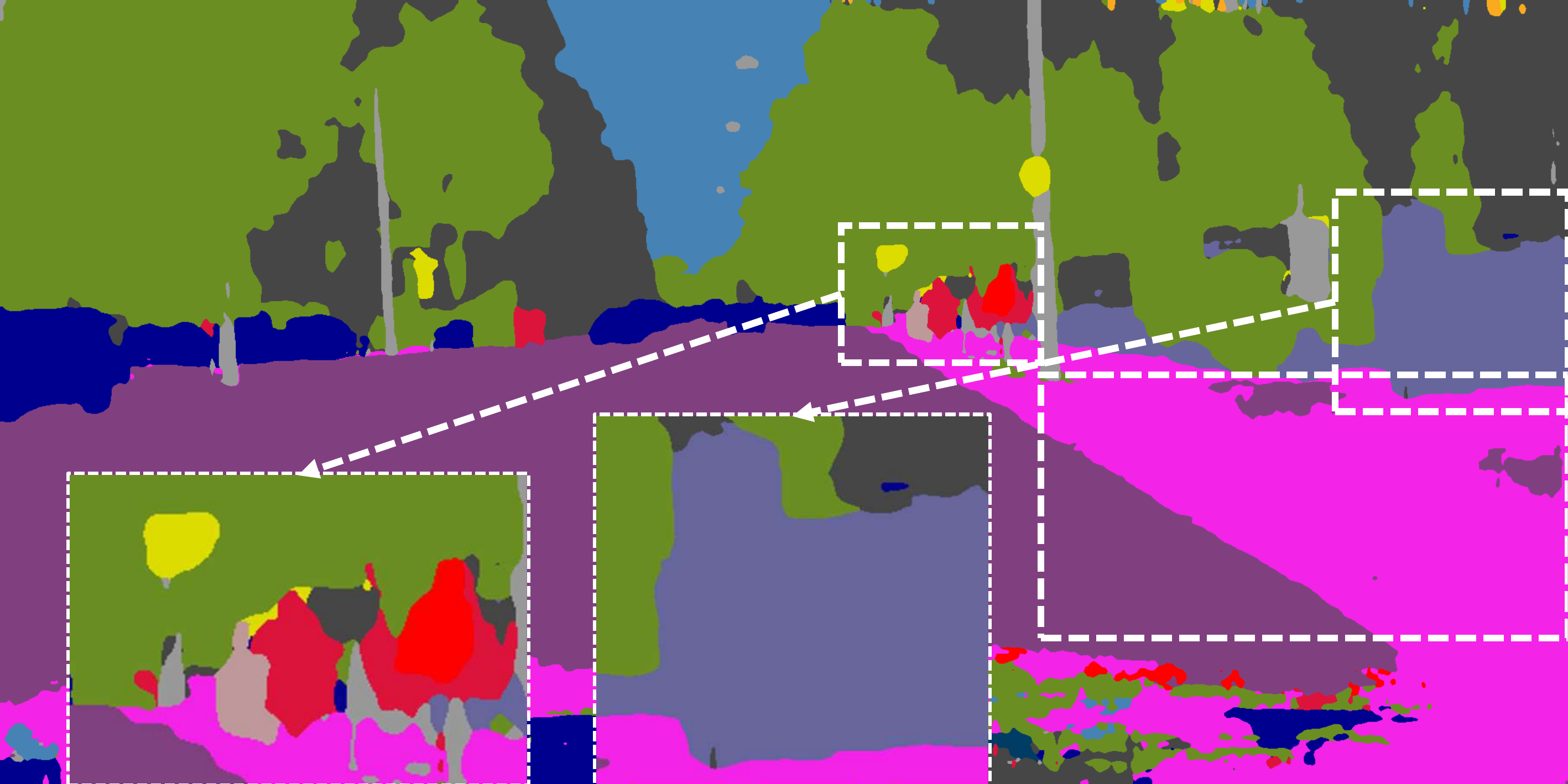}} \hfill
	\subfloat{\includegraphics[width=0.198\linewidth]{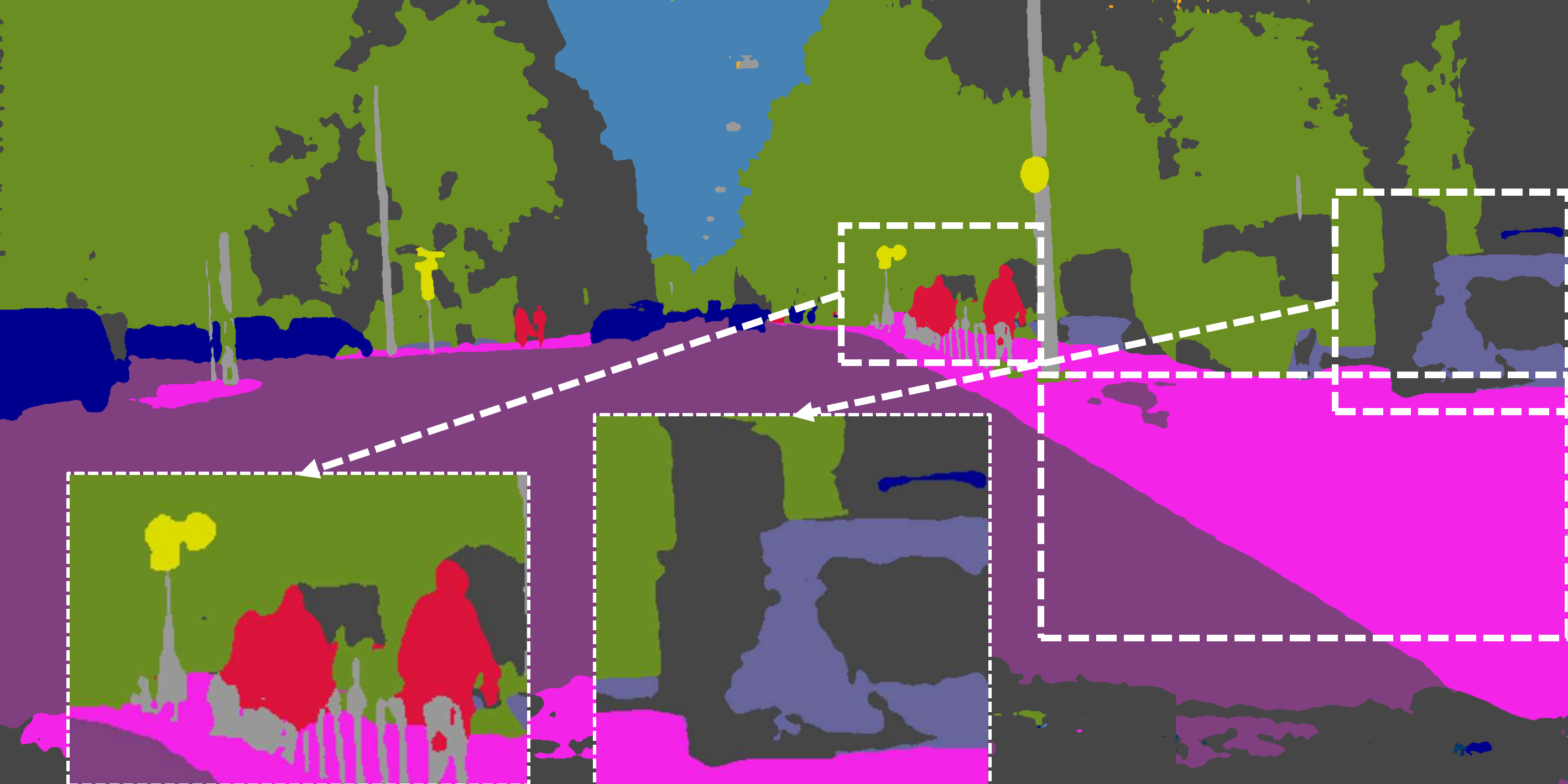}} \hfill
	\subfloat{\includegraphics[width=0.198\linewidth]{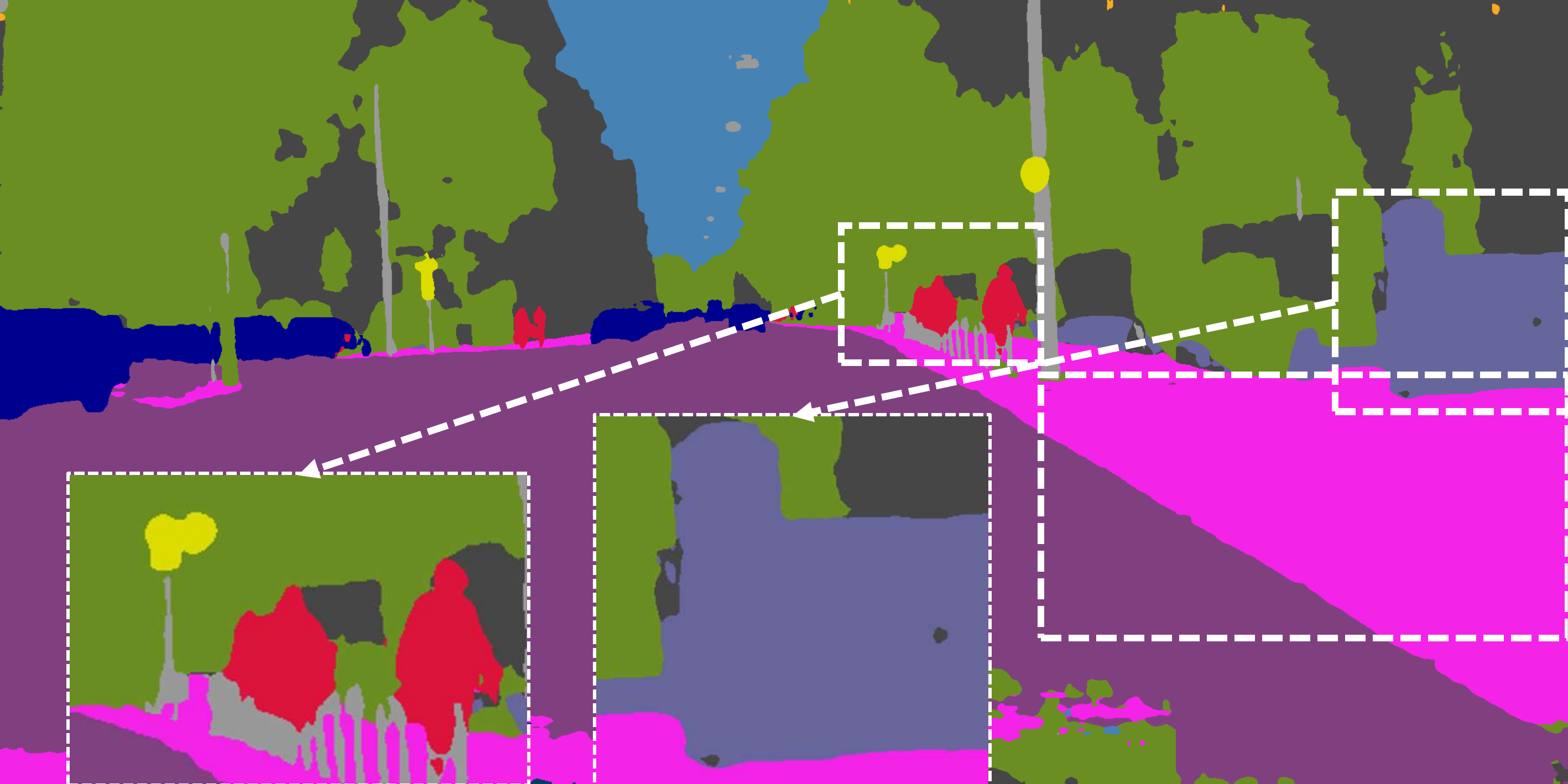}} \\ \vspace{-0.30cm}
        \subfloat{\includegraphics[width=0.198\linewidth]{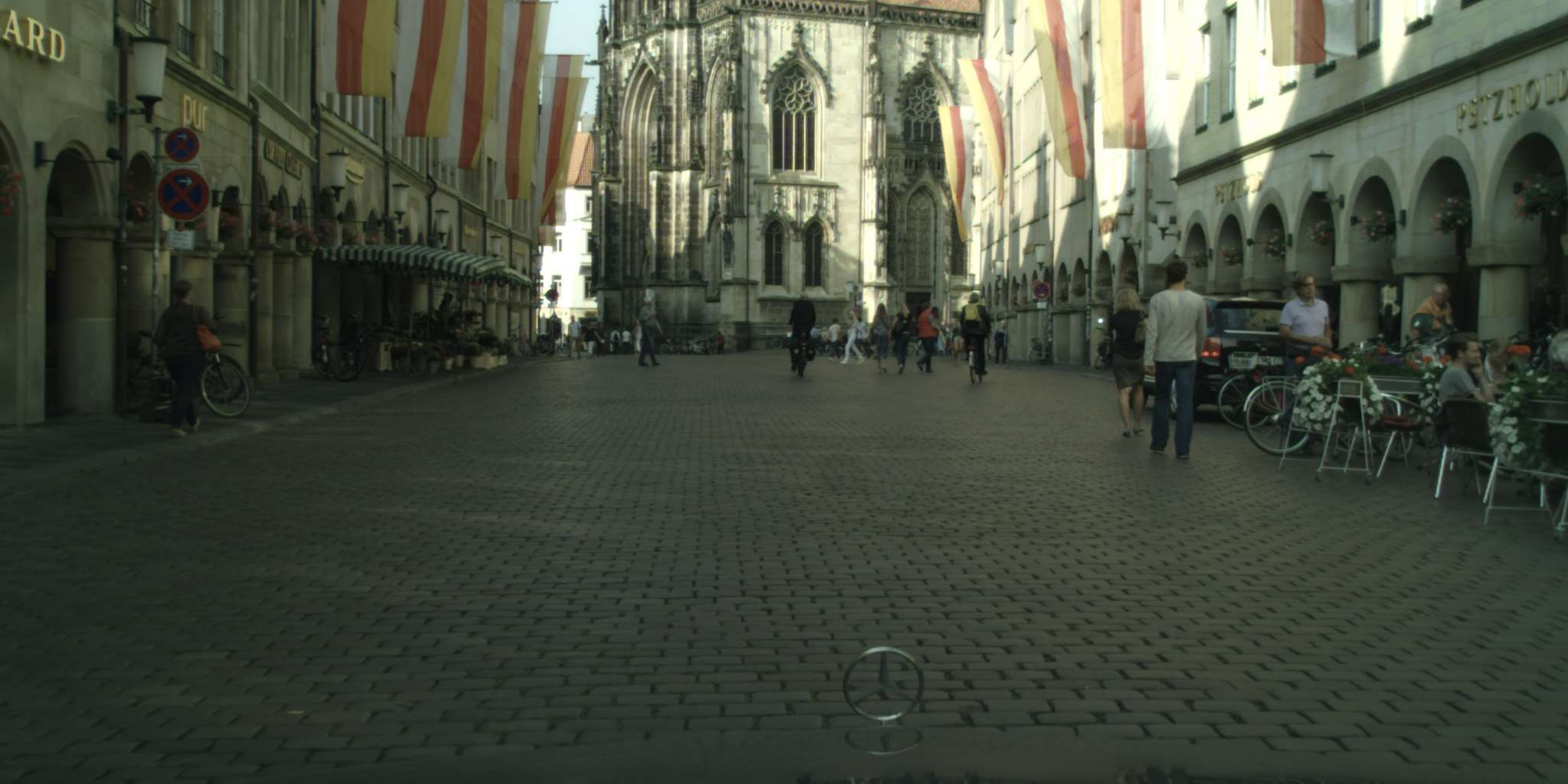}} \hfill
	\subfloat{\includegraphics[width=0.198\linewidth]{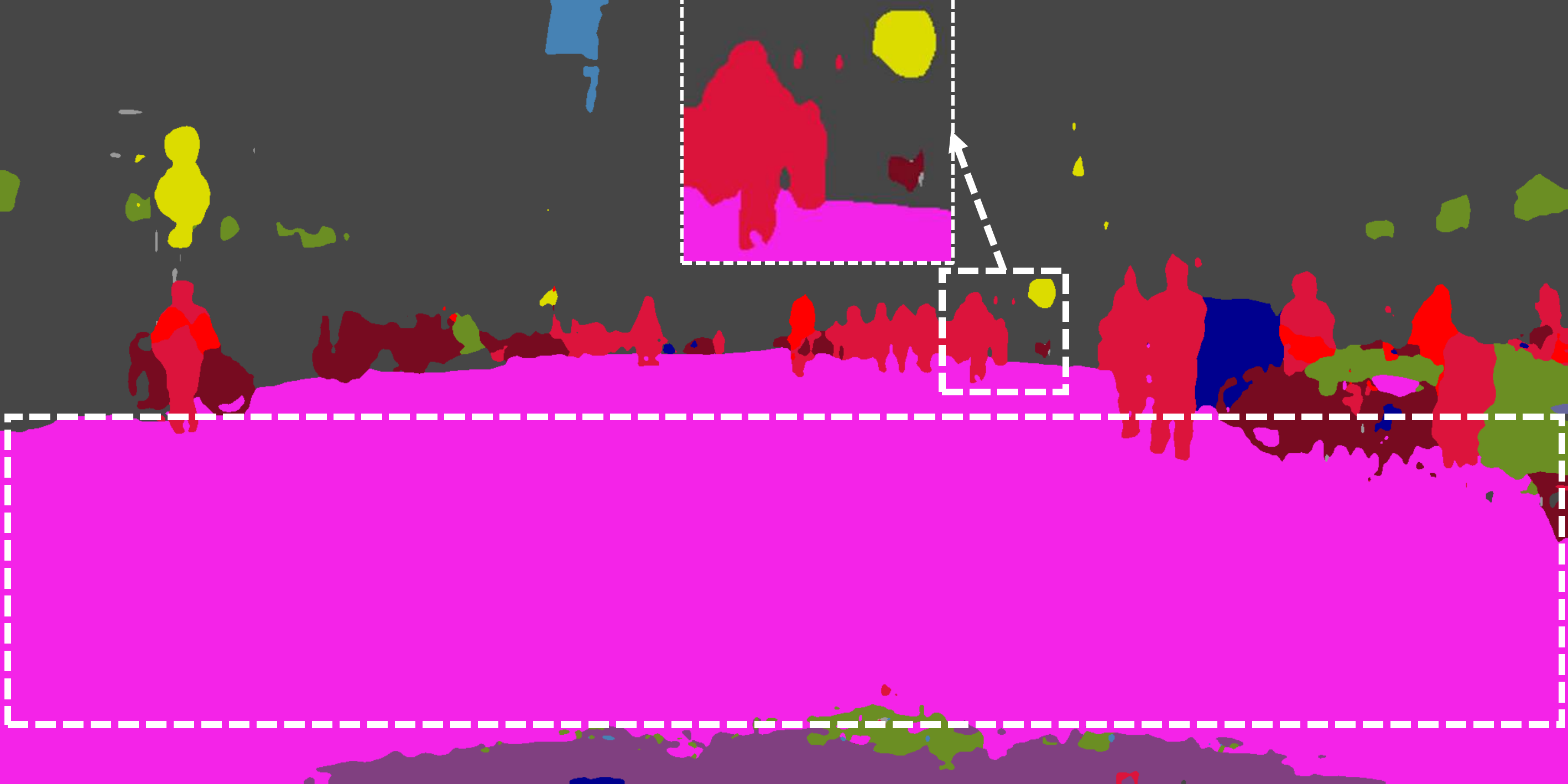}} \hfill 
	\subfloat{\includegraphics[width=0.198\linewidth]{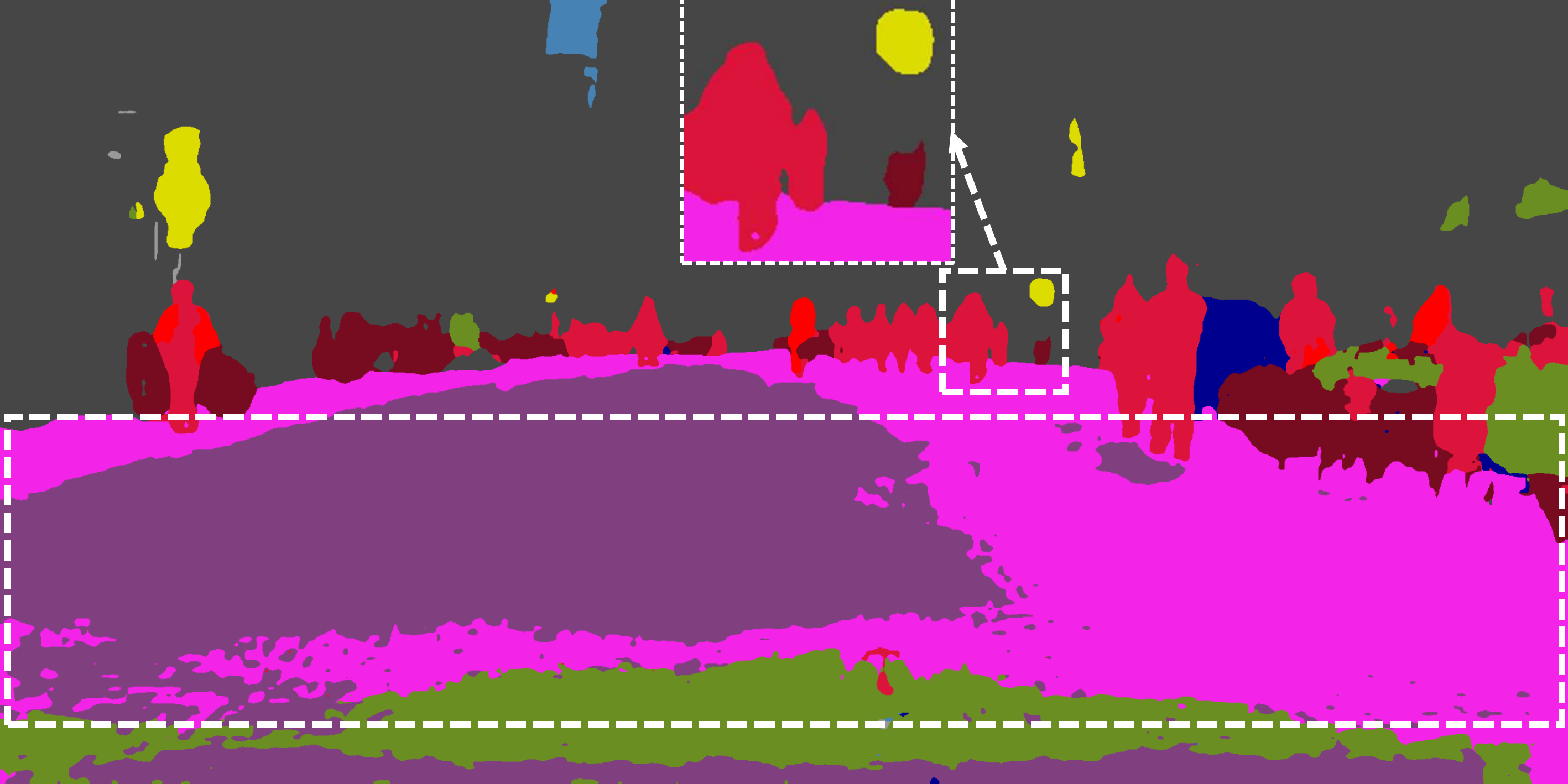}} \hfill 
	\subfloat{\includegraphics[width=0.198\linewidth]{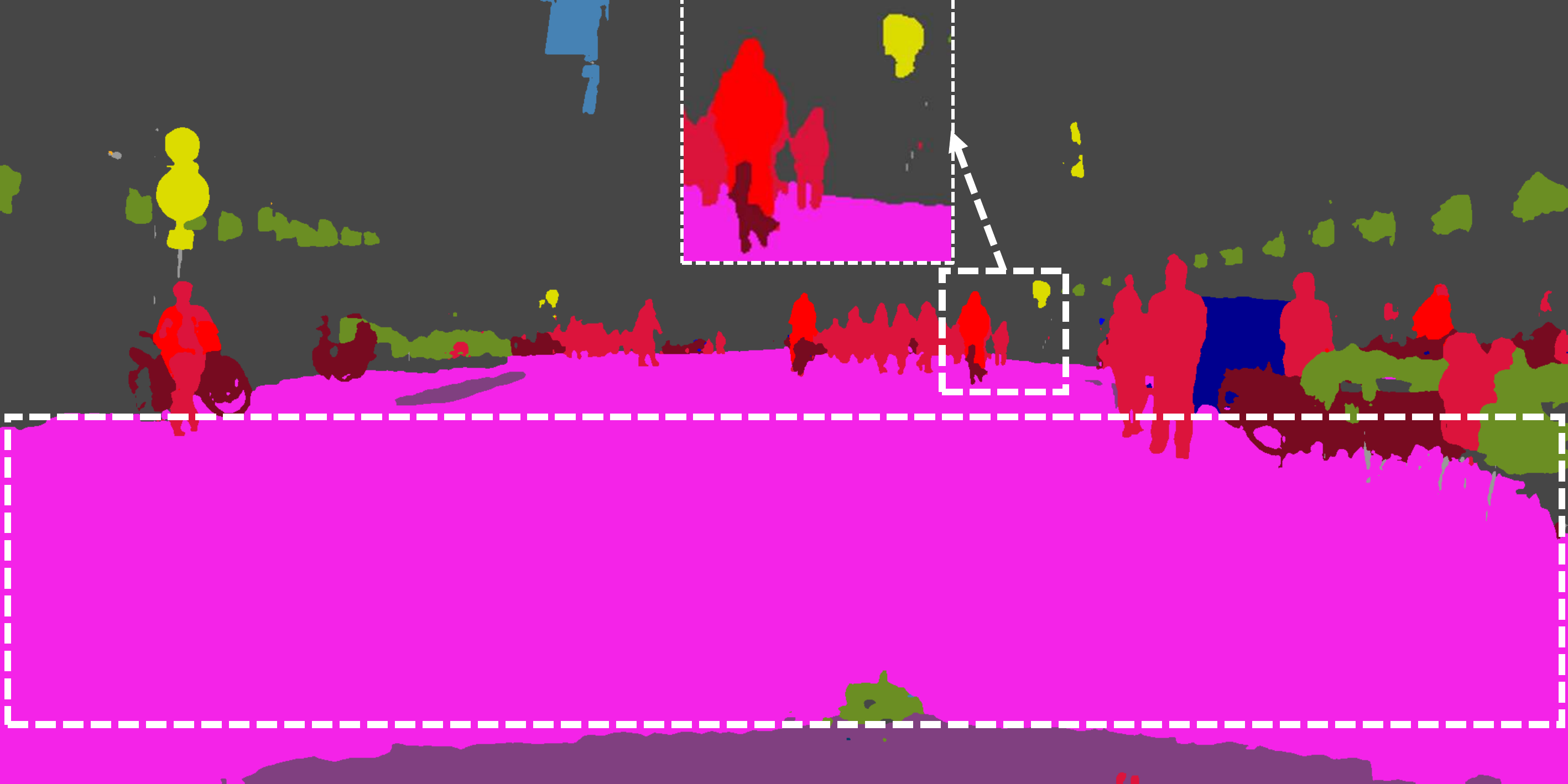}} \hfill 
	\subfloat{\includegraphics[width=0.198\linewidth]{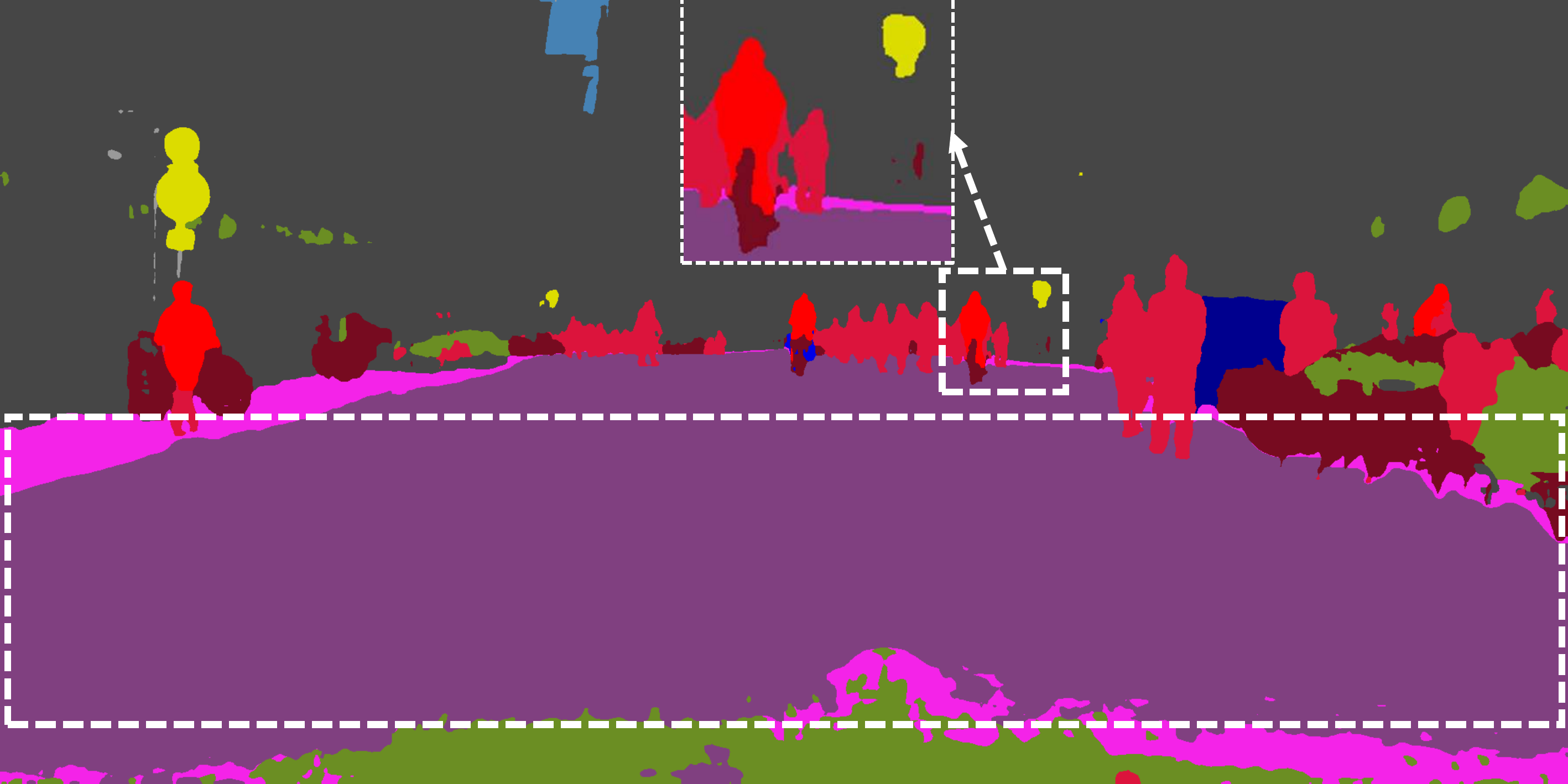}} \\  \vspace{-0.30cm} 
        \subfloat{\includegraphics[width=0.198\linewidth]{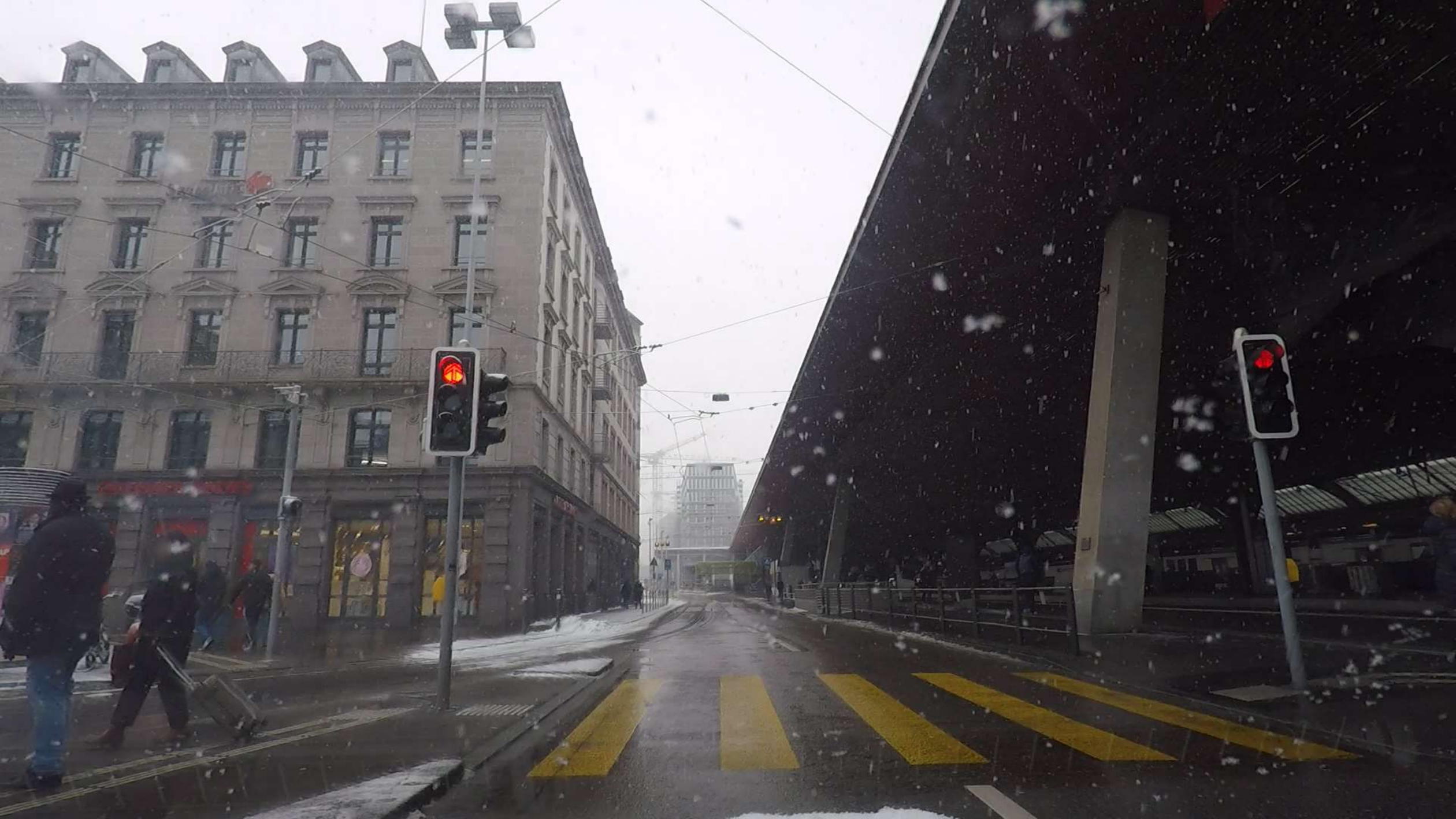}} \hfill
	\subfloat{\includegraphics[width=0.198\linewidth]{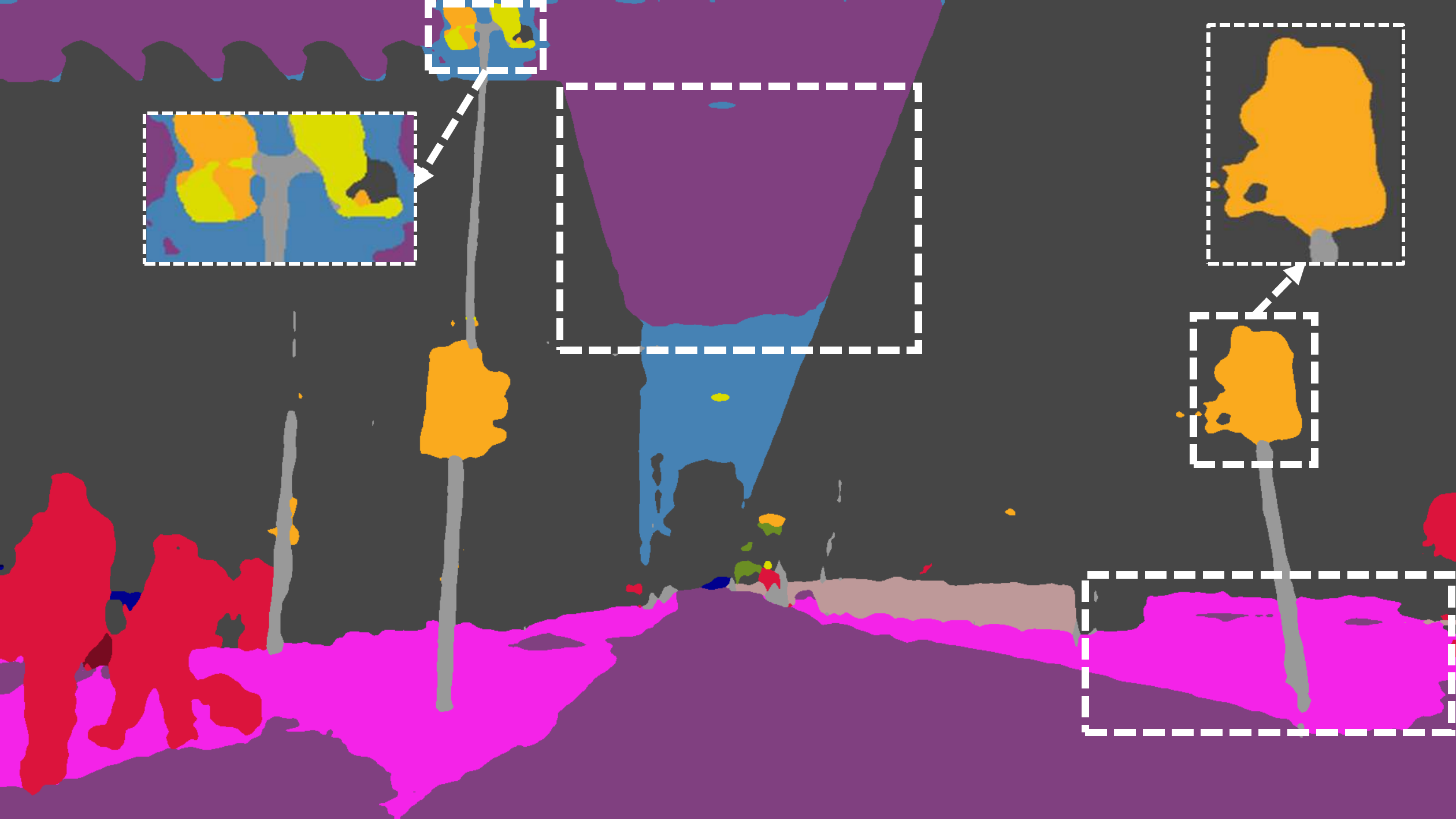}} \hfill
	\subfloat{\includegraphics[width=0.198\linewidth]{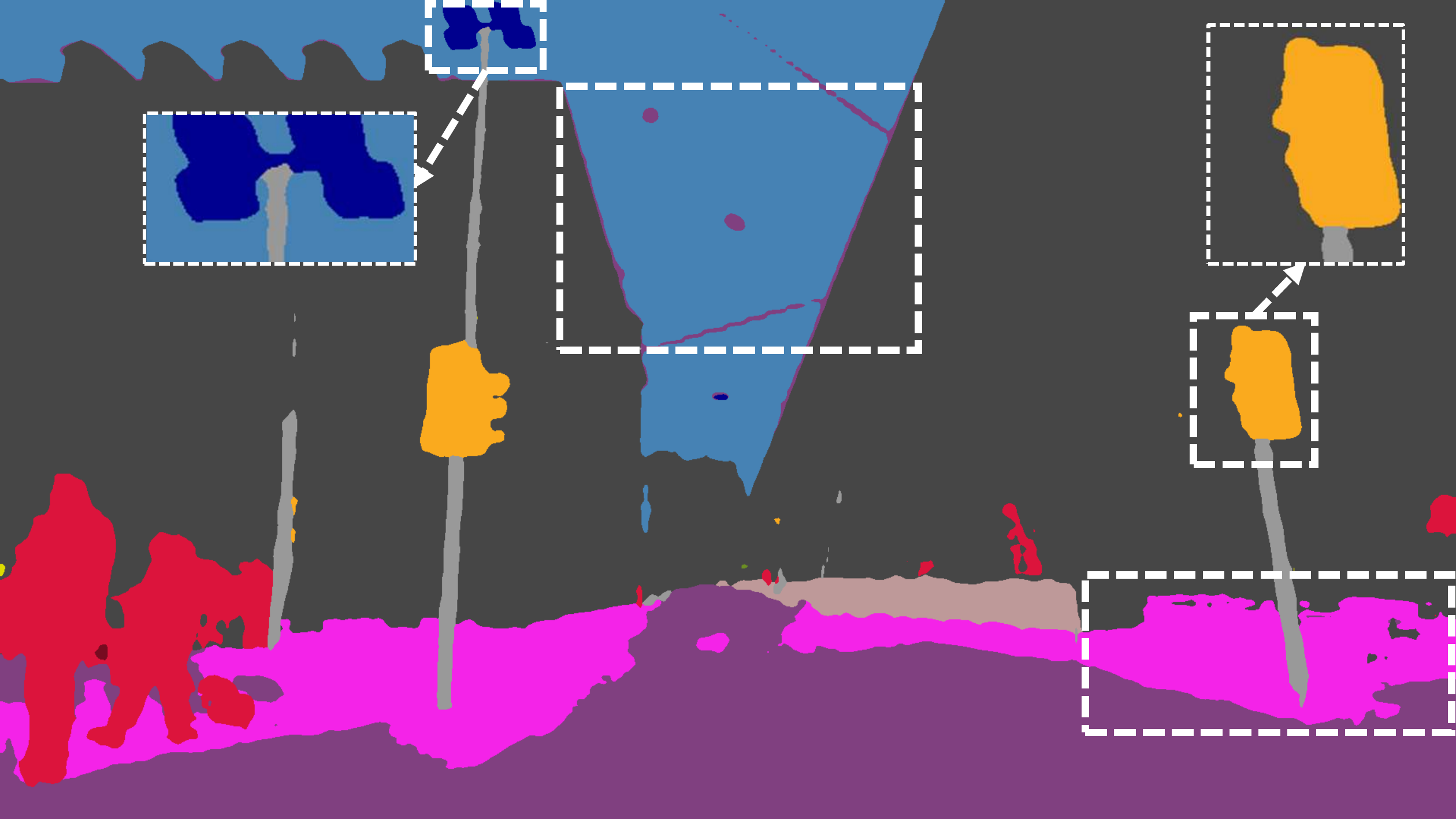}} \hfill
	\subfloat{\includegraphics[width=0.198\linewidth]{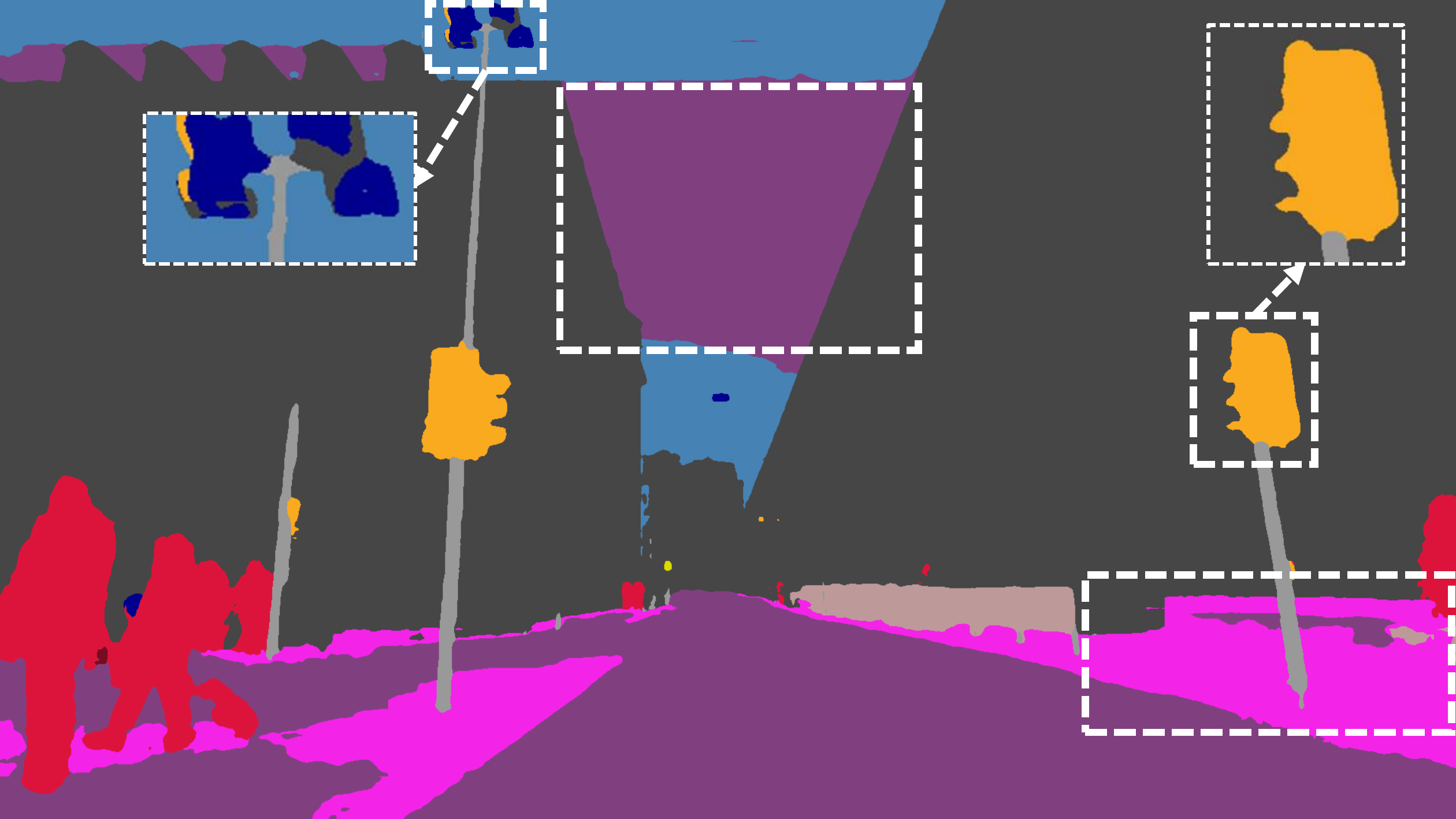}} \hfill
	\subfloat{\includegraphics[width=0.198\linewidth]{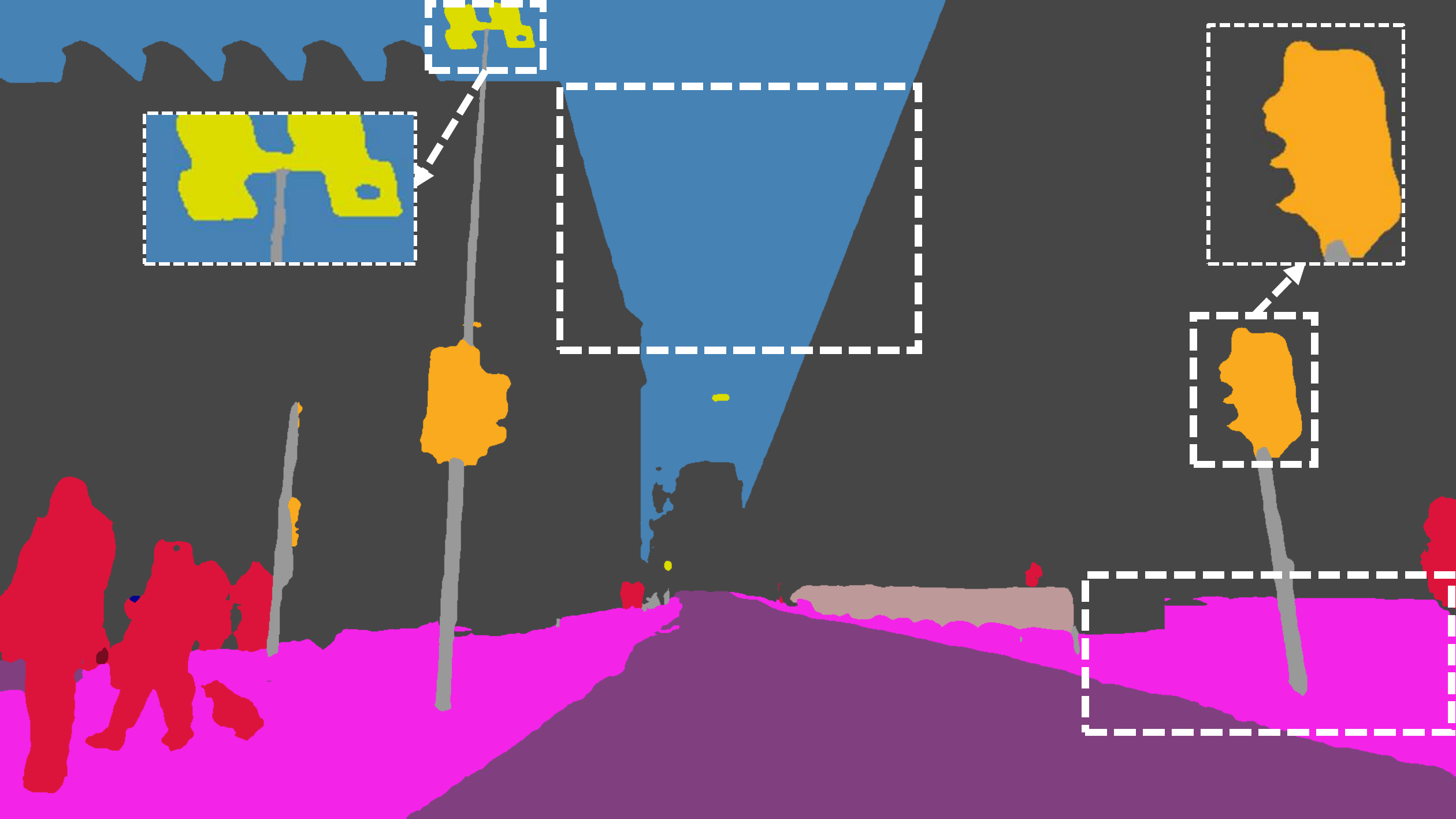}} \\ \vspace{-0.30cm}
        \subfloat[Image]{\includegraphics[width=0.198\linewidth]{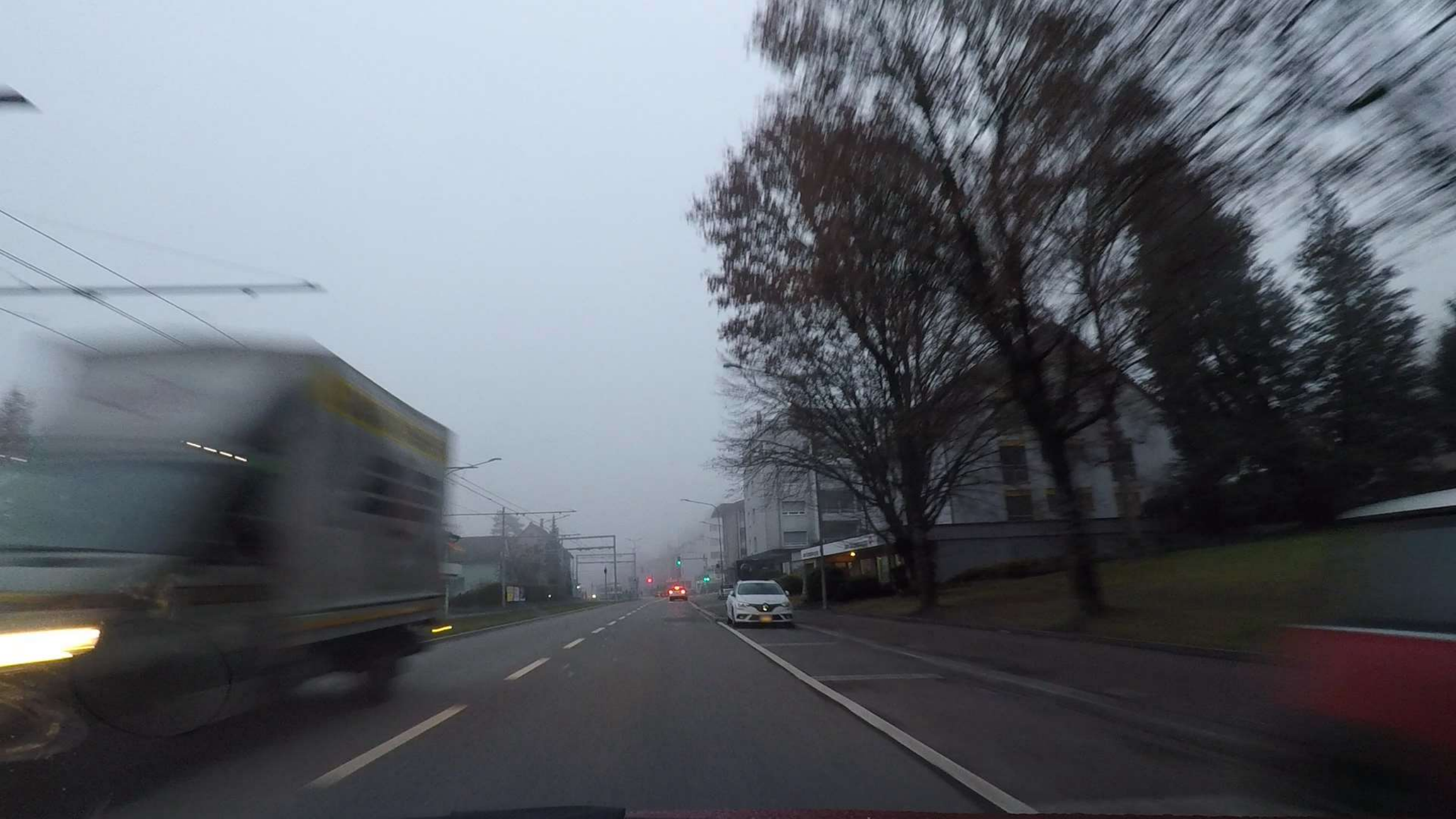}} \hfill
	\subfloat[DAFormer]{\includegraphics[width=0.198\linewidth]{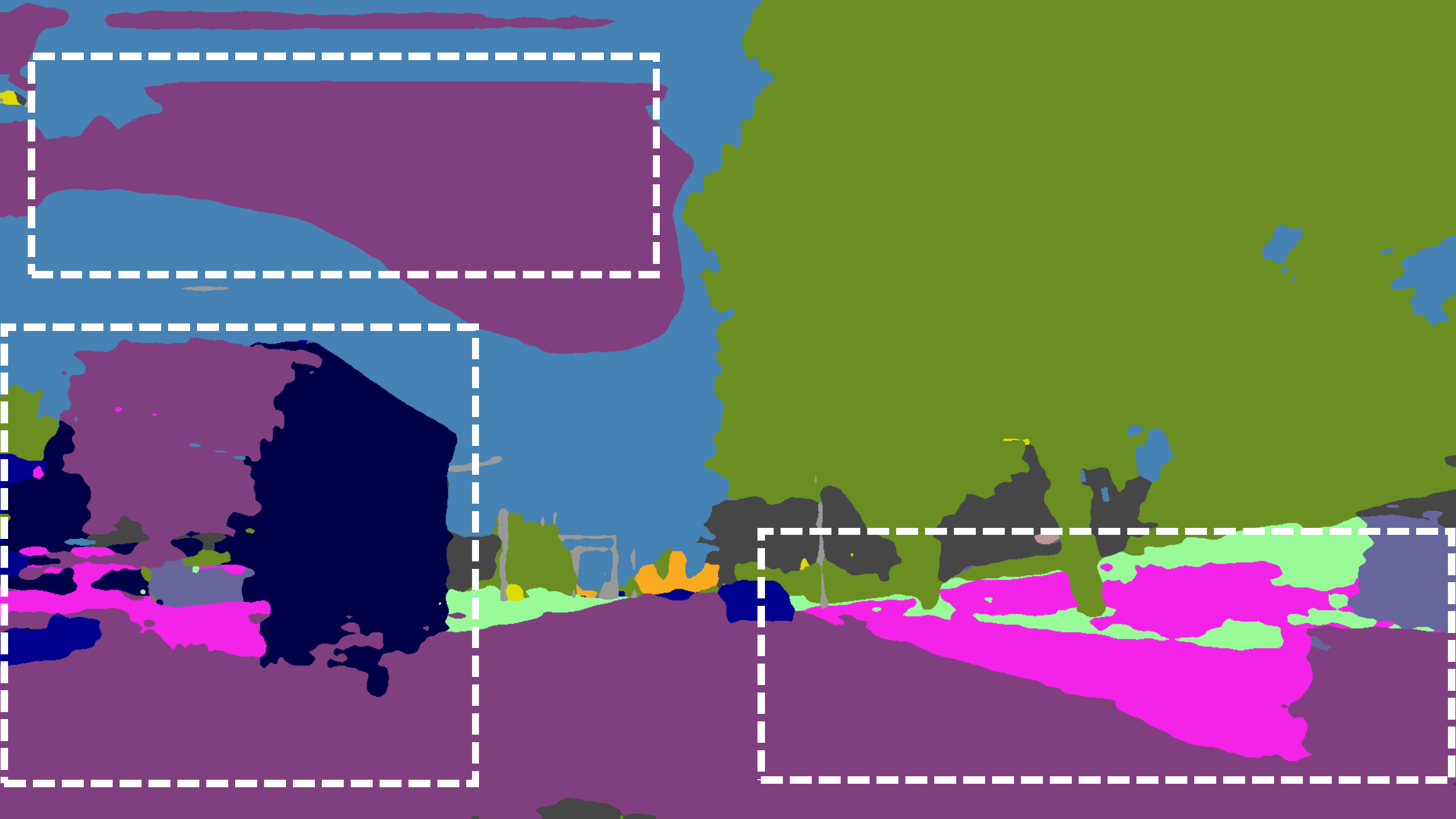}} \hfill
	\subfloat[DAFormer+SAM4UDASS]{\includegraphics[width=0.198\linewidth]{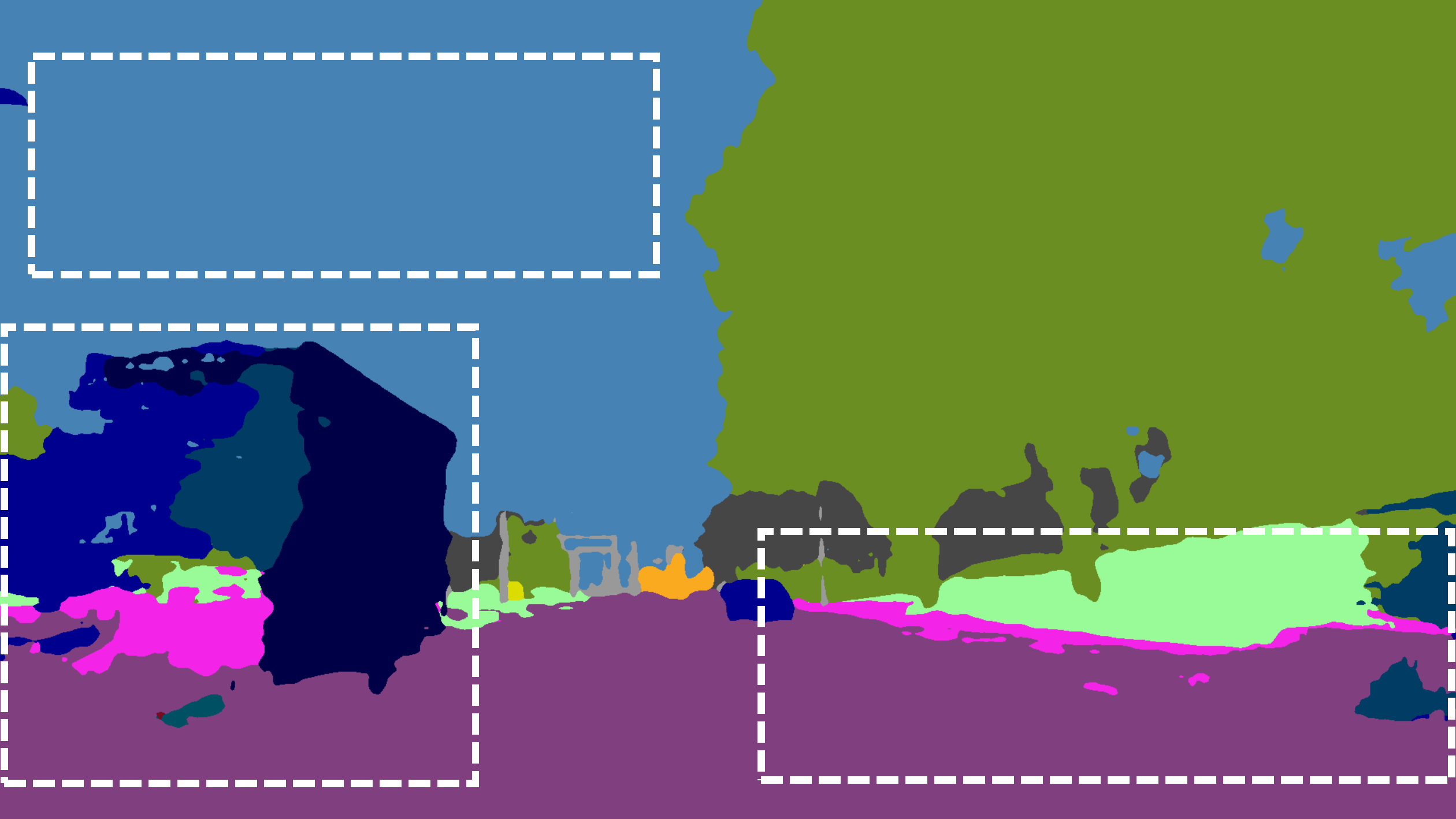}} \hfill
	\subfloat[MIC]{\includegraphics[width=0.198\linewidth]{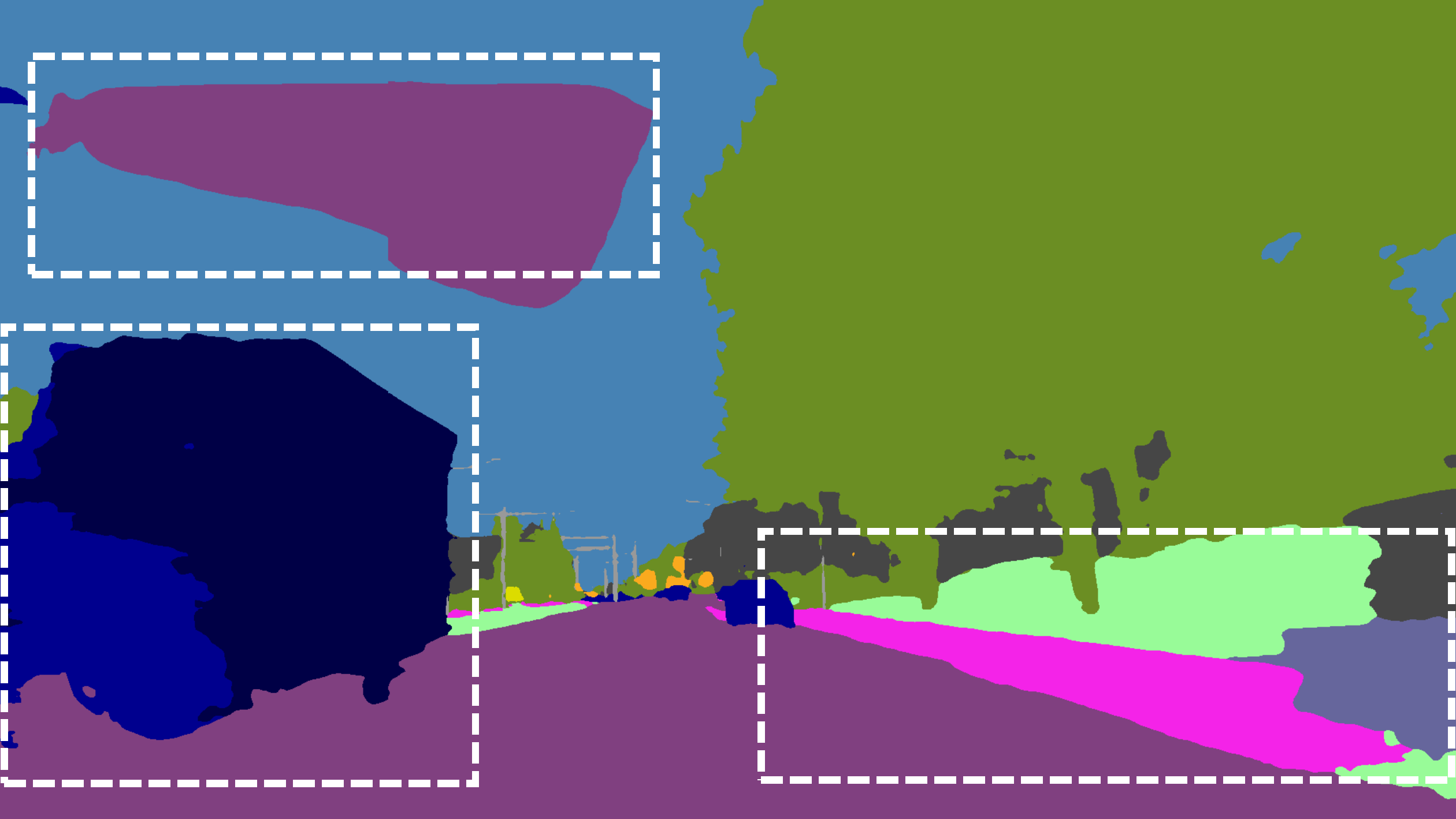}} \hfill
	\subfloat[MIC+SAM4UDASS]{\includegraphics[width=0.198\linewidth]{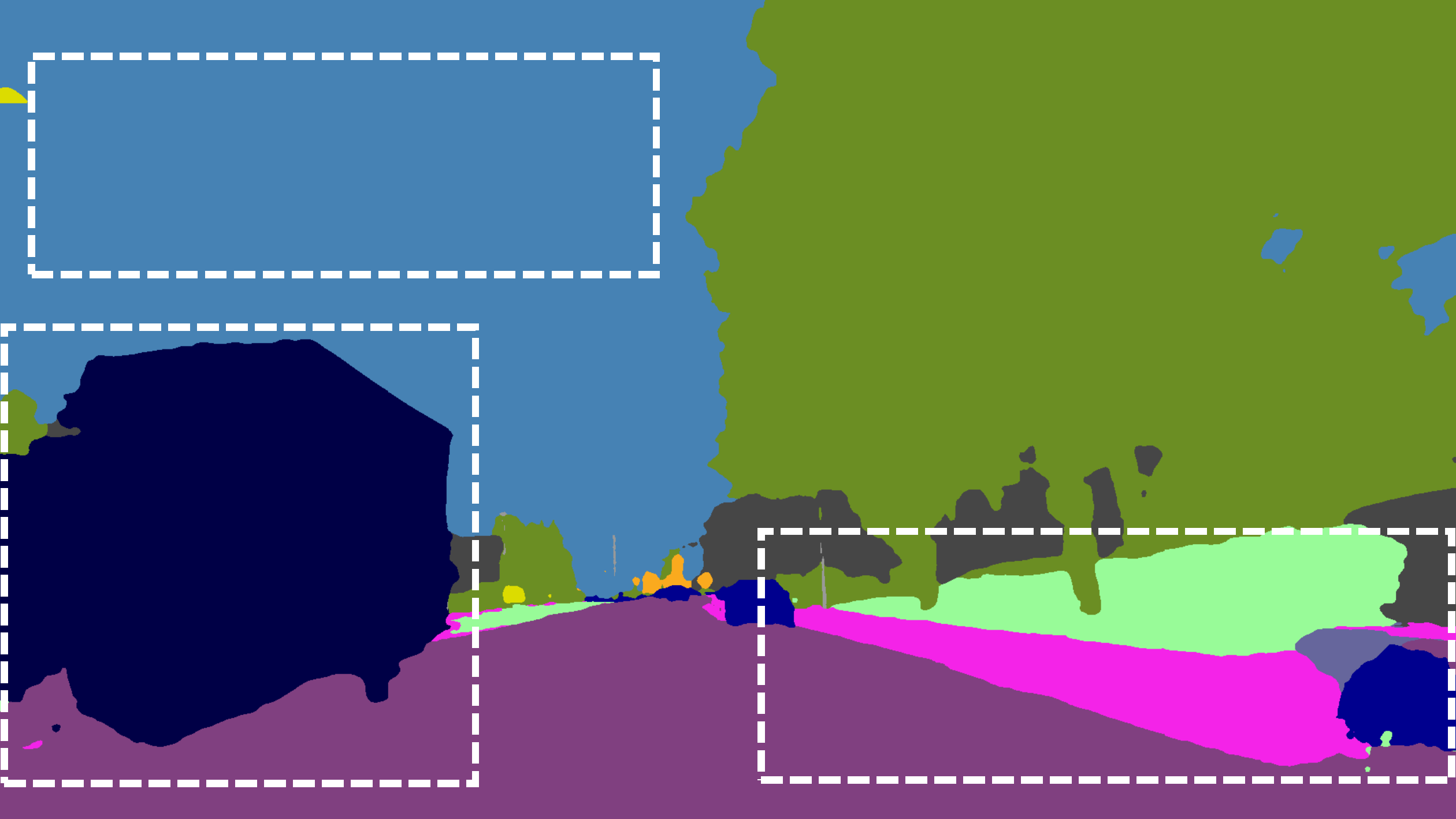}}
	\caption{Qualitative results of different self-training UDA methods on GTA5-to-Cityscapes (rows 1-2), SYNTHIA-to-Cityscapes (rows 3-4), and Cityscapes-to-ACDC (rows 5-6). 
	From left to right: the image, the prediction of DAFormer, DAFormer + SAM4UDASS, MIC, MIC + SAM4UDASS.
        } 
        \label{fig:pics_gta5}
\end{figure*}

Table~\ref{tab:gta5_comparison} shows the comprehensive performance of SAM4UDASS on GTA5-to-Cityscapes. It delivers mIoU enhancements of 1.9\%, 3.2\%, 3.0\%, and 1.9\% for TUFL, SePiCo, DAFormer, and MIC, respectively. It should be noted that we only tune the parameters based on DAFormer and then directly apply SAM4UDASS to other self-training UDA methods. When using MIC, SAM4UDASS achieves SOTA 77.3\% mIoU and performs best in most categories. The most significant improvements are observed in classes like fence, pole, rider, truck, and bike, where fence, pole, and bike belong to $C_s$, highlighting the capability of SAM4UDASS in refining pseudo-labels, especially for rare and small objects.

The first two rows of Fig.~\ref{fig:pics_gta5} show the qualitative results. The integration of SAM4UDASS leads to improved segmentation of small classes like pole, traffic light, and bicycle, while also yielding more precise boundaries due to SAM's inherent characteristics. Meanwhile, the fence in the first row is better classified, which is a rare class in GTA5 and Cityscapes.

\subsubsection{SYNTHIA-to-Cityscapes}

\begin{table*}[!ht]
        \centering
        \captionsetup{format=myformat}
        \setlength{\tabcolsep}{3pt}
        \renewcommand{\arraystretch}{1.1}
	\caption{The adaptation performance and comparison on SYNTHIA-to-Cityscapes} 
	\label{tab:syn_comparison}
	\resizebox{\linewidth}{!}{
                \begin{tabular}{c|c|cccccccccccccccc|c}
                \hline
                Methods & Network & road & sw & build & wall & fence & pole & light & sign & vege & sky & person & rider & car & bus & motor & bike & mIoU \\ \hline
		TUFL\cite{TUFL} & BiSeNet & 89.6  & 56.2  & 83.9  & 16.9  & 0.6  & 41.6  & 30.2  & 47.4  & 87.3  & 88.6  & 62.4  & 26.7  & 72.0  & 31.0  & 11.0  & 51.1 & 49.8 \\ 
		SePiCo\cite{sepico} & DeeplabV2 & 80.5  & 38.2  & 84.0  & 12.8  & 2.5  & 41.7  & 51.0  & 53.5  & 83.6  & 80.1  & 73.4  & 44.5  & 85.4  & 52.0  & 50.5  & 65.0  & 56.2 \\ \hline 
		TUFL+\textbf{SAM4UDASS} & BiSeNet & 92.7  & 59.7  & 84.6  & 14.9  & 0.5  & 41.6  & 32.3  & 49.9  & 87.7  & 90.0  & 63.5  & 29.0  & 77.8  & 31.2  & 11.1  & 54.5 & 51.3  \\ 
                SePiCo+\textbf{SAM4UDASS} & DeeplabV2 & 90.5  & 54.5  & 84.1  & 17.9  & 4.2  & 43.2  & 54.4  & 57.0  & 84.2  & 86.0  & 75.1  & 49.7  & 89.1  & 52.0  & 54.7  & 68.5 & 60.3 \\ \hline
                DAFormer\cite{daformer} & \multirow{5}{*}{\begin{tabular}{c} DAFormer\end{tabular} } & 86.0  & 46.9  & 88.8  & 42.5  & 8.1  & 50.1  & 54.4  & 54.2  & 85.3  & 88.3  & 73.2  & 46.5  & 87.3  & 59.7  & 49.2  & 60.6  & 61.3 \\
                HRDA\cite{hrda} &  & 88.1  & 50.5  & 89.0  & 46.3  & 7.2  & 56.9  & 66.0  & 62.4  & 86.9  & 94.5  & 79.1  & 54.2  & 90.2  & 63.2  & 65.5  & 65.6 & 66.6 \\
		MIC\cite{mic} &  & 85.7  & 47.9  & 87.5  & 42.5  & 8.9  & 57.7  & 66.7  & 63.5  & 87.0  & 94.6  & 81.2  & 58.2  & \textbf{90.8}  & \textbf{69.0} & \textbf{67.8} & 65.1 & 67.1 \\ \cline{1-1} \cline{3-19}
                DAFormer+\textbf{SAM4UDASS} & & \textbf{93.7} & \textbf{66.7} & \textbf{89.1} & 41.7  & \textbf{11.1}  & 51.4  & 55.8  & 56.3  & 86.5  & 90.8  & 75.2  & 49.7  & 88.7  & 60.3  & 53.8  & 68.5 & 65.0  \\
                MIC+\textbf{SAM4UDASS} &  & 90.8  & 61.1  & 88.2  & \textbf{46.8} & 10.1  & \textbf{60.7} & \textbf{68.5} & \textbf{64.4} & \textbf{87.2} & \textbf{94.8} & \textbf{82.1} & \textbf{60.0} & 90.7  & 63.8  & 67.0  & \textbf{72.8} & \textbf{69.3} \\ \hline
	\end{tabular}}
        \vspace{-0.5cm}
\end{table*}

Table~\ref{tab:syn_comparison} shows the overall performance of SAM4UDASS on SYNTHIA-to-Cityscapes. It consistently improves mIoU by 1.5\%, 4.1\%, 3.7\%, and 2.2\% for TUFL, SePiCo, DAFormer, and MIC, respectively. It attains SOTA 69.3\% mIoU and excels in majority of the classes based on MIC. For DAFormer, improvements are observed in road, sidewalk, bicycle, motorcycle, rider, fence, and traffic light. Similarly, MIC achieves enhancements in road, sidewalk, wall, pole, and bicycle. Alongside rare or small categories like the pole and bicycle, the road and sidewalk also exhibit prominent improvements, a testament to the efficacy of the road assumption in SGML. 

The third and fourth rows of Fig.~\ref{fig:pics_gta5} present the qualitative segmentation performance on SYNTHIA-to-Cityscapes. Enhanced segmentation of small classes like traffic signs and persons, with refined boundaries, is evident. Meanwhile, the rare class wall is better segmented in the third row. Moreover, it classifies the road and sidewalk in the fourth row more accurately, consistent with the qualitative results in Table~\ref{tab:syn_comparison}.

\subsubsection{Cityscapes-to-ACDC}

\begin{table*}[!ht]
        \centering
        \setlength{\tabcolsep}{3pt}
        \renewcommand{\arraystretch}{1.1}
        \captionsetup{format=myformat}
	\caption{The adaptation performance and comparison on Cityscapes-to-ACDC} 
	\label{tab:acdc_comparison}
	\resizebox{\linewidth}{!}{
	\begin{tabular}{c|c|ccccccccccccccccccc|c}
		\hline
		Methods & Network & road & sw & build & wall & fence & pole & light & sign & vege & terrain & sky & person & rider & car & truck & bus & train & motor & bike & mIoU \\ \hline
                TUFL\cite{TUFL} & BiSeNet  & 68.5  & 49.6  & 57.2  & 12.2  & 22.8  & 35.1  & 49.3  & 49.5  & 51.3  & 45.2  & 68.5  & 36.8  & 19.1  & 72.0  & 27.0  & 35.1  & 58.1  & 13.8  & 39.0 & 42.6 \\
                SePiCo\cite{sepico} & DeeplabV2 & 57.4  & 44.4  & 76.2  & 31.2  & 24.8  & 39.0  & 33.3  & 56.4  & 65.6  & 46.7  & 52.3  & 52.0  & 21.0  & 76.7  & 31.0  & 33.2  & 54.7  & 32.1  & 33.9 & 45.4 \\ \hline
                TUFL+\textbf{SAM4UDASS} & BiSeNet & 77.8  & 52.0  & 59.8  & 15.8  & 21.7  & 40.0  & 53.7  & 52.3  & 60.6  & 47.8  & 78.3  & 44.3  & 24.1  & 74.3  & 29.2  & 38.2  & 57.8  & 19.2  & 42.5 & 46.8 \\
                SePiCo+\textbf{SAM4UDASS} & DeeplabV2 & 56.0  & 45.7  & 79.5  & 38.0  & 27.5  & 41.2  & 42.7  & 57.5  & 71.6  & 48.1  & 63.0  & 55.1  & 23.6  & 79.5  & 38.6  & 37.4  & 56.6  & 33.8  & 42.7 & 49.4 \\ \hline
                DAFormer\cite{daformer} & \multirow{5}{*}{\begin{tabular}{c} DAFormer \end{tabular}} & 58.4  & 51.3  & 84.0  & 42.7  & 35.1  & 50.7  & 30.0  & 57.0  & 74.8  & 52.8  & 51.3  & 58.3  & 32.6  & 82.7  & 58.3  & 54.9  & 82.4  & 44.1  & 50.7  & 55.4 \\
                HRDA\cite{hrda} &  & 88.8  & 60.6  & 87.2  & 56.7  & 38.8  & 53.7  & {62.0} & 61.5  & 74.4  & 59.5  & 86.2  & 67.7  & \textbf{49.2} & 88.4  & 74.8  & 75.3  & 86.5  & 54.7  & 61.3 & 67.8 \\
		MIC\cite{mic} & & {89.2} & {64.1} & 88.4  & \textbf{59.5} & \textbf{43.1} & \textbf{57.8}  & 48.2  & {68.1} & 76.1  & \textbf{61.2} & 85.7  & 70.5  & 46.1  & \textbf{89.6} & \textbf{76.5} & {82.4}  & 89.0  & \textbf{55.5} & \textbf{62.6} & 69.1 \\ \cline{1-1} \cline{3-22}
                DAFormer+\textbf{SAM4UDASS} &  & 79.4  & 42.0  & 85.8  & 45.6  & 35.0  & 52.2  & 49.2  & 58.3  & 75.8  & 51.5  & 86.2  & 61.4  & 35.0  & 83.2  & 62.2  & 64.6  & 83.6  & 45.2  & 53.0 & 60.5 \\
                MIC+\textbf{SAM4UDASS} &  & \textbf{91.2}  & \textbf{67.5}  & \textbf{89.1}  & 54.0  & 40.5  & 57.5  & \textbf{63.7}  & \textbf{69.3}  & \textbf{78.9}  & 61.0  & \textbf{90.7}  & \textbf{71.5}  & 46.3  & \textbf{89.6}  & 74.1  & \textbf{86.1}  & \textbf{89.1}  & 54.6  & 61.9 & \textbf{70.3}\\ \hline
	\end{tabular}}
\end{table*}
Table~\ref{tab:acdc_comparison} presents the performance of SAM4UDASS on Cityscapes-to-ACDC. TUFL, SePiCo, DAFormer and MIC gain 4.2\%, 4.0\%, 5.1\% and 1.2\% mIoU improvement.  Notably, when $\hat{y}_{uda}$ exhibits low (TUFL, SePiCo) or moderate (DAFormer) precision, SAM masks effectively refine pseudo-labels. Conversely, when the precision of $\hat{y}_{uda}$ is high (MIC), SAM4UDASS's benefits are less pronounced, akin to the synthetic-to-real results. The qualitative results are shown in the last two rows of Fig~\ref{fig:pics_gta5}, further highlighting the enhancement brought by SAM4UDASS.

\subsection{Ablation Study}
\subsubsection{Grounded SAM vs. Majority Voting vs. SGML}

\begin{figure*}[htbp]
        \centering
	\captionsetup[subfloat]{font=scriptsize,labelfont=scriptsize,labelformat=empty}
        \vspace{-0.5cm}
	\subfloat{\includegraphics[width=0.166\linewidth]{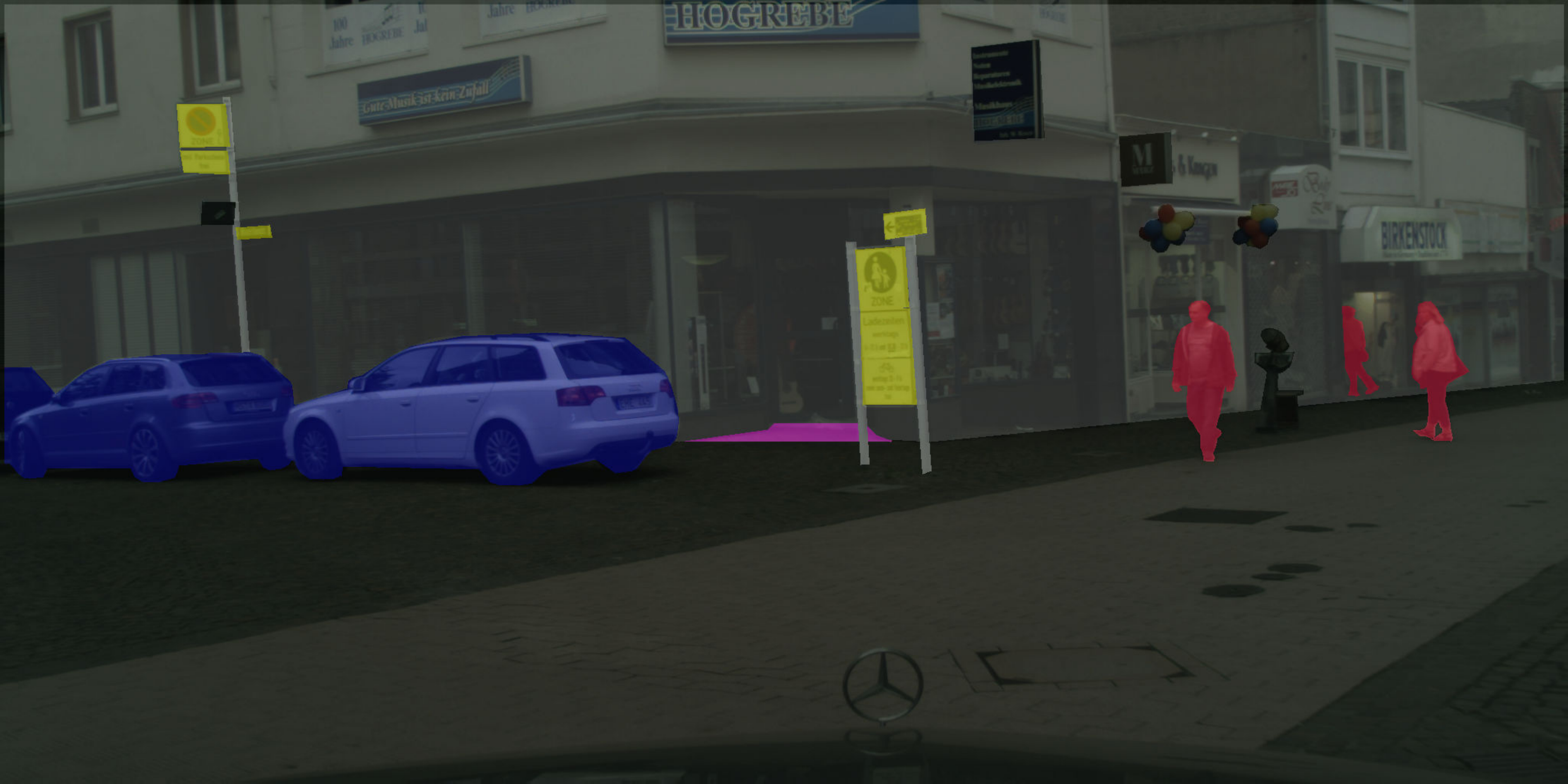}} \hfill
	\subfloat{\includegraphics[width=0.166\linewidth]{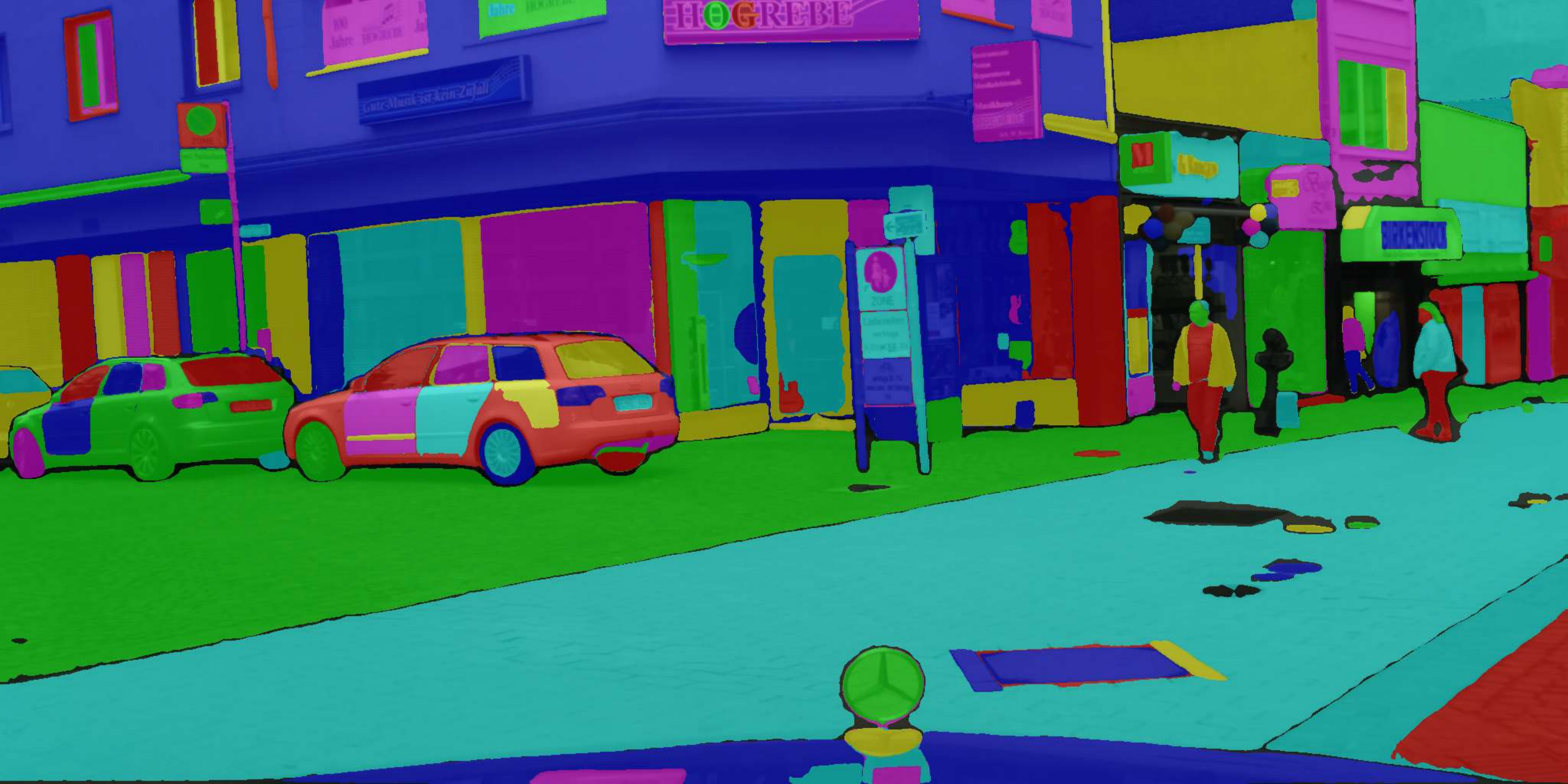}} \hfill
	\subfloat{\includegraphics[width=0.166\linewidth]{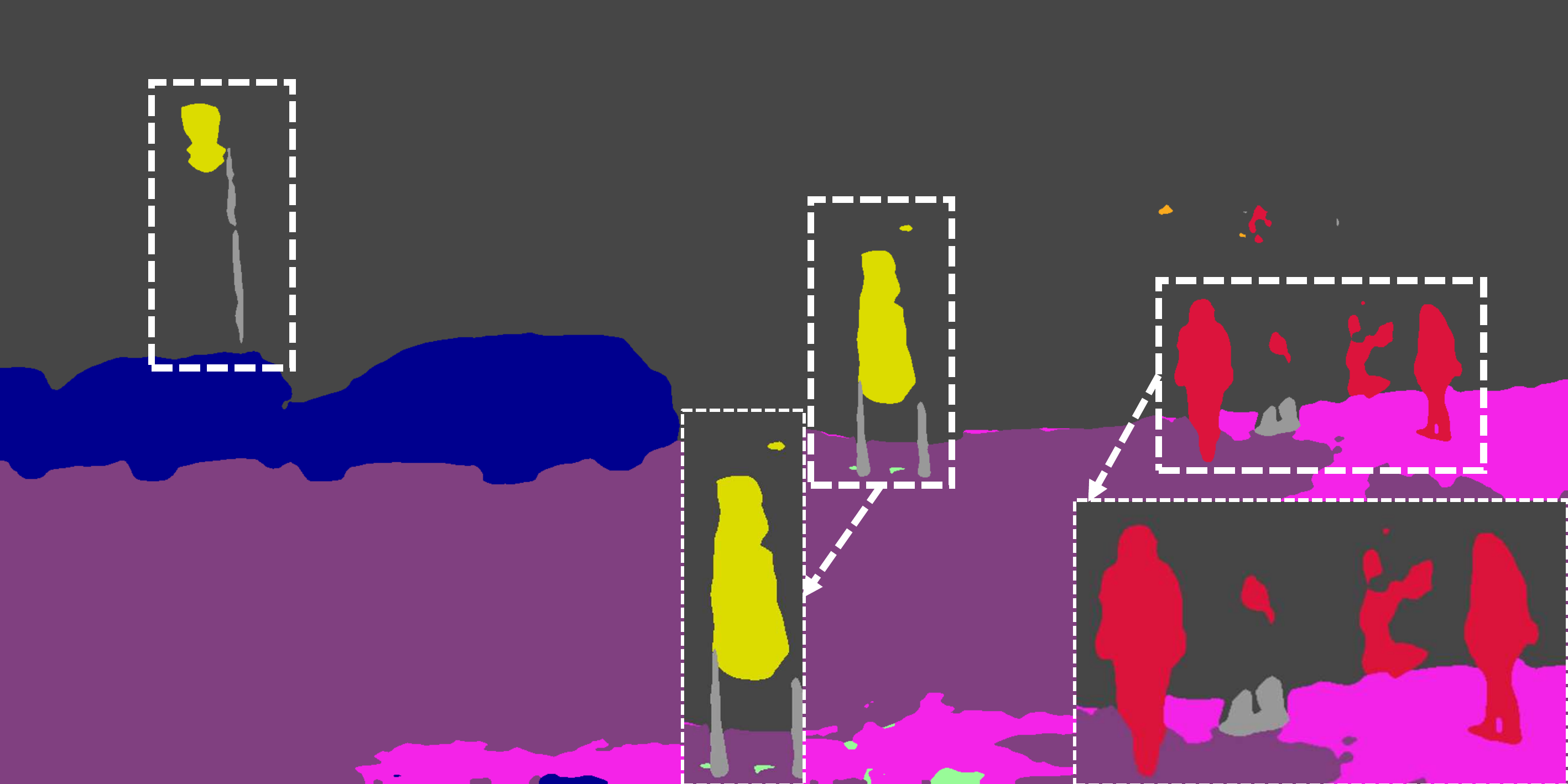}} \hfill
	\subfloat{\includegraphics[width=0.166\linewidth]{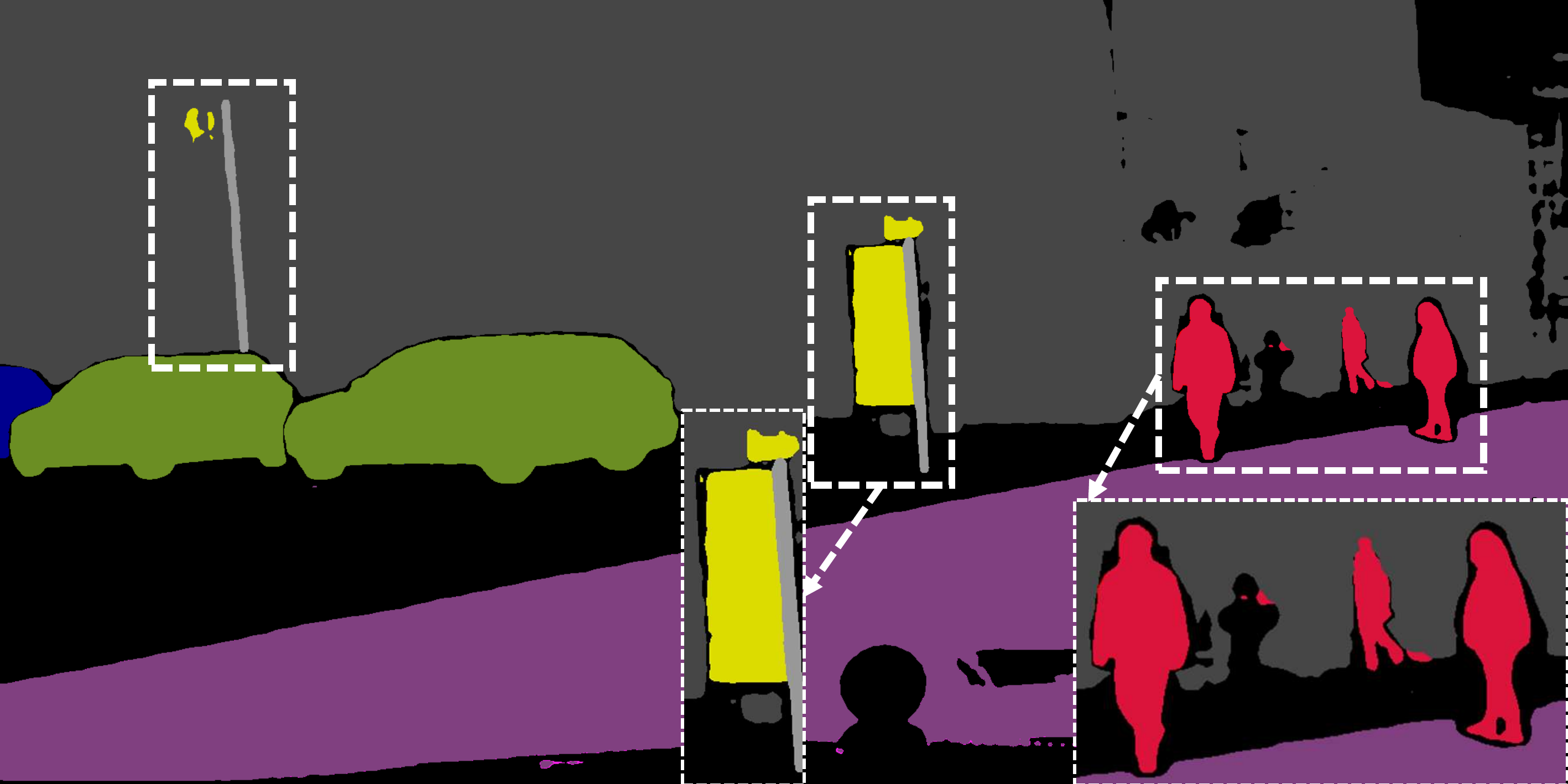}} \hfill
	\subfloat{\includegraphics[width=0.166\linewidth]{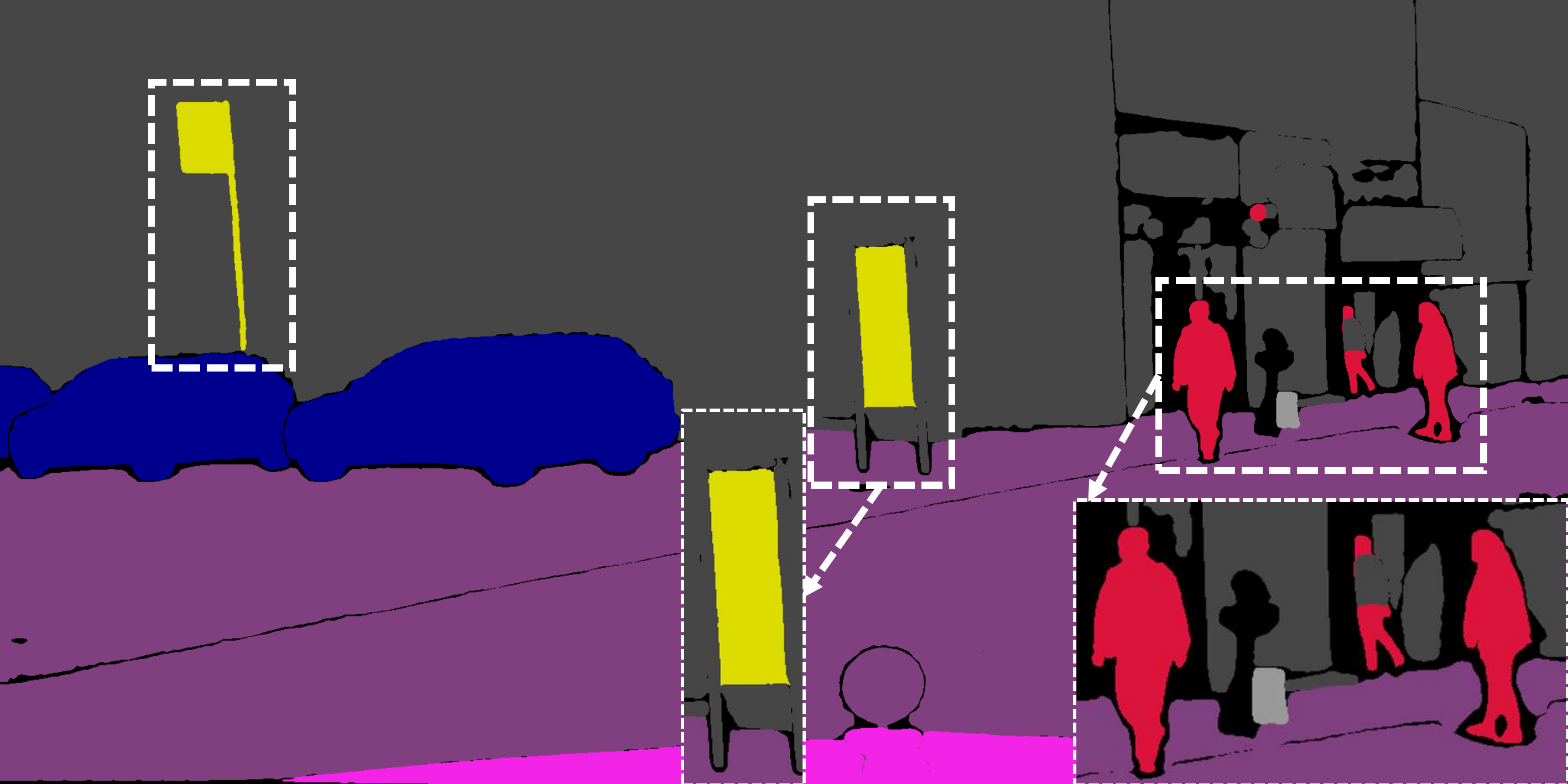}} \hfill
        \subfloat{\includegraphics[width=0.166\linewidth]{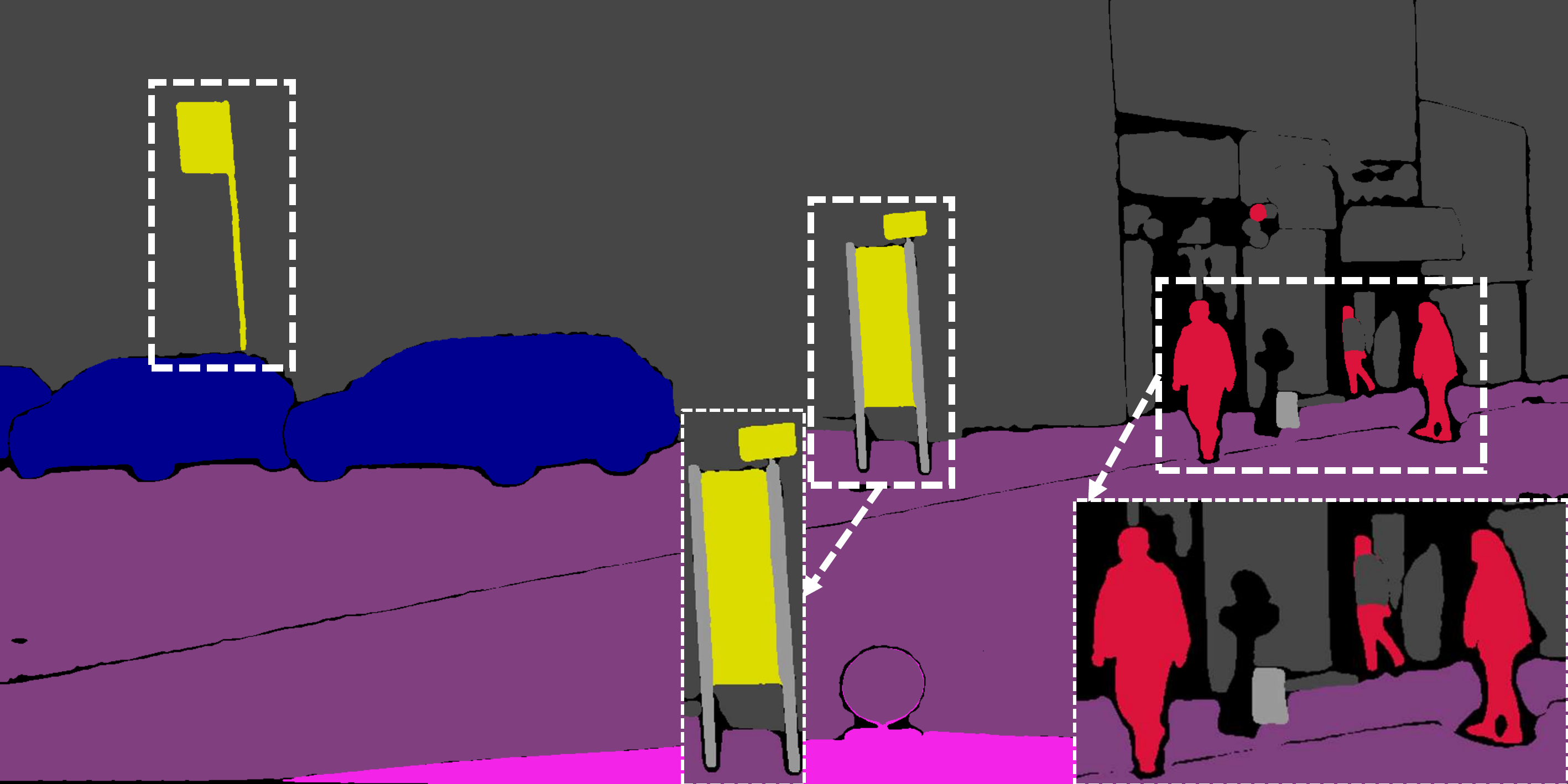}} \\ \vspace{-0.30cm}
        \subfloat[Image with ground truth]{\includegraphics[width=0.166\linewidth]{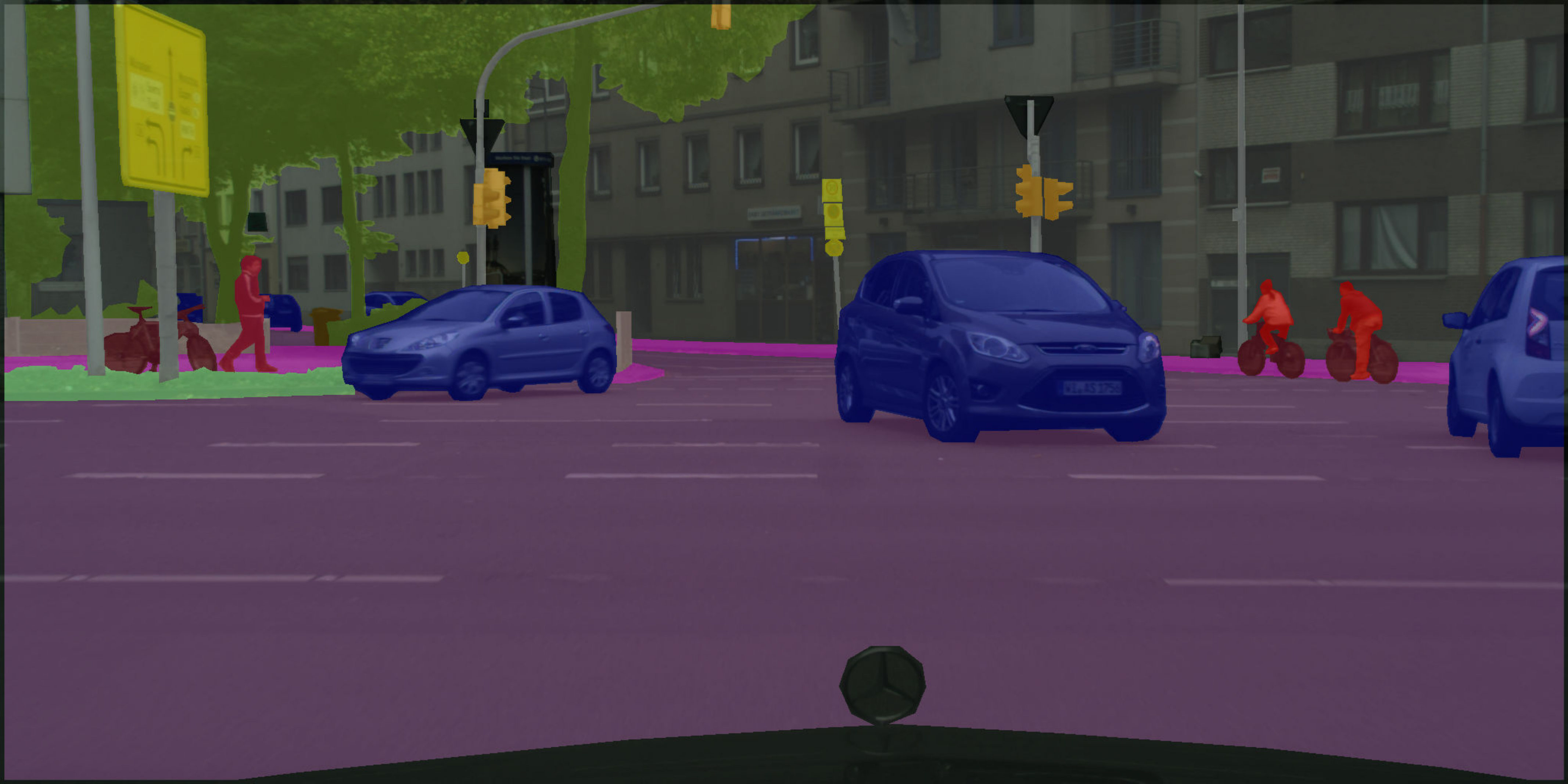}} \hfill
	\subfloat[SAM Masks]{\includegraphics[width=0.166\linewidth]{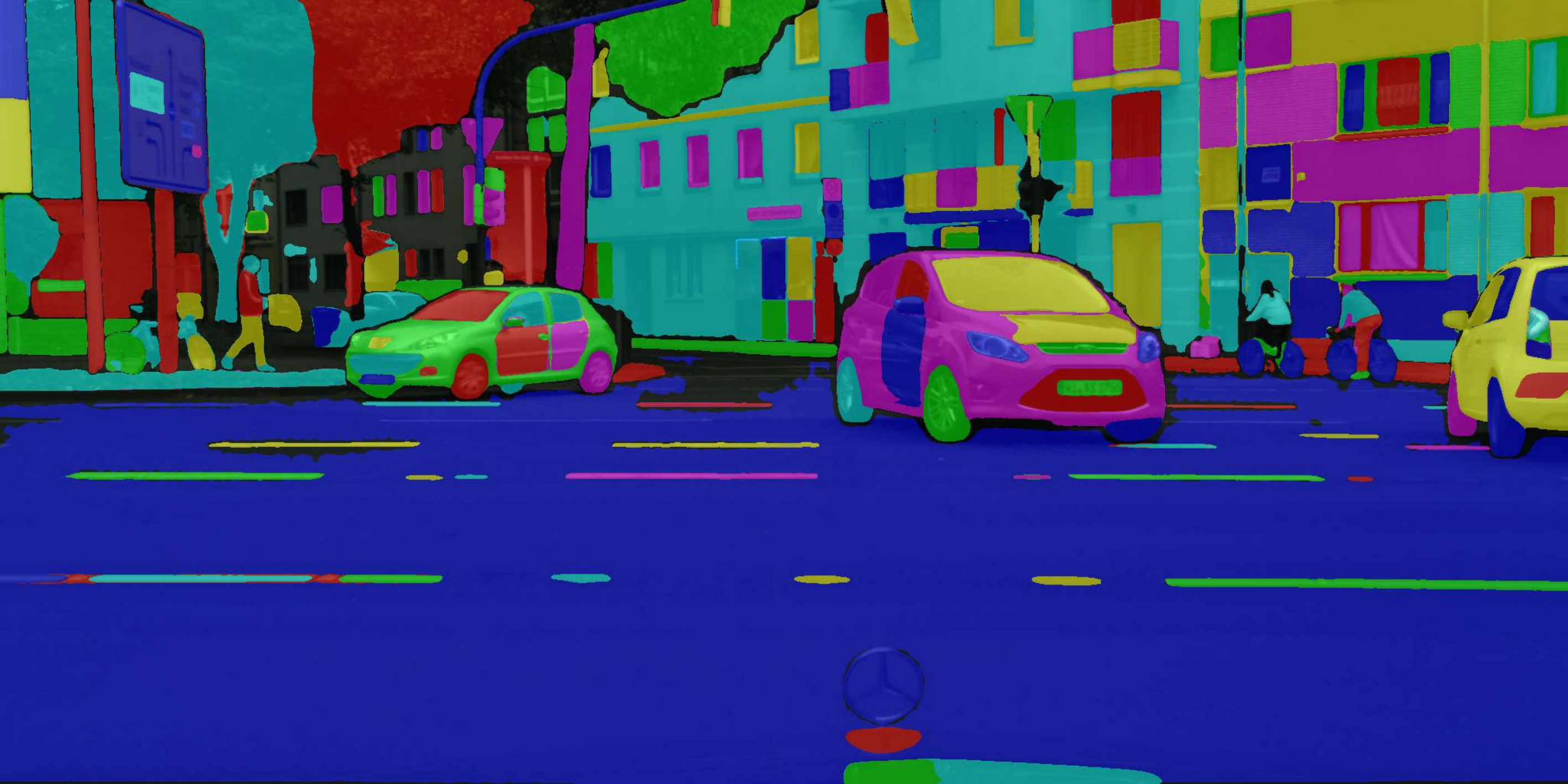}} \hfill
	\subfloat[UDA Pseudo-label]{\includegraphics[width=0.166\linewidth]{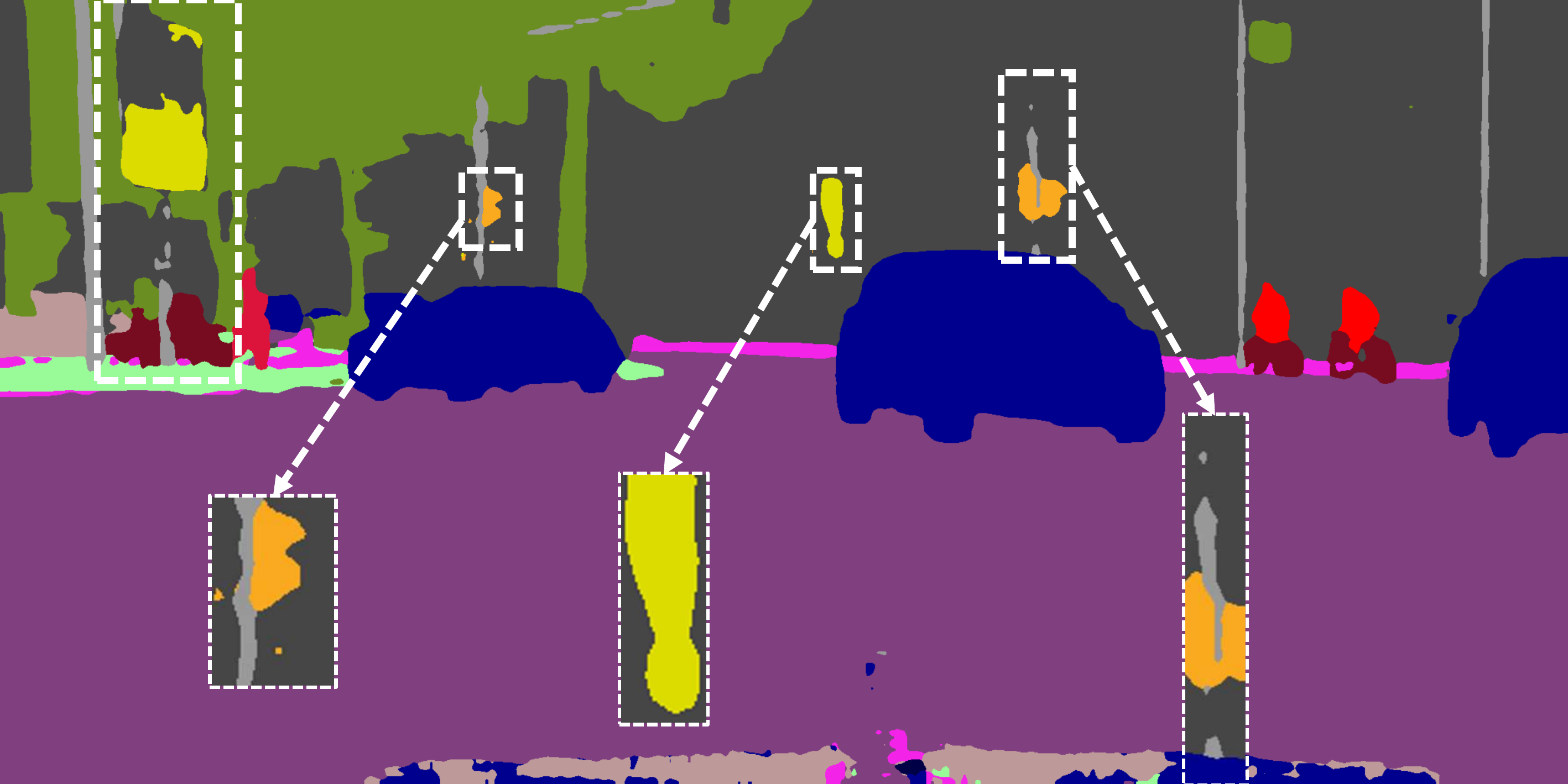}} \hfill
	\subfloat[Grounded SAM]{\includegraphics[width=0.166\linewidth]{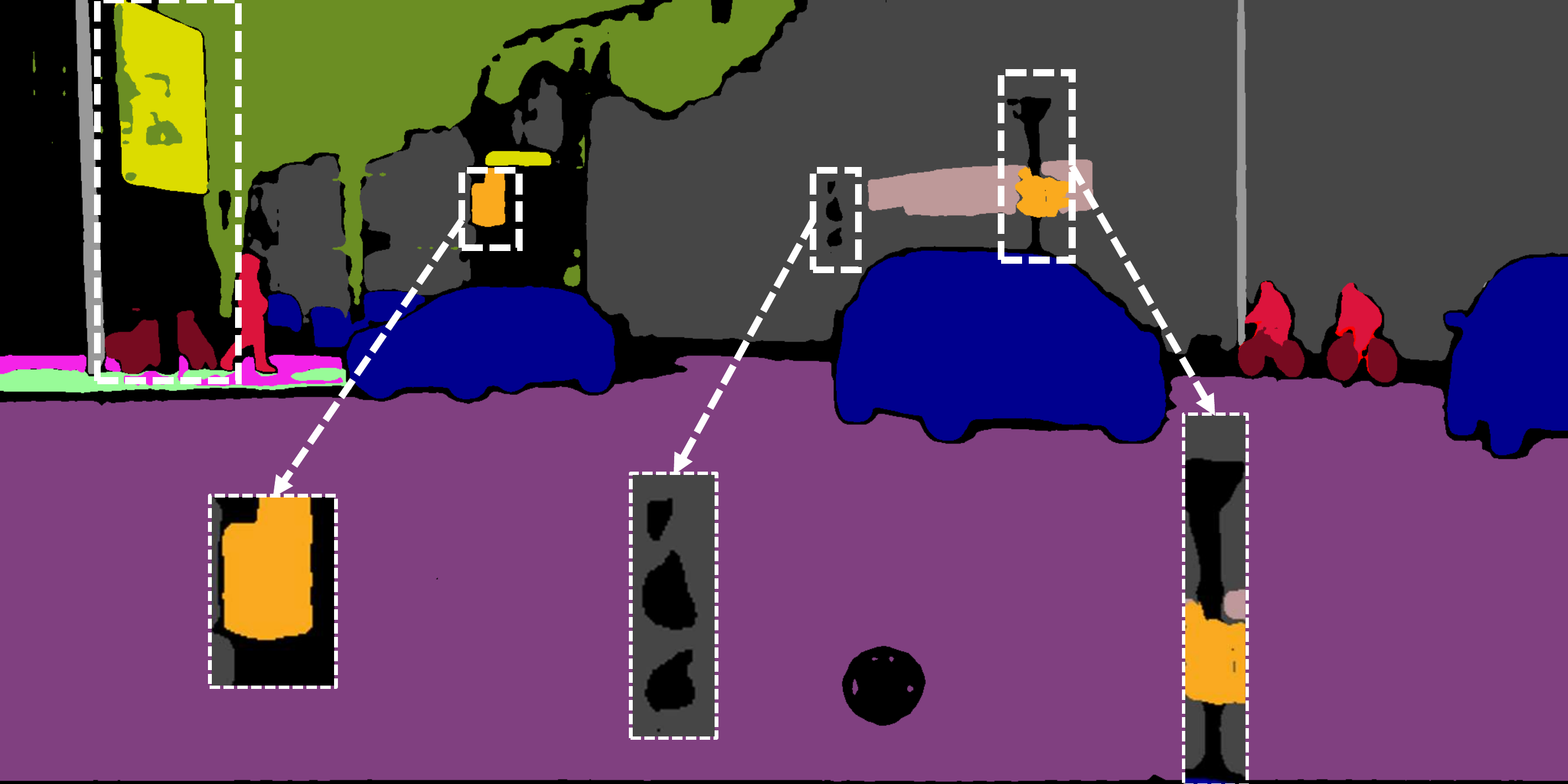}} \hfill
	\subfloat[Majority Voting]{\includegraphics[width=0.166\linewidth]{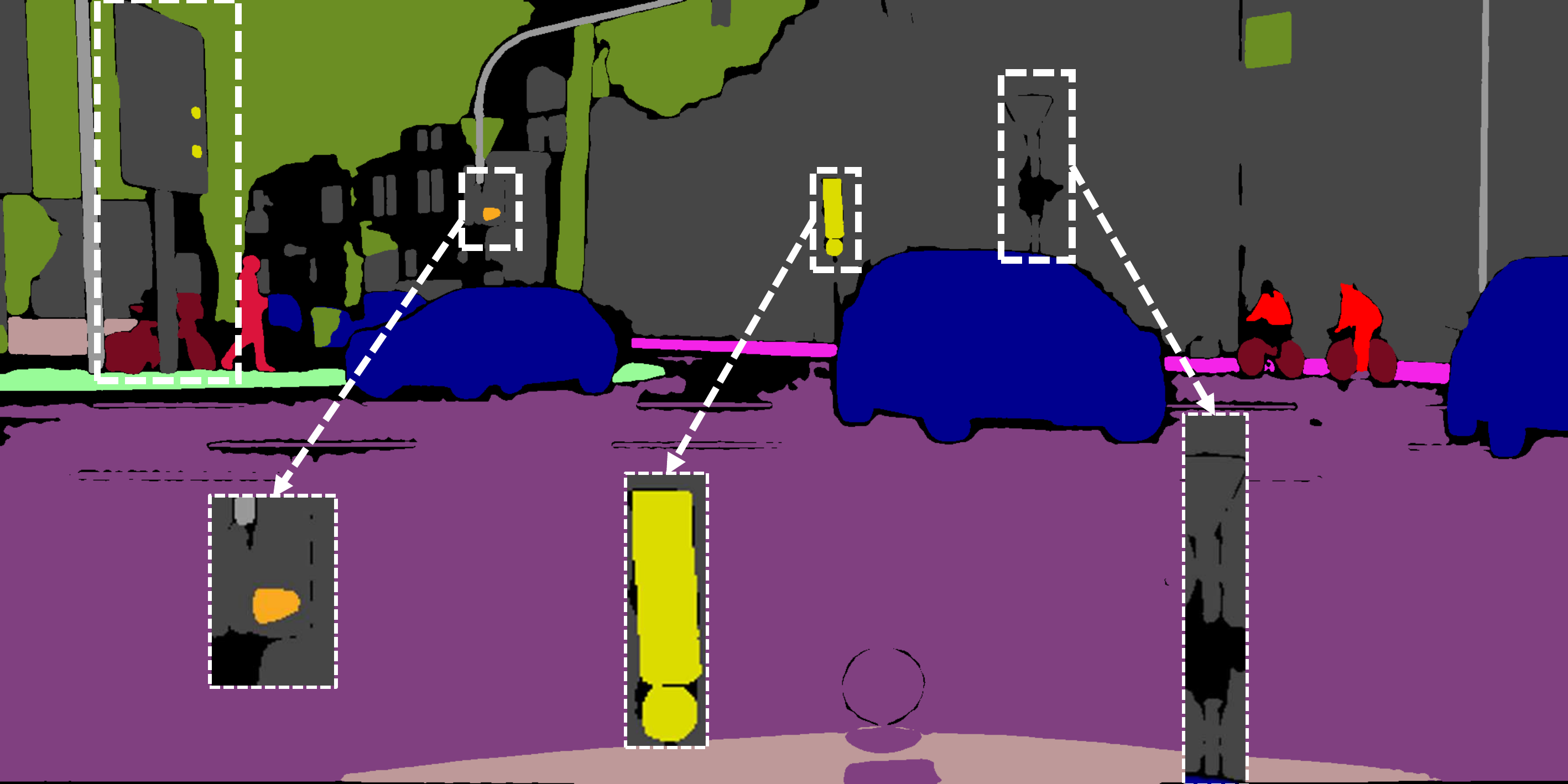}} \hfill
        \subfloat[SGML]{\includegraphics[width=0.166\linewidth]{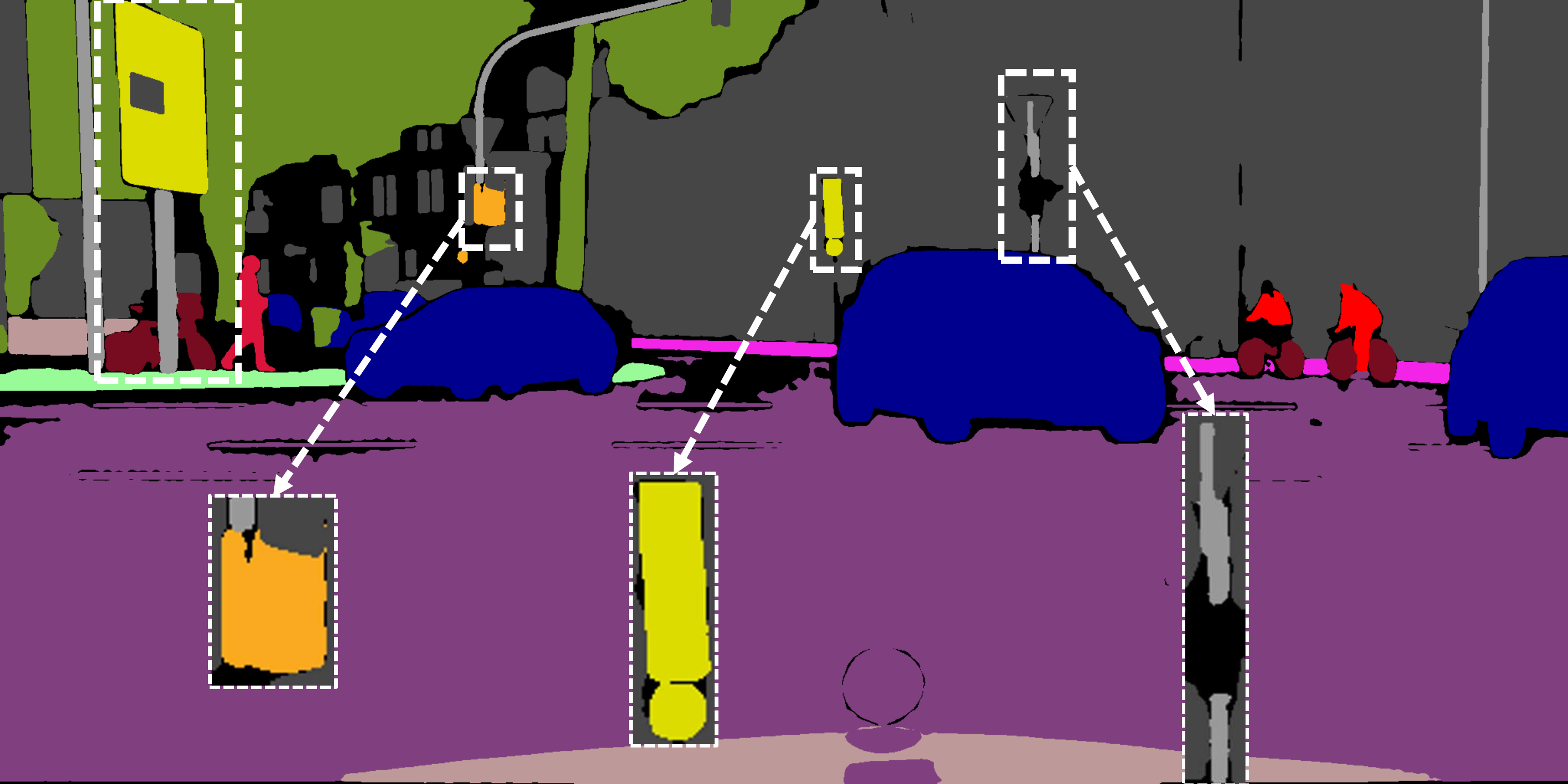}}
        \caption{The performance of different methods in assigning semantic categories to unlabeled SAM masks. From left to right: the image mixed with ground truth, SAM masks, UDA pseudo-label, and the results of Grounded SAM, Majority Voting, and SGML, respectively. Certain small objects have been magnified to enhance visibility.}
        \vspace{-0.2cm}
        \label{fig:get_sam}
\end{figure*}

A comparison is made between Grounded SAM\cite{dino}, Majority Voting\cite{ssm}, and SGML on GTA5-to-Cityscapes using DAFormer. Grounded SAM employs the 19 class names from Cityscapes as prompts. The qualitative results are exhibited in Fig.~\ref{fig:get_sam}. Grounded SAM's performance is compromised on background classes due to its heavy reliance on DINO. For instance, sidewalk and building classes demonstrate inadequacies, resulting in void regions (black area). In some cases, cars are even misclassified as vegetation as shown in row 1, column 4. Benefiting from semantic and area ratio information, SGML can assign semantic labels to SAM masks more appropriately on some small classes than Majority Voting, like the pole in the first row and the traffic sign in the second row. Table~\ref{tab:sgml} shows the quantitative results of different methods when using Fusion Strategy 1, which takes $\hat{y}_{sam}$ as the base and can better show the quality of SAM pseudo-label. It should be noted that Table~\ref{tab:sgml} only shows the results of the $C_s$ classes, road, and sidewalk that refined by SGML. Consistent with the qualitative results, SGML outperforms Majority Voting on some small or rare classes such as the wall, pole, traffic sign, terrain and bike. Although Grounded SAM performs better on some foreground categories like the traffic sign and bike that DINO can detect well, it lags behind SGML in background categories like road, sidewalk, wall, and terrain.

\begin{table}[htbp]
        \centering
        \setlength{\tabcolsep}{2pt}
        \captionsetup{format=myformat}
        \caption{Comparison of different methods
        for assigning SAM unlabeled masks using Fusion Strategy 1}
        \label{tab:sgml}
        \begin{tabular}{c|ccccccc} \hline
                mIoU(\%) & raod & sidewalk & wall & pole  & sign & terrain & bike \\ \hline
                Grounded SAM\cite{dino} & 94.16 & 61.09 & 31.08 & 50.03  & \textbf{63.06} & 36.85 & \textbf{69.62} \\
                Majority Voting\cite{ssm} & 94.45 & 66.55 & 48.40 & 49.09  & 60.50 & 57.19 & 63.48 \\
                SGML (ours) & \textbf{95.20} & \textbf{69.22} & \textbf{49.01} & \textbf{50.37} & 61.96 & \textbf{57.92} & 64.26 \\ \hline
        \end{tabular}
        \vspace{-0.3cm}
\end{table}

\subsubsection{Three Fusion Strategies}
\begin{table}[htbp]
        \centering
        \setlength{\tabcolsep}{3pt}
        \captionsetup{format=myformat}
        \caption{The performance of three fusion strategies on different datasets based on DAFormer}
        \label{tab:fusion_strategy}
        \begin{tabular}{c|cccc}
                \hline
                mIoU(\%) & $\hat{y}_{uda}$ & Strategy 1 & Strategy 2 & Strategy 3 \\ \hline
                GTA5-to-Cityscapes & 68.28 & 70.79 & 70.39 & \textbf{71.09} \\
                SYNTHIA-to-Cityscapes & 60.49 & 64.11 & 62.02 & \textbf{64.20} \\ 
                Cityscapes-to-ACDC & 56.73 & \textbf{62.07} & 57.88 & 61.87 \\ \hline
        \end{tabular}
\end{table}

%
The performance of three fusion strategies is compared on the Cityscapes and ACDC training sets, shown in Table~\ref{tab:fusion_strategy}. Strategy 3 performs optimally on GTA5-to-Cityscapes and SYNTHIA-to-Cityscapes, while Strategy 1 performs best on Cityscapes-to-ACDC. Strategy 1 can better leverage the masks generated by SAM than Strategy 2, resulting in enhanced performance. Meanwhile, Strategy 3, building upon Strategy 1, further employs the confidence of $\hat{y}_{uda}$ to refine the results. As the quality of $\hat{y}_{uda}$ improves, Strategy 3 gradually demonstrates its superiority.

\subsubsection{The area ratio $\alpha$} 
\begin{figure}[htbp]
        \centering
        \captionsetup[subfloat]{font=scriptsize,labelfont=scriptsize}
        \vspace{-0.4cm}
        \subfloat{\includegraphics[width=0.5\linewidth]{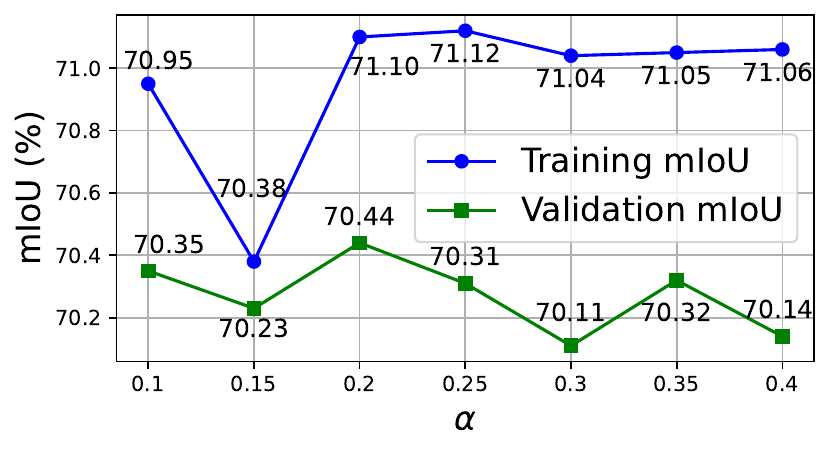}}
        \subfloat{\includegraphics[width=0.5\linewidth]{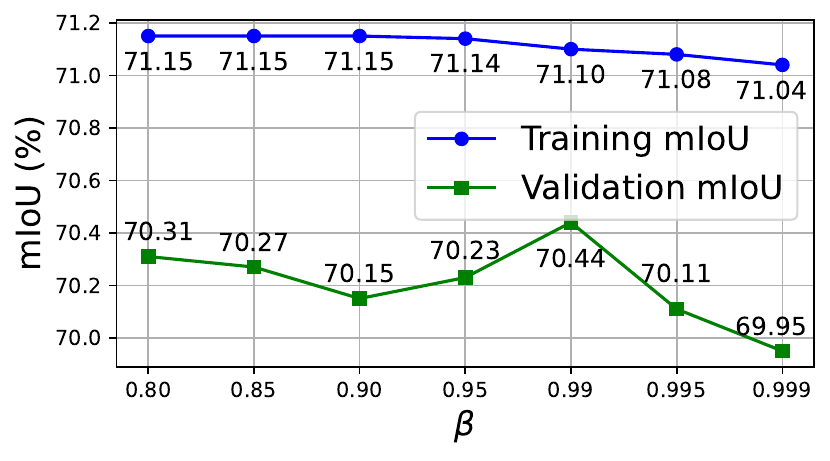}}
        \caption{The influence of area ratio $\alpha$ (left) and confidence threshold $\beta$ (right) on GTA5-to-Cityscapes using DAFormer.}
        \label{fig:alpha_ablation}
\end{figure}

The impact of the area ratio $\alpha$ is explored using DAFormer on GTA5-to-Cityscapes, where $\alpha$ is varied as 0.1, 0.15, 0.2, 0.25, 0.3, 0.35, and 0.4. As illustrated in Fig~\ref{fig:alpha_ablation}, SGML's performance remains relatively insensitive to $\alpha$. The optimal performance is achieved at $\alpha$ values of 0.25 and 0.2 on the training and validation sets, respectively. In our experiments, we adopt $\alpha=0.2$.

\subsubsection{The confidence threshold $\beta$} 


Fig.~\ref{fig:alpha_ablation} depicts the sensitivity of Fusion Strategy 3 to the confidence threshold $\beta$ using DAFormer on GTA5-to-Cityscapes, where $\beta$ is set as 0.8, 0.85, 0.9, 0.95, 0.99, 0.995, and 0.999. The performance of Fusion Strategy 3 remains robust within the range of 0.80 to 0.99, with a smooth transition in mIoU. Performance decreases as $\beta$ exceeds 0.99, primarily due to significantly reduced pixel selection from $\hat{y}_{uda}$, which diminishes its corrective influence on ${y}_t^{(1)}$ when $\beta$ is set too high.

\subsubsection{Run time experiments}
\begin{figure}
        \centering
        \captionsetup[subfloat]{font=scriptsize,labelfont=scriptsize}
        \vspace{-0.4cm}
        \subfloat{\includegraphics[width=0.5\linewidth]{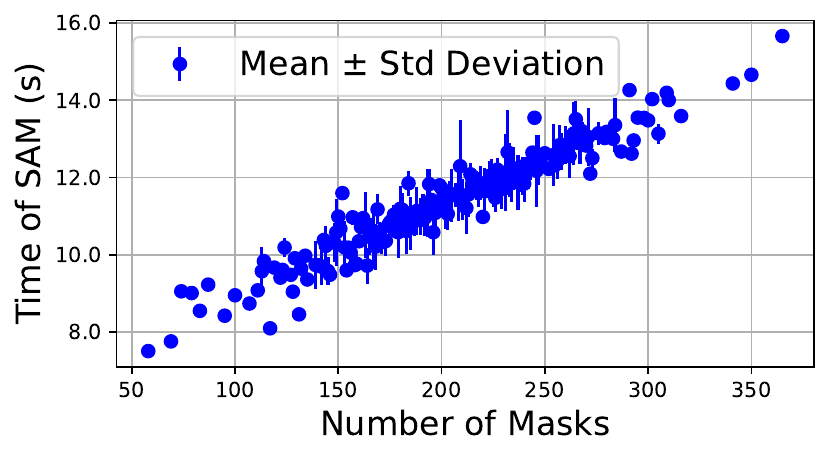}}
        \subfloat{\includegraphics[width=0.5\linewidth]{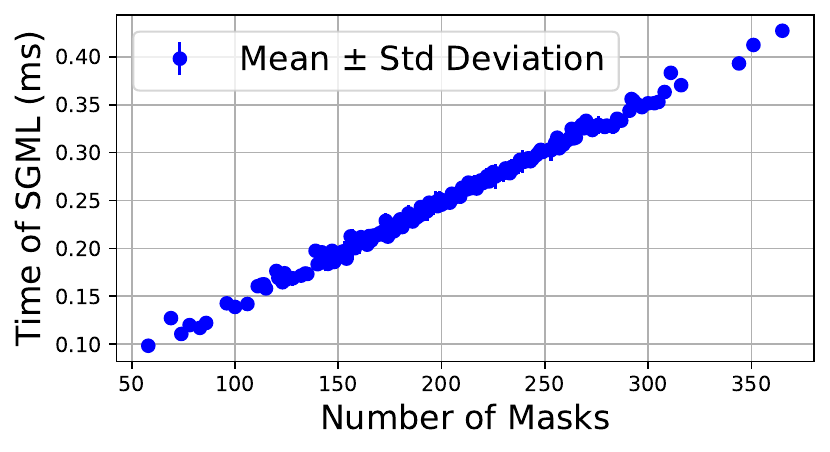}}
        \caption{The correlation between processing time and the number of masks for SAM (left) and SGML (right).}
        \vspace{-0.3cm}
        \label{fig:sam_time}
\end{figure}

We conducted runtime experiments using 500 images from the Cityscapes training set. Each runtime experiment was independently repeated three times to ensure accuracy, and the resulting data was averaged, as detailed in Table~\ref{tab:time}. On average, the processing time per image for SAM was measured at 11.4 s, while SGML required 253.5 ms. Their processing times are positively correlated with the number of masks, as shown in Fig~\ref{fig:sam_time}. Three fusion strategies exhibited processing times of 7.8 ms, 27.6 ms, and 29.2 ms, respectively. The cumulative runtime, which includes SAM, SGML, and Fusion Strategy 3, equates to 11.7 seconds per image. Nevertheless, SAM masks for target domain images can be generated offline, with an estimated time of around 8 hours to produce SAM masks for the complete Cityscapes training set. In this context, the processing time per image would be approximately 282.7 ms.

\begin{table}[htbp]
        \centering
        \setlength{\tabcolsep}{3pt}
        \captionsetup{format=myformat}
        \caption{Average run times of SAM, SGML, and three fusion strategies on 500 images from Cityscapes}
        \label{tab:time}
        \begin{tabular}{ccccc|c}
                \hline
                SAM & SGML & Strategy 1 & Strategy 2 & Strategy 3 & Time (ms) \\ \hline
                $\surd$ & & & & & $11415.9$ \\ 
                & $\surd$ & & & & $253.5$ \\ 
                & & $\surd$ & & & $7.8$ \\ 
                & & & $\surd$ & & $27.6$  \\ 
                & & & & $\surd$ & $29.2$  \\ \hline 
                & $\surd$ & & & $\surd$ & $282.7$ \\
                $\surd$ & $\surd$ & & & $\surd$ & $11698.6$ \\ \hline
        \end{tabular}
        \vspace{-0.2cm}
\end{table}

\section{Conclusion} 
In this paper, we propose SAM4UDASS, a novel framework for enhancing self-training UDA methods in cross-domain driving scenarios, which integrates the Segment Anything Model into UDASS for the first time. SGML is designed to assign semantic categories to unlabeled SAM masks, enhancing the quality of pseudo-labels for rare or small objects. Three simple yet effective fusion strategies are developed to combine the pseudo-labels from self-training methods and SAM for refined ones, which mitigates the semantic granularity inconsistency between SAM masks and the target domain. Extensive experiments on synthetic-to-real and normal-to-adverse driving datasets demonstrate that SAM4UDASS can consistently improve the performance of self-training methods.

This paper provides a pioneering attempt to use SAM for UDASS in driving scenes. However, it only utilizes SAM's whole-image segmentation results, and the prompt-based results have not achieved satisfactory performance. Meanwhile, further optimizing SAM4UDASS's runtime is possible and needed. These aspects will be our future work.

\bibliographystyle{IEEEtran}
\bibliography{IEEEabrv,reference.bib}
\vspace{-14pt}
\begin{IEEEbiography}[{\includegraphics[width=1in,height=1.2in,clip,keepaspectratio]{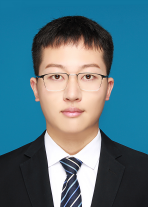}}]{\textbf{Weihao Yan}}
        received the B.S. degree in automation from Shanghai Jiao Tong University, Shanghai, China, in 2020. He is currently working toward the Ph.D. degree in Control Science and Engineering with Shanghai Jiao Tong University.\\
        His main research interests include computer vision, scene segmentation, domain adaptation and their applications in intelligent transportation systems.
\end{IEEEbiography}
\vspace{-13.5 mm}
\begin{IEEEbiography}[{\includegraphics[width=1in,height=1.2in,clip,keepaspectratio]{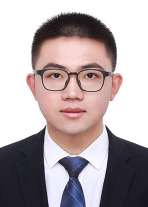}}]{\textbf{Yeqiang Qian}}
        received a Ph.D. degree in control science and engineering from Shanghai Jiao Tong University, Shanghai, China, in 2020. He is currently a postdoctoral fellow at the Global Institute of Future Technology at Shanghai Jiao Tong University. \\ 
        His main research interests include computer vision, pattern recognition, machine learning, and their applications in intelligent transportation systems.
\end{IEEEbiography}
\vspace{-13.5 mm}
\begin{IEEEbiography}[{\includegraphics[width=1in,height=1.2in,clip,keepaspectratio]{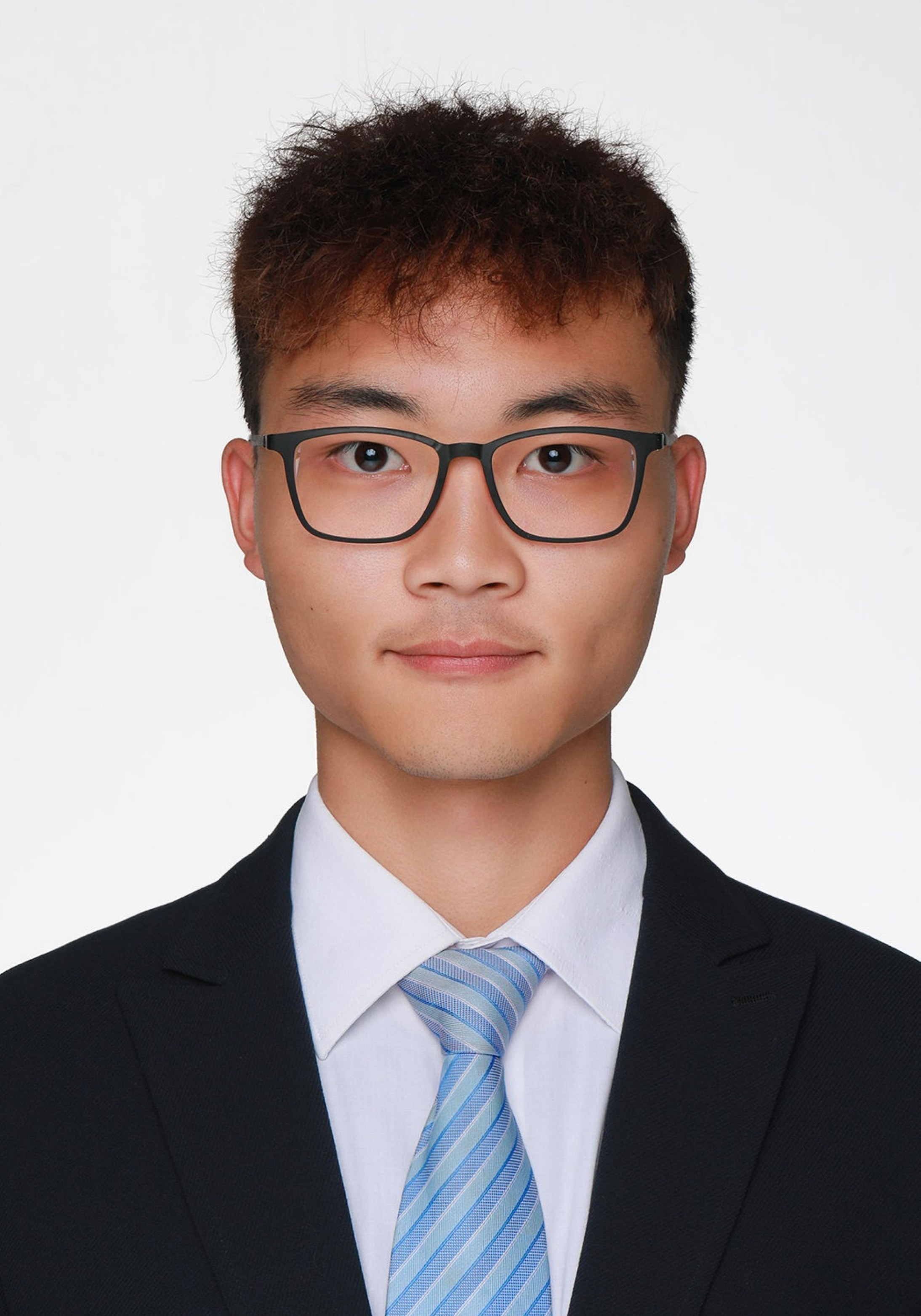}}]{\textbf{Xingyuan Chen}}
        received the B.S. degree in Artificial Intelligence from Shanghai Jiao Tong University, Shanghai, China, in 2023. He is currently working towards the Master. degree in Control Science and Engineering with Joint-Institution of Shanghai Jiaotong University.\\
        His main research interests include Computer Vision, Reinforcement Learning and their applications in autonomous vehicle research.
\end{IEEEbiography}
\vspace{-13.5 mm}
\begin{IEEEbiography}[{\includegraphics[width=1in,height=1.2in,clip,keepaspectratio]{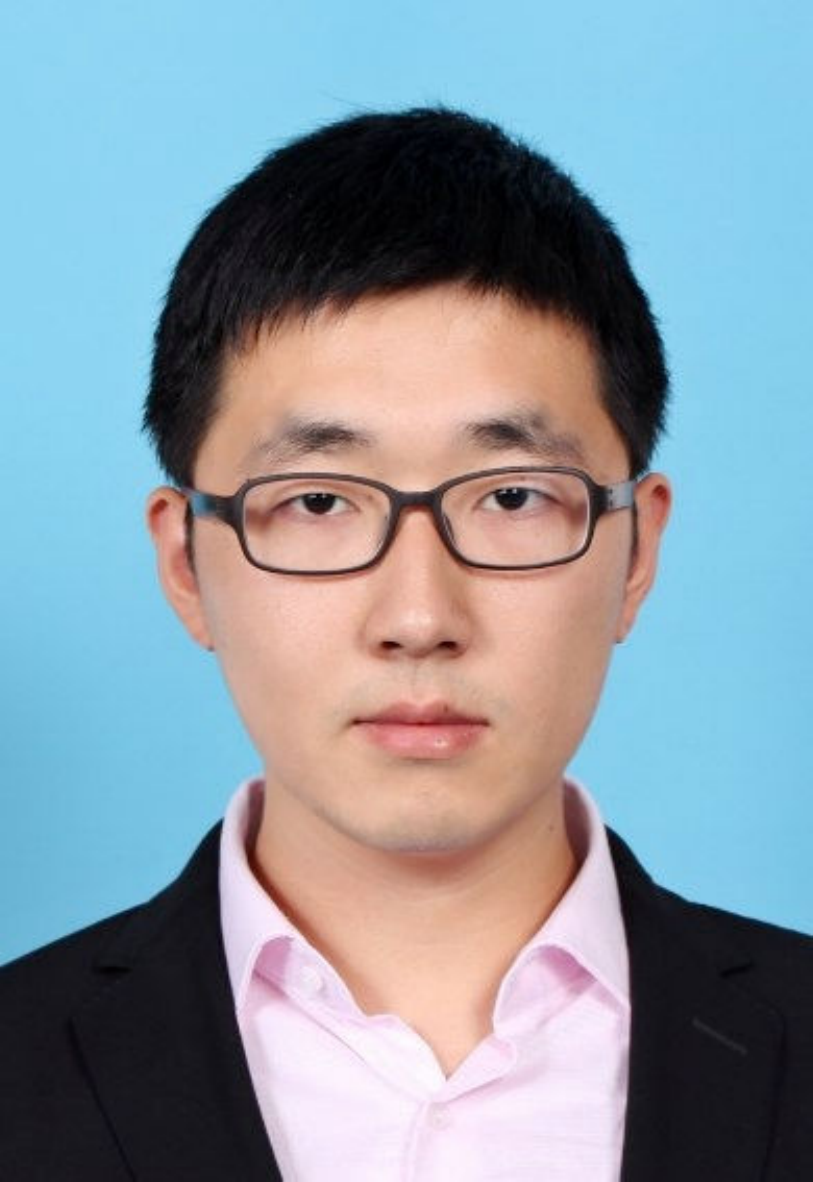}}]{\textbf{Hanyang Zhuang}}
        received the Ph.D. degree from Shanghai Jiao Tong University, Shanghai, China, in 2018. 
        He worked as a postdoctoral reseracher at Shanghai Jiao Tong University from 2020 to 2022. He is currently an assistant research professor at Shanghai Jiao Tong University implementing research works related to intelligent vehicles.\\ 
        His research focus is on AD/ADAS system design, high-precision localization, environment perception, and cooperative driving.
\end{IEEEbiography}
\vspace{-13.5 mm}
\begin{IEEEbiography}[{\includegraphics[width=1in,height=1.2in,clip,keepaspectratio]{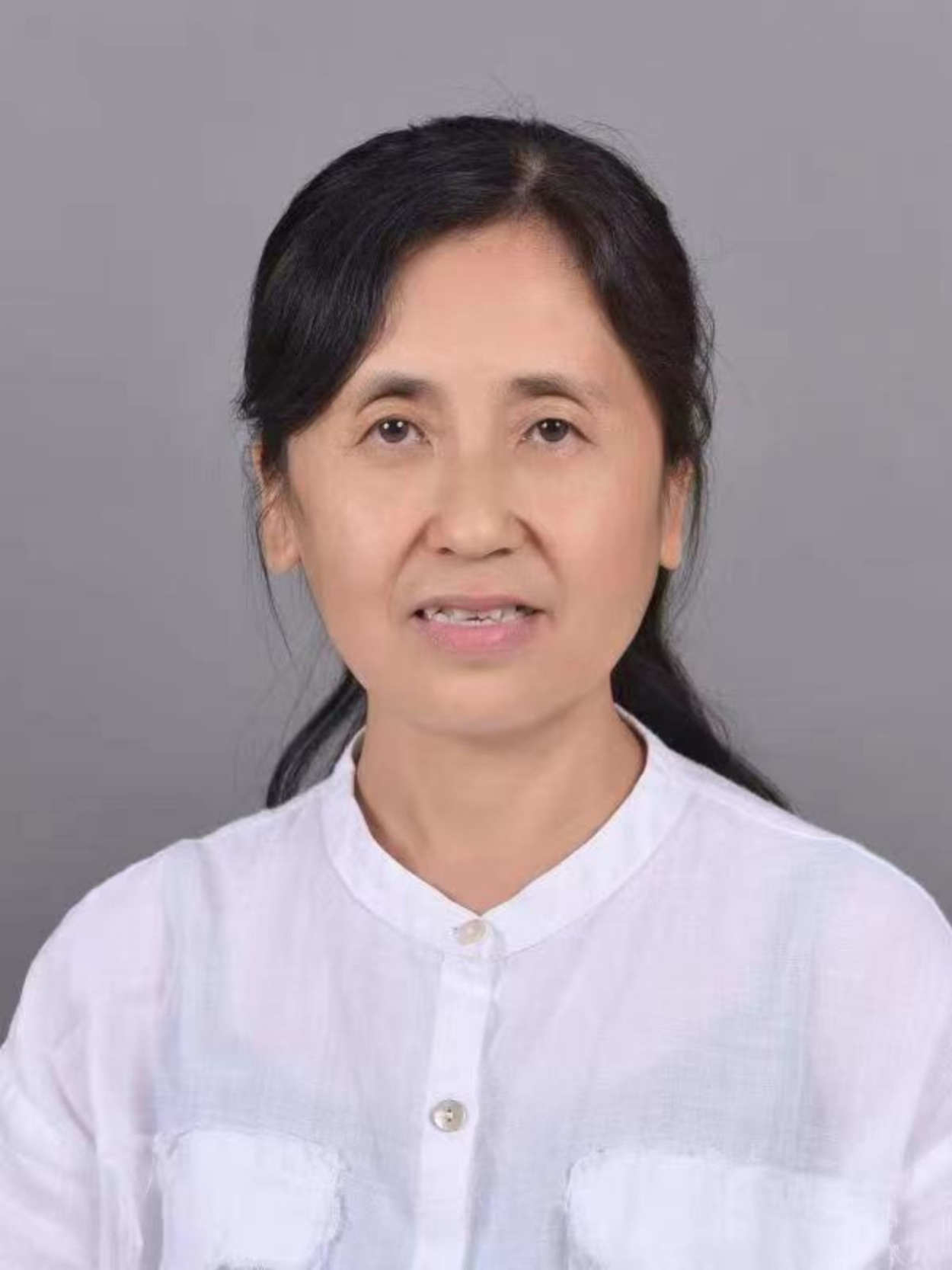}}]{\textbf{Chunxiang Wang}}
received the Ph.D. degree in mechanical engineering from the Harbin Institute of Technology, Harbin, China, in 1999.\\
She is currently an Associate Professor with the Department of Automation at Shanghai Jiao Tong University, Shanghai, China. Her research interests include robotic technology and electromechanical integration.
\end{IEEEbiography}
\vspace{-13.5 mm}
\begin{IEEEbiography}[{\includegraphics[width=1in,height=1.2in,clip,keepaspectratio]{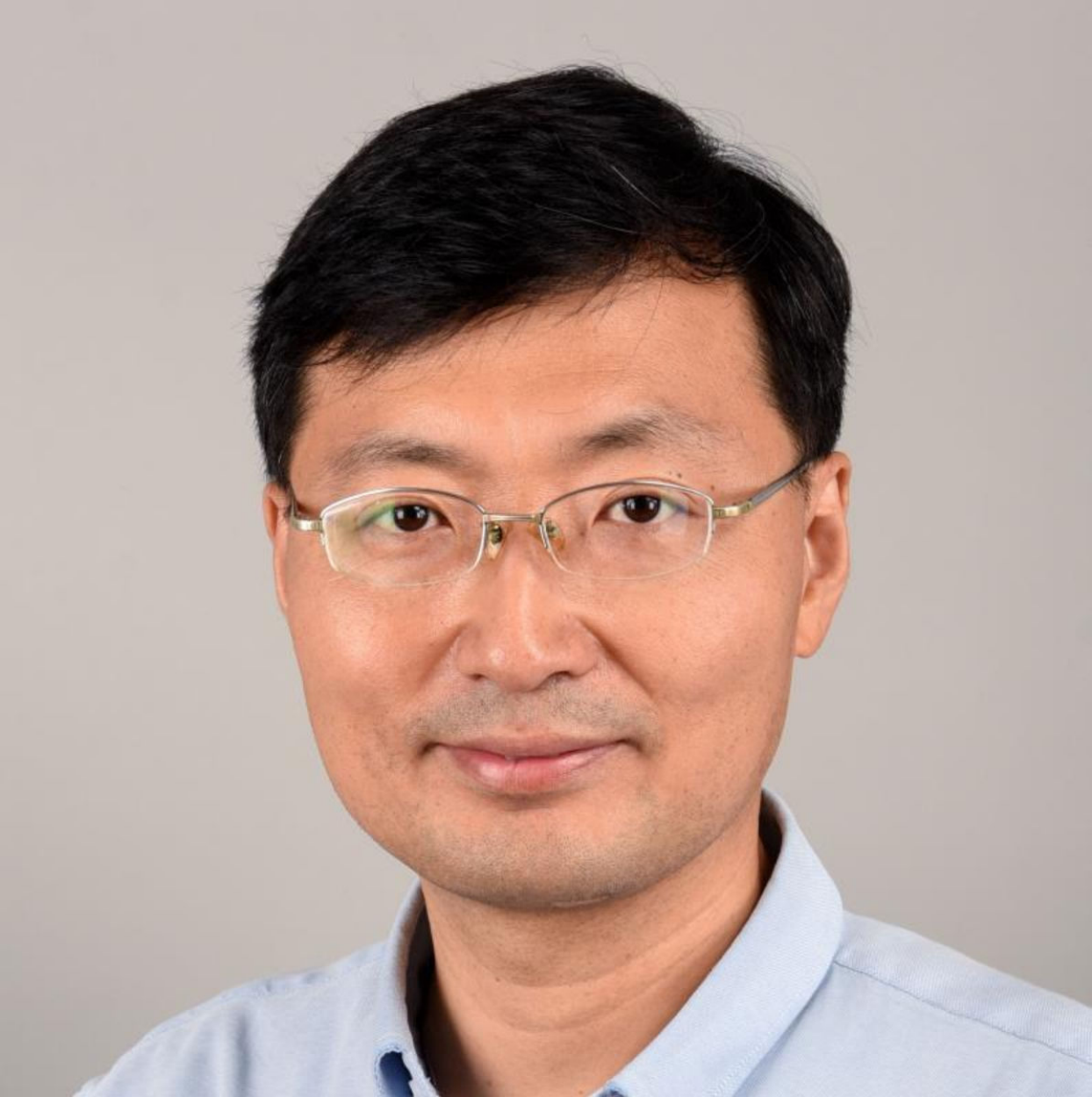}}]{\textbf{Ming Yang}}
received the Master and Ph.D. degrees from Tsinghua University, Beijing, China, in 1999 and 2003, respectively.\\
He is currently the Full Tenure Professor at Shanghai Jiao Tong University, the deputy director of the Innovation Center of Intelligent Connected Vehicles. He has been working in the field of intelligent vehicles for more than 20 years. He participated in several related research projects, such as the THMR-V project (first intelligent vehicle in China), European CyberCars and CyberMove projects, CyberC3 project, CyberCars-2 project, ITER transfer cask project, AGV, etc.
\end{IEEEbiography}

\vfill

\end{document}